\documentclass{article}

\usepackage[preprint]{neurips_2021}

\usepackage[utf8]{inputenc} 
\usepackage[T1]{fontenc}    
\usepackage{url}            
\usepackage{booktabs}       
\usepackage{amsfonts}       
\usepackage{nicefrac}       
\usepackage{microtype}      
\usepackage[dvipsnames]{xcolor}         
\usepackage{pifont}
\usepackage[most]{tcolorbox}
\usepackage{xstring}
\usepackage{xurl}
\usepackage{dsfont}

\usepackage{amsmath,bm,amsthm,amssymb,mathtools}
\usepackage{thmtools,thm-restate}

\usepackage{amsmath,amsfonts,bm}
\usepackage{mathtools}

\def\eqref#1{equation~\ref{#1}}

\def\1{\bm{1}}

\def\vm{{\bm{m}}}

\def\vq{{\bm{q}}}

\def\vs{{\bm{s}}}

\def\vv{{\bm{v}}}

\def\vx{{\bm{x}}}

\def\mA{{\bm{A}}}

\def\mS{{\bm{S}}}
\def\mT{{\bm{T}}}

\DeclareMathAlphabet{\mathsfit}{\encodingdefault}{\sfdefault}{m}{sl}
\SetMathAlphabet{\mathsfit}{bold}{\encodingdefault}{\sfdefault}{bx}{n}

\def\gG{{\mathcal{G}}}

\newcommand{\R}{\mathbb{R}}

\DeclarePairedDelimiter{\norm}{\lVert}{\rVert}

\newcommand{\defeq}{\vcentcolon=}

\newcommand*{\ldblbrace}{\{\mskip-5mu\{}
\newcommand*{\rdblbrace}{\}\mskip-5mu\}}

\def\a{{\mathbf a}}

\def\R{{\mathbb R}}

\def\s{{\mathbf s}}

\def\1{{\mathbf{1}}}

\def\vs{{\bm{s}}}

\usepackage{soul, color}
\newcommand{\Note}[1]{}
\renewcommand{\Note}[1]{#1}  

\usepackage{caption}
\usepackage{subcaption}
\usepackage{float}
\usepackage{graphicx}
\usepackage{multirow}
\usepackage{framed}
\usepackage{enumitem}
\usepackage{wrapfig}
\usepackage{algorithmic}
\usepackage{textcomp}
\usepackage{booktabs} 
\usepackage{tikz}
\usepackage{colortbl}
\usepackage{listings}
\usepackage{url}
\usepackage{todonotes}
\usepackage{tikz}
\usepackage{tikz-cd}
\usepackage{enumitem}
\usepackage{xcolor,colortbl}
\usepackage[autostyle]{csquotes}
\usepackage{marvosym}
\usepackage{xspace}
\usepackage{epigraph}
\usepackage[scr=boondoxo,scrscaled=1.05]{mathalfa}
\usepackage{pdflscape}

\usepackage[
    pagebackref=true,
    colorlinks = true,
    linkcolor = OliveGreen,
    urlcolor  = Cerulean,
    citecolor = MidnightBlue,
    anchorcolor = blue
]{hyperref} 
\usepackage[nameinlink]{cleveref}
\renewcommand*\backref[1]{\ifx#1\relax \else (Cited on page #1) \fi}

\def\upplus{\textsuperscript{\ding{60}}\ }
\newcommand{\appref}[1]{\hyperref[#1]{\upplus}}

\newcommand{\goingfurther}[1]{\begin{tcolorbox}[enhanced,attach boxed title to top left={yshift=-2mm,yshifttext=-1mm,xshift = 10mm},
colback=Goldenrod!10!white,colframe=Goldenrod!75!black,colbacktitle=white, coltitle=black, title=Going further,fonttitle=\bfseries,
boxed title style={size=small,colframe=Goldenrod!75!black} ]
#1
\end{tcolorbox}}

\newcommand{\badge}[2]{
\tcbox[on line,boxsep=0pt,left=2pt,right=2pt,top=2pt,bottom=2pt,colback=#1,fontupper=\small\color{white},boxrule=0pt,frame hidden]{#2}
}

\newcommand{\resourcetag}[1]{%
    \IfEqCase{#1}{%
        {article}{\badge{ForestGreen}{#1}}%
        {video}{\badge{RedOrange}{#1}}%
        {blog}{\badge{Rhodamine}{#1}}%
        {book}{\badge{Brown}{#1}}%
        {technical}{\badge{BlueViolet}{#1}}%
        {visual}{\badge{TealBlue}{#1}}%
    }[\PackageError{tag}{Undefined option to tag: #1}{}]%
}%

\title{A Hitchhiker's Guide to Geometric GNNs \\ for 3D Atomic Systems}

\author{
  Alexandre Duval\thanks{Equal first authors.}$^{*, 1, 2}$\quad Simon V. Mathis$^{*, 3}$\quad Chaitanya K. Joshi$^{*, 3}$\quad Victor Schmidt$^{*, 1, 4}$ \vspace{5pt} \\
  \textbf{Santiago Miret$^{5}$\quad Fragkiskos D. Malliaros$^{2}$\quad Taco Cohen$^{6}$} \vspace{5pt} \\
  \textbf{Pietro Liò$^{3}$\quad Yoshua Bengio$^{1, 4}$\quad Michael Bronstein$^{7}$} \vspace{10pt} \\
  $^1$Mila\quad $^2$Université Paris-Saclay\thanks{Université Paris-Saclay, CentraleSupélec, Inria.}\quad $^3$University of Cambridge\quad  $^4$Université de Montréal \vspace{5pt} \\
  $^5$Intel Labs\quad $^6$Qualcomm AI Research\thanks{Qualcomm AI Research is an initiative of Qualcomm Technologies, Inc.}\quad $^7$University of Oxford \vspace{10pt} \\
  $^*$Equal first authors. Correspondence to: \vspace{5pt} \\
  \texttt{\{alexandre.duval, schmidtv\}@mila.quebec}\vspace{5pt} \\ \texttt{\{simon.mathis, chaitanya.joshi\}@cl.cam.ac.uk}
}

\begin{document}

\maketitle

\begin{abstract}


    Recent advances in computational modelling of atomic systems, spanning molecules, proteins, and materials, represent them as \emph{geometric graphs} with atoms embedded as nodes in 3D Euclidean space. In these graphs, the geometric attributes transform according to the inherent physical symmetries of 3D atomic systems, including rotations and translations in Euclidean space, as well as node permutations. 
    In recent years, \emph{Geometric Graph Neural Networks} have emerged as the preferred machine learning architecture powering applications ranging from protein structure prediction to molecular simulations and material generation. Their specificity lies in the inductive biases they leverage --- such as physical symmetries and chemical properties --- to learn informative representations of these geometric graphs.
    
    In this opinionated paper, we provide a comprehensive and self-contained overview of the field of Geometric GNNs for 3D atomic systems. We cover fundamental background material and introduce a pedagogical taxonomy of Geometric GNN architectures: 
    (1) invariant networks, (2) equivariant networks in Cartesian basis, (3) equivariant networks in spherical basis, and (4) unconstrained networks. 
    Additionally, we outline key datasets and application areas and suggest future research directions. The objective of this work is to present a structured perspective on the field, making it accessible to newcomers and aiding practitioners in gaining an intuition for its mathematical abstractions.

\end{abstract}

\clearpage
\tableofcontents
\clearpage

\section*{Notation}

\renewcommand{\l}[1]{\ensuremath{\underline{#1}}} 
\newcommand{\g}[1]{\ensuremath{\vec{#1}}}         
\renewcommand{\c}[1]{\ensuremath{\mathbf{#1}}}    

\renewcommand{\s}[1]{\ensuremath{\MakeLowercase{#1}}}  
\renewcommand{\v}[1]{\ensuremath{\c{#1}}} 
\newcommand{\cs}[1]{\ensuremath{\c{\s{#1}}}}  
\newcommand{\cv}[1]{\ensuremath{\c{#1}}}  
\newcommand{\m}[1]{\ensuremath{\c{\MakeUppercase{#1}}}}  
\newcommand{\cm}[1]{\ensuremath{\c{\MakeUppercase{#1}}}} 

\newcommand{\ls}[1]{\ensuremath{\l{\s{#1}}}}  
\newcommand{\lv}[1]{\ensuremath{\l{\v{#1}}}}  
\newcommand{\lm}[1] {\ensuremath{\l{\m{#1}}}} 

\newcommand{\gv}[1]{\ensuremath{\g{#1}}}                  
\newcommand{\gt}[1]{\ensuremath{\g{#1}}}  

\newcommand{\lgv}[1]{\ensuremath{\l{\gv{#1}}}}  
\newcommand{\lgt}[1]{\ensuremath{\l{\gt{#1}}}}  

\newcommand{\cgv}[1]{\ensuremath{\g{\cv{#1}}}} 
\newcommand{\cgt}[1]{\ensuremath{\gt{\c{#1}}}} 

\newcommand{\clgv}[1]{\ensuremath{\c{\lgv{#1}}}}  
\newcommand{\clgt}[1]{\ensuremath{\c{\lgt{#1}}}}  

\newcommand{\cart}{\ensuremath{\mathfrak{c}}} 
\newcommand{\sphe}{\ensuremath{l}}            
\newcommand{\ucart}{\ensuremath{{[\cart]}}}   
\newcommand{\usphe}{\ensuremath{{(\sphe)}}}   
\newcommand{\ci}{\ensuremath{q}}              
\newcommand{\si}{\ensuremath{m}}              
\newcommand{\chan}{\ensuremath{c}}            
\newcommand{\node}{\ensuremath{n}}            

\newcommand{\gs}[1]{\ensuremath{\sum_{#1=1}^3}}

\newcommand{\graph}[1]{\ensuremath{\mathcal{\MakeUppercase{#1}}}}
\newcommand{\group}[1]{\ensuremath{\mathscr{{#1}}}}  

\newcommand{\gel}[1]{\ensuremath{\mathfrak{\MakeLowercase{#1}}}} 
\renewcommand*{\ldblbrace}{\{\mskip-5mu\{}
\renewcommand*{\rdblbrace}{\}\mskip-5mu\}}
\newcommand{\multiset}[1]{\ensuremath{\ldblbrace #1 \rdblbrace}}
\newcommand{\rot}{\ensuremath{\m{R}}}
\newcommand{\perm}{\ensuremath{\m{P}}}
\newcommand{\trans}{\ensuremath{\gv{t}}}
\newcommand{\reals}{\ensuremath{\mathbb{R}}}
\newcommand{\nei}{\ensuremath{\mathcal{N}}}
\newcommand{\basis}{\ensuremath{\psi}}


We assume the reader is familiar with basic machine learning terminology and common neural network architectures such as multi-layer perceptrons (MLPs). To keep the text clear and concise for readers with varying levels of prior knowledge, we include explanations of certain concepts in \Cref{app:lexicon}. These concepts are marked with a \upplus symbol.

\begin{tcolorbox}[enhanced,attach boxed title to top left={yshift=-2mm,yshifttext=-1mm,xshift = 10mm},
colback=cyan!3!white,colframe=cyan!75!black,colbacktitle=white, coltitle=black, title=Key notations,fonttitle=\bfseries,
boxed title style={size=small,colframe=cyan!75!black} ]
Throughout this paper, we use an intuitive \emph{visual grammar} to help readers separate key concepts mentally: (1) Bold characters represent collections (lists) of objects of the same type and have channel indices; (2) Underlined characters are learnable; (3) Characters with an arrow on top have a geometric meaning; (4) Geometric characters may carry component/node/channel indices as subscript and representation indices as superscript for higher tensor types. 
\end{tcolorbox}

This high-level visual grammar gives rise to the following notation: 

\begin{itemize}
    \item \textbf{Scalars}: Scalar quantities (simple numbers) are denoted by lowercase Latin letters $\s{s} \in \mathbb{R}$.
    
    \item \textbf{Geometric vectors}: Vector quantities with a geometric meaning carry an arrow on top to emphasise their geometric significance.
    In this work, all geometric vectors\appref{app:sec:geom-voc} are assumed to lie in a 3D space: $\cgv{v} \in \mathbb{R}^{b\times3},\gv{x} \in \mathbb{R}^3$. When evident from the context, we refer to geometric vectors as \emph{vectors}.
    
    \item \textbf{Geometric tensors}: Higher-order tensors\appref{app:sec:geom-voc} with geometric meaning are denoted by uppercase letters with an arrow on top $(\gt{T},\ldots)$. 
    To explictly distinguish spherical tensors $\gt{T}^\usphe$ we use a bracketed $l$ superscript with $\sphe$ indicating the type of the tensor\appref{app:sec:geom-voc}. For Cartesian tensors\appref{app:sec:geom-voc} $\gt{T}^\ucart$ of type $\cart$, we use a square-bracket superscript instead. 
    
    \item \textbf{Lists of quantities}: To represent lists of multiple quantities of the same type, we print characters in bold. For example, $\v{a}$ is a list of scalar quantities, $\cgv{v}$ is a list of geometric vectors, and $\cgt{T}^{(l)}$ is a list of spherical tensors of type $l$. For matrices of scalars (e.g. the adjacency matrix), we use bold uppercase letters $\m{a}$. Its entries are written $\s{a}_{ij}$ and row vectors $\cs{a}_{i}$. 
    
    \item \textbf{Learnable quantities}: Learnable quantities are marked with an underline. For example $\ls{a}$ is a learnable scalar, $\lv{a}$ is a list of learnable scalars, and $\lm{W}$ is a learnable weight matrix. 

    \item \textbf{Node indices}: We use $i, j, k, l$ to denote specific node indices. For example, $\gv{v}_{i}$ is a geometric vector at node $i$. Directed edges are denoted as tuples $(i, j)$. 

    \item \textbf{Channel indices}: We use $\chan$ to denote channel indices for lists or matrices of objects. For example, $\cgv{v}_{\chan_1}$ is the $\chan_1$-th geometric vector in $\cgv{v}$. Channel indices serve to distinguish the feature dimensions associated with the same atom, like atom type, atomic mass and electronegativity. 

    \item \textbf{Component indices}: To refer to the dimensions representing the different components or features of a data point (e.g. dimension of geometric vectors), we use $\ci$ for Cartesian and $\si$ for spherical geometric tensors. For instance, $\gt{T}^\ucart_{\ci_1}$ is a component of a Cartesian tensor while $\gt{Y}^\usphe_{\si}$ is a component of a spherical tensor.   
\end{itemize}

\paragraph{Special symbols}
\begin{itemize}
    \item \graph{G} is used to refer to a graph with vertex set $\mathcal{V}_{\graph{G}}$ and edge set $\mathcal{E}_{\graph{G}}$.
    \item \group{G} refers to the group \group{G}, e.g. $\group{G} = \text{SO}(3)$.
    \item $\sigma(\cdot)$ denotes a non-linear activation function $\sigma: \reals \to \reals$. When applied to a list of objects $\v{u}$, $\sigma(\v{u})$ is understood to act channel-wise.
    \item $f(\cdot)$ is a general function, often representing a neural network. 
    \item $\l{\m{W}}$ stands for the learnable weights of a neural network.
    \item $\basis(\cdot)$ stands for any basis function (e.g. radial basis, bessel function). See \Cref{app:subsec:basis-functions}.
    \item $\odot$ denotes the channel-wise product.
    \item $\otimes$ denotes the tensor product.
    \item $\vert \vert$ denotes concatenation.
    \item $\rot$, $\perm$, $\trans$ denote a rotation matrix, a permutation matrix and a translation vector, respectively.
    \item $\measuredangle ijk = \measuredangle(\g{x}_{ij}, \vec{x}_{jk})$ and 
    $\measuredangle ijkl = \measuredangle (\vec{x}_{ij}, \vec{x}_{jk}, \vec{x}_{kl})$ denote bond angle and dihedral (torsion) angles respectively.
    \item $a, b, c, d, e, n,\ldots$ are scalar quantities. $a, b$ are often used for space dimensions (e.g. number of features) and $d$ for the Cartesian space dimension ($d=3$ since we focus on 3D atomic systems). $c$ is used as the scalar cutoff value to create a graph from a point cloud. $n$, $e$ point to the number of nodes and edges in a graph.
    \item $(\m{A}, \m{S}, \cgt{v}, \cgv{x}, \cgv{c}, \cgv{o})$ denotes the graph $\graph{G}$ with adjacency matrix $\m{A} \in \reals^{n \times n}$, scalar feature matrix $\m{S} \in \reals^{n \times a}$, atom positions $\cgv{x} \in \reals^{n \times d}$, geometric feature vector $\cgt{v} \in \reals^{n \times b \times d}$. For periodic graphs, we  also include unit cell $\cgv{c} \in \mathbb{R}^{d \times d}$ and cell offsets $\cgv{o} \in \{-1,0,1\}^{e \times d}$ parameters. Nodes-labels or graph-labels are written $\s{y} \in \mathbb{R}$ or $\gv{y} \in \mathbb{R}^{n \times 3}$.
    \item $\nei_i$ refers to the neighbors of node $i$ in the graph $\graph{G}$. $\nei_i = \{j \in \mathcal{V}_\graph{G} \setminus \{i\}\text{ s.t. } \a_{ij} \neq 0\}$.
    \item $\v{h}_i, \v{m}_{ij}, \cgv{m}_{ij}, \gv{x}_{ij}, \s{d}_{ij}, \hat{x}_{ij}$ relate to message passing. $\v{h}_i$ refers to atom $i$'s hidden representation, $\v{m}_{ij}  \in \reals^{a}$ to the scalar message from node $j$ to node $i$ and $\cgv{m}_{ij} \in \reals^{a \times d}$ to the geometric message. We use the superscript $\bullet^{(t)}$ to denote the $t^{th}$message passing layer or iteration. $\gv{x}_{ij} = \gv{x}_{i} - \gv{x}_{j}$ is the relative position or displacement between two nodes, $\s{d}_{ij}=\vert \vert \gv{x}_{ij} \vert \vert$ the distance separating them, and $\hat{x}_{ij} = \gv{x}_{ij} /  \vert \vert \gv{x}_{ij} \vert \vert $ the unit directional vector. 
\end{itemize}

\vspace{3mm}

\begin{tcolorbox}[enhanced,attach boxed title to top left={yshift=-2mm,yshifttext=-1mm,xshift = 10mm},
colback=cyan!3!white,colframe=cyan!75!black,colbacktitle=white, coltitle=black, title=Clarification,fonttitle=\bfseries,
boxed title style={size=small,colframe=cyan!75!black} ]
We use the words \emph{vector}, \emph{matrix} and \emph{tensor} from the mathematical perspective, which is distinct from the common usage of these words in the wider machine learning literature. Tensors are \textbf{not} simply multidimensional arrays of numbers. Instead, the word \emph{tensor} signifies that the object, in addition to being representable via a multidimensional array of numbers, also has certain properties and follows transformations in multidimensional spaces. In machine learning language, our word \emph{tensor} therefore refers to multidimensional arrays of numbers which follow certain transformation rules. Similarly, \emph{lists} denote ordered collections, not computer science data structures.
\end{tcolorbox}


\clearpage

\begin{figure}[t!]
    \centering
    \makebox[\textwidth][c]{\includegraphics[width=1.2\textwidth]{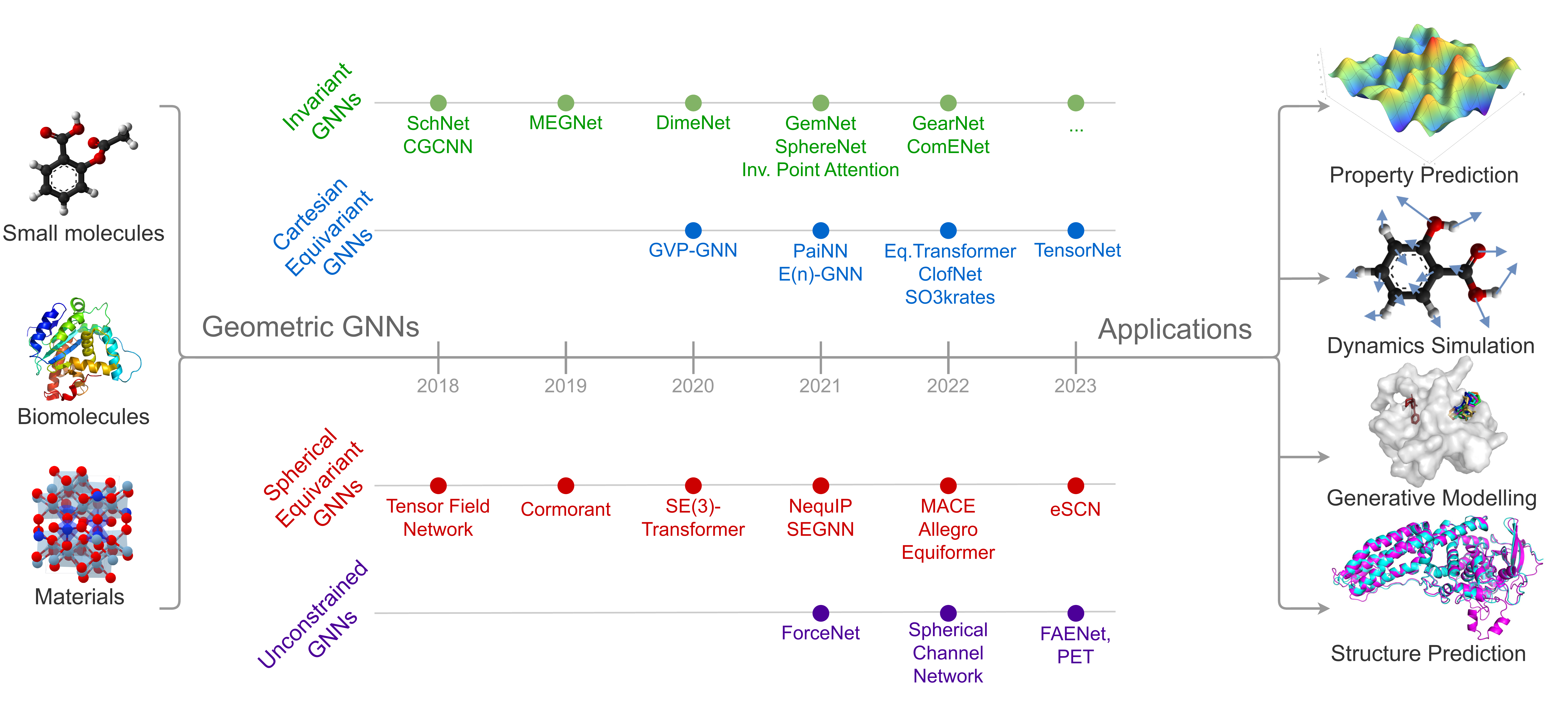}}%
    \caption{\textbf{Timeline of key Geometric GNNs for 3D atomic systems}, characterised by the type of intermediate representations within layers\protect\footnotemark.
    This survey presents a self-contained overview of Geometric GNN architectures and their applications in modeling 3D atomic systems.
    }
    \label{fig:timeline}
\end{figure}

\footnotetext{This is a partial selection of representative architectures; an exhaustive list is provided on \href{https://github.com/AlexDuvalinho/geometric-gnns}{Github}.}


\section{Introduction}
\label{sec:intro}

Graphs are a powerful and general mathematical abstraction. They can represent complex relationships and interactions across fields as diverse as social networks, recommendation systems, molecular structures, and biological interactomes. 
Graph Neural Networks (GNNs) \citep{ scarselli2008graph, kipf2017semi, velickovic2018graph} are the current state-of-the-art machine learning methods for processing graph data and making predictions over nodes, edges or at the global graph level.
GNNs learn latent representations of nodes through \emph{message passing} \citep{gilmer2017neural}\appref{app:subsec:message-passing}, which enables the model to extract information about the local subgraph around each node. 
The Transformer architecture \citep{vaswani2017attention} for natural language processing can also be viewed as a type of GNN, where the nodes are words and the graph is assumed to be fully connected \citep{joshi2020transformers}.

\begin{figure}[t!]
    \centering
    \includegraphics[width=0.8\linewidth]{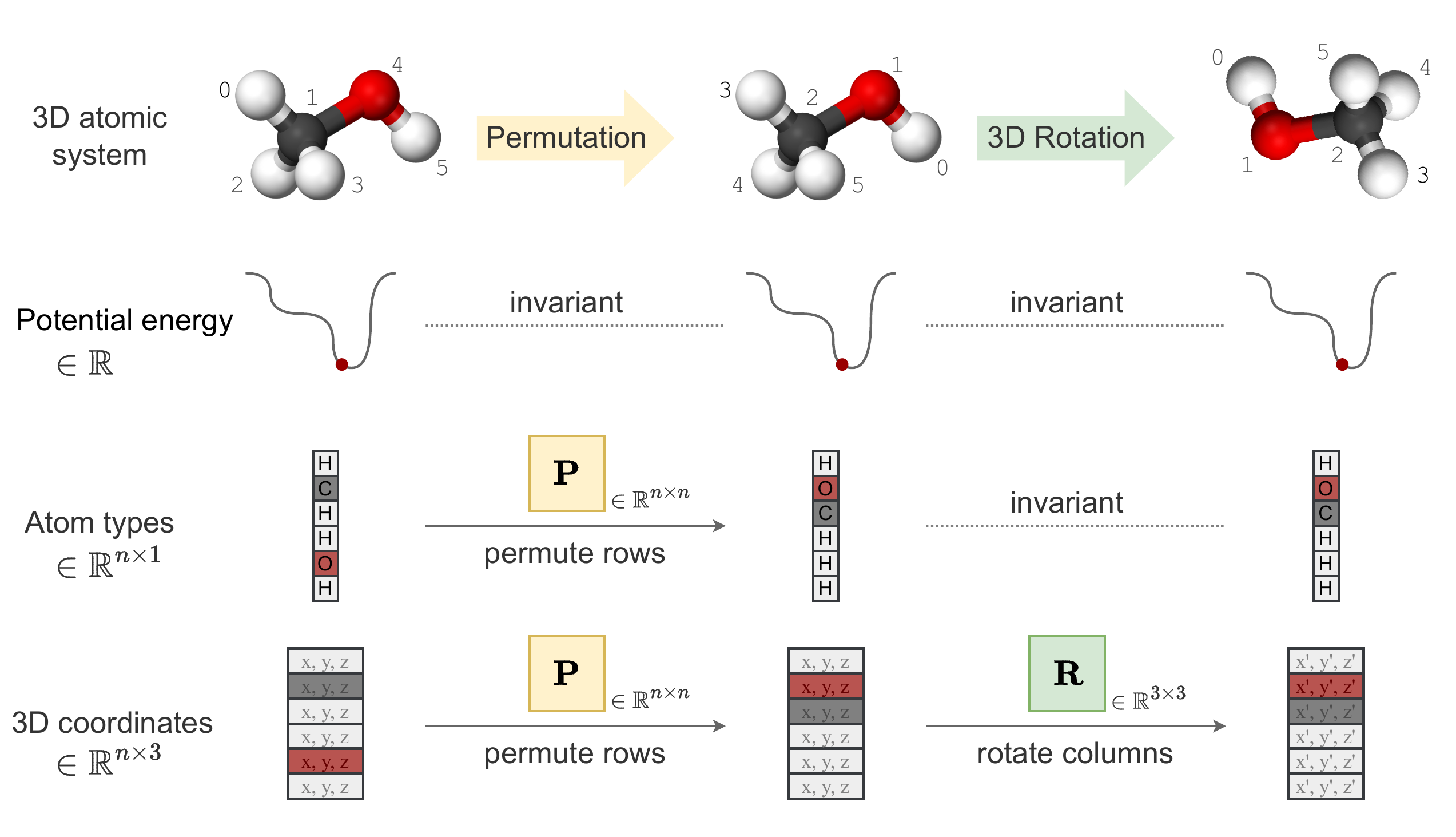}
    \caption{\textbf{Physical symmetries of 3D atomic systems.} 
    The ordering of atoms/nodes in the system is arbitrary.
    Additionally, global rotations or translations of the system in 3D Euclidean space will lead to an equivalent transformation of 3D coordinates and other geometric attributes. 
    Global properties of the system such as the potential energy are invariant to both permutation and physical symmetries.
    Geometric GNNs explicitly account for both permutation symmetry and physical transformation behaviours when modeling 3D atomic systems, while standard GNNs solely account for permutations.
    }
    \label{fig:mol-energy-forces}
\end{figure}

Graphs are purely topological objects, in the sense that they specify only how entities (nodes) are connected, but not their spatial layout (`geometry'). For example, a social network represents friendship relations between people, but not where these people live.  
\emph{Geometric graphs} are a type of graphs where nodes are additionally endowed with geometric information pertaining to the physical world, such as their spatial positions. 
A prototypical example we consider in this paper are molecules: the nodes represent atoms embedded in 3D Euclidean space with scalar attributes (e.g. atom type) and geometric attributes (e.g. position, velocity, or forces). Both are essential to accurately model the properties of a physical system.
Because these properties are independent of the chosen reference frame\appref{app:sec:geom-voc}, the geometric attributes are typically either invariant (independent) or equivariant (changing in the same way) under symmetry groups acting on them\appref{app:sec:groups}.
Consider the illustration in \Cref{fig:mol-energy-forces}: molecular properties such as the potential energy of an isolated molecule remain the same no matter how we rotate or translate the molecule in space; it is thus \emph{invariant} to Euclidean transformations.
On the other hand, rotating or translating the molecule will lead to an equivalent transformation of the directional forces acting on each atom; atomic forces are \emph{equivariant} to Euclidean transformations. 


Thus, unlike generic graph data, geometric graphs contain additional attributes with known transformation behaviours under physical symmetries. GNNs which do not take physical symmetries into account are considered ill-suited to model geometric graphs, as treating geometric attributes in the same manner as standard node features would no longer retain their physical meaning and transformation semantics \citep{bronstein2021geometric, bogatskiy2022symmetry}. 

Geometric Graph Neural Networks are an emerging class of GNNs for modeling geometric graphs constructed from 3D atomic systems. Geometric GNNs learn latent representations which enforce the appropriate physical symmetries on geometric attributes, enabling the model to better capture both geometric and relational structure in 3D systems. Geometric GNNs are the core architecture behind recent applications in protein structure prediction \citep{jumper2021highly}, protein design \citep{dauparas2022robust}, molecular simulation \citep{batzner2022nequip} and materials discovery \citep{zitnick2020introduction}.



Given the progress in Geometric GNNs for 3D atomic systems, summarised in \Cref{fig:timeline},
newcomers to the field often find themselves lost in the current `zoo' of different models.  
This hitchhiker's guide aims to provide a comprehensive and pedagogical overview of the Geometric GNN modeling pipeline, describing all the architectural building blocks, highlighting key conceptual ideas, and outlining impactful future directions.
Our primary goal is to serve as a guide for both newcomers and experienced researchers alike to navigate the exciting field of geometric graph learning. 




The rest of the paper is organized as follows:
\begin{itemize}
    \item \Cref{sec:background} provides all necessary \textbf{background materials} for geometric graphs and Geometric GNNs, including explanations of key conceptual ideas. We aim to provide a solid foundation for understanding the subsequent content.
    \item \Cref{sec:geom-gnns} describes \textbf{all components of the Geometric GNN pipeline} such as input creation, embedding, interaction, and output blocks. We detail the variations within each component, allowing readers to grasp the intricacies and design choices involved.
    \item \Cref{sec:invariant-gnns}, \ref{ssec:equivariant_gnns}, and \ref{subsec:non-spm-gnns} introduce a \textbf{novel taxonomy} that categorizes Geometric GNNs into four distinct families of methods: invariant, equivariant with Cartesian tensors, equivariant with spherical tensors, and unconstrained. This taxonomy offers a nuanced classification of existing architectures and establishes links between the different families.
    \item \Cref{sec:appli-data-coding} explores various \textbf{datasets} and \textbf{applications} of Geometric GNNs, guiding the selection of evaluation methodology. 
    \item \Cref{sec:discussion} concludes the survey by identifying key areas for \textbf{future research}, shedding light on untapped opportunities in the field.
\end{itemize}
Additionally, the appendix contains definitions, refreshers, and complementary information on various topics. The accompanying \href{https://github.com/AlexDuvalinho/geometric-gnns}{Github repository} offers an exhaustive list of Geometric GNNs and datasets along with their key properties, which we hope the community will keep up-to-date. 





\clearpage

\section{Preliminaries}
\label{sec:background}


\subsection{Graphs and Graph Neural Networks}
\label{sec:normal-graph}

\begin{figure}[h!]
    \centering
    \begin{subfigure}[b]{0.3\linewidth}
        \centering
        \includegraphics[width=\linewidth]{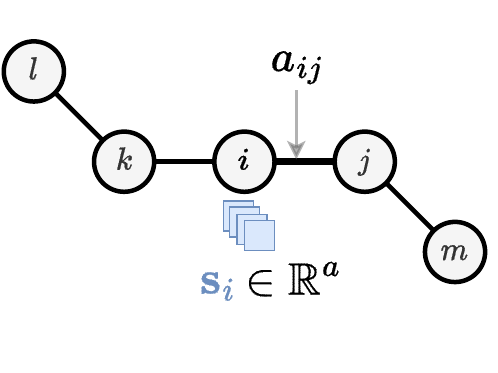}
        \caption{An attributed graph}
        \label{fig:normal-graph}
    \end{subfigure}
    \hfill
    \begin{subfigure}[b]{0.3\linewidth}
        \centering
        \includegraphics[width=\linewidth]{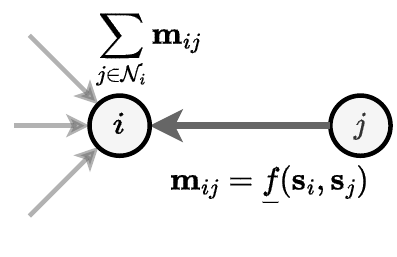}
        \caption{Message passing}
        \label{fig:mpnn}
    \end{subfigure}
    \hfill
    \begin{subfigure}[b]{0.3\linewidth}
        \centering
        \includegraphics[width=\linewidth]{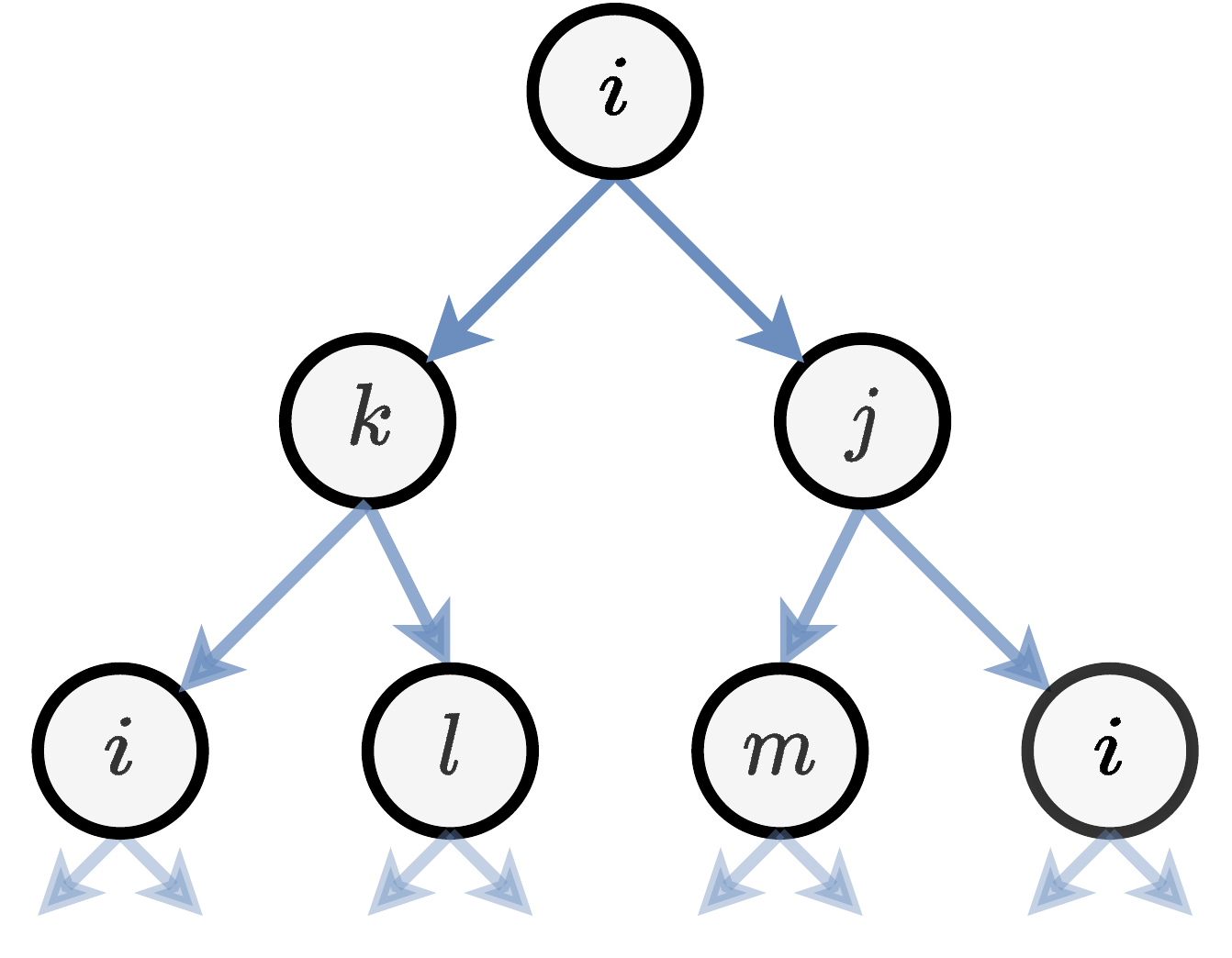}
        \caption{GNN computation tree}
        \label{fig:comp-tree}
    \end{subfigure}
    \caption{\textbf{Graphs and Graph Neural Networks.}
    (a) Graphs model a set of entities as nodes, with edges denoting relationships and structure among them. 
    (b) GNNs build latent representations of graph data through message passing operations, where each node performs learnable feature aggregation from its local neighbourhood.
    (c) Stacking $L$ message passing layers enables GNNs to send and aggregate information from $L$-hop subgraphs around each node.
    }
    \label{fig:graph-mpnn-iso}
\end{figure}

\textbf{Graphs. }
Graphs are used to model complex and interconnected systems in the real world, ranging from molecules and knowledge graphs to social networks and recommendation systems.
Formally, an attributed graph $\graph{G} = ( \m{A}, \m{S} )$ is a set $\mathcal{V}$ of $n$ nodes connected by edges, as shown in \Cref{fig:normal-graph}.
$\m{A}$ denotes an $n \times n$ adjacency matrix where each entry $a_{ij} \in \{ 0, 1 \}$ indicates the presence or absence of an edge connecting nodes $i$ and $j$.
The matrix of \emph{scalar} features $\m{S} \in \reals^{n \times a}$ stores attributes $\cs{s}_i \in \reals^a$ associated with each node $i$. For example, in molecular graphs (2D), each node contains information about the atom type (e.g. hydrogen, carbon), and edges represent bonds among atoms. 

Typically, the nodes in a graph have no canonical or fixed ordering and can be shuffled arbitrarily, resulting in an equivalent shuffling of the rows and columns of the adjacency matrix $\m{A}$.
Thus, accounting for permutation symmetry is a critical consideration when designing machine learning models for graphs.
One can also consider more complex definitions of a graph, including multi-relational graphs or higher-order topological variants such as hypergraphs, but we will proceed with a basic definition. 


\textbf{Graph Neural Networks. }
Graph Neural Networks (GNNs) \citep{goller1996learning, sperduti1997supervised, gori2005new, scarselli2008graph} are bespoke neural networks for graph data that incorporate permutation symmetry.
In recent years, modern variants of GNNs have emerged as the architecture of choice for machine learning with large-scale and real-world graph data \citep{kipf2017semi, velickovic2018graph}. GNNs build actionable node representations through message passing operations \citep{gilmer2017neural,battaglia2018relational} where each node updates its feature vector by aggregating features from its local neighbourhood $\nei_i$ in the graph. In simpler terms, neighbouring nodes (or edges) exchange information and influence each other’s embedding update. Thus, node features represent the local sub-graph structure around the node and stacking several message passing layers propagates node features beyond local neighbourhoods.

Node features $\v{s}_i$ are updated from iteration $t$ to $t+ 1$ in three steps. (1) Compute ``messages'' between the node of interest $i$ and each of its neighbour $\nei_i$ in the graph, via a learnable  $\l{\textsc{MSG}}$ function; (2) Aggregate all messages coming from $\nei_i$ via a fixed permutation-invariant aggregation operator $\oplus$ (e.g. sum, mean); (3) Update the representation of node $i$ via a learnable function $\l{\textsc{UPD}}$, typically using both aggregated messages and its own representation as input. In practice, $\l{\textsc{MSG}}$ and $\l{\textsc{UPD}}$ are neural networks whose definition has been the focus of much of GNN methodology research. Formally, the message passing GNN paradigm is expressed as:
\begin{align}
    \v{m}_{ij}^{(t)} &= \l{\text{MSG}} \ \big( \v{s}_i^{(t)}, \ \v{s}_j^{(t)}) \nonumber \\
    \v{s}_i^{(t+1)} &= \l{\text{UPD}} \ \big(\v{s}_i^{(t)}, \ \underset{j \in \nei_i}{\bigoplus} \ \v{m}_{ij}^{(t)}\big) 
    \label{eq:standard-mpnn}
\end{align}

The features derived in the final iteration $L$, i.e. the last message passing layer, are mapped to graph-level, node-level or edge-level predictions via a permutation-equivariant readout. For both node-level and edge-level tasks, we can learn a shared classifier (e.g. MLP) on node/edge representations, $\v{s}_i^{(L)}$ or $f (\v{s}_i^{(L)}, \v{s}_j^{(L)})$, where $f$ is any function, e.g. a simple concatenation. For graph regression or classification tasks, we first need to derive a graph representation from learned node representations $\{ \cs{s}_i^{(L)} \}$, using a permutation-invariant readout function $\underset{i \in \mathcal{V}}{\bigoplus} \v{h}_i^{L}$. This is called \textit{graph pooling}\footnote{This operation is similar to the pooling layers commonly found in CNNs: their goal is to coarsen representations and aggregate information from all the node features into a single feature for the entire graph.}. Then, we can learn a classification or regression head over the resulting flattened vector.

\textbf{Applications of GNNs. }
GNNs have demonstrated their utility across a wide range of applications, ranging from
recommendation systems \citep{hamilton2017inductive},
social networks \citep{monti2019fake, benamira2020semisupervised},
transportation networks \citep{derrow2021eta},
weather forecasting \citep{lam2023learning},
and, perhaps most importantly, for accelerating and augmenting scientific discovery \citep{zhang2023artificial, wang2023scientific}.
This survey focuses on the later, introducing the family of graphs and GNNs powering recent advances in modeling atomic systems in 3D space, including molecular dynamics simulation \citep{batzner2022nequip}, protein folding
\citep{jumper2021highly} and design \citep{dauparas2022robust}, as well as materials discovery \citep{zitnick2020introduction}.



\begin{figure}[h!]
    \centering
    \includegraphics[width=\linewidth]{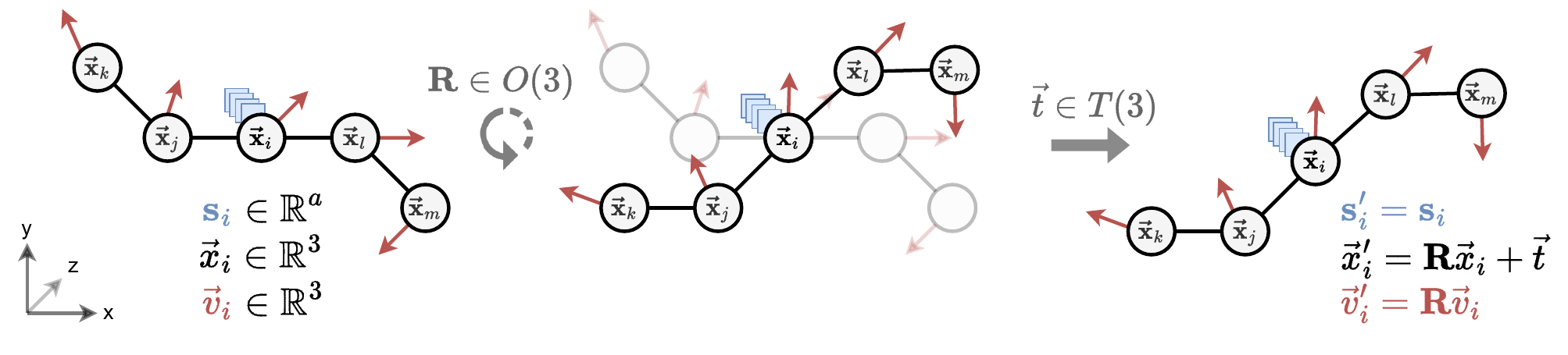}
    \caption{\textbf{Geometric graphs and Euclidean symmetries. }
        Geometric graphs embedded in 3D Euclidean space model systems with both geometry and relational structure, such as molecules and materials.
        The geometric attributes transform along with Euclidean transformations of the system:
        (1) The group of rotations $\text{SO}(3)$, or rotations and reflections $\text{O}(3)$, acts on the vector features $\cgv{v}$ and coordinates $\cgv{x}$;
        and
        (2) The translation group $\text{T}(3)$ acts on the coordinates $\cgv{x}$.
        Scalar features remain invariant to Euclidean transformations.
        Note that this setup generalises to multiple vector features $\cgv{v}$ or higher-order tensor type features. 
    }
    \label{fig:geom-graph-rot}
\end{figure}

\subsection{Geometric Graphs}
\label{subsec:geom-graphs}

\textbf{Geometric graphs.}
Geometric graphs are used to model systems containing both relational structure and geometry embedded in $d$-dimensional Euclidean space (for most real-world applications, $d=3$D space). As illustrated in \Cref{fig:geom-graph-rot}, a geometric graph $\graph{G} = ( \m{A}, \m{S}, \cgt{v}, \cgv{x} )$ is an attributed graph with scalar features $\m{S}$ that is also decorated with geometric attributes: node coordinates $\cgv{x} \in \reals^{n \times d}$ and (optionally) $b$ vector features $\cgt{v} \in \reals^{n \times b \times d}$, sometimes denoted $\cgv{v}$ for simplicity.

In biochemistry and material science, geometric graphs have to be constructed from the underlying point cloud $( \m{S}, \cgt{v}, \cgv{x} )$, which constitutes the typical input data of a set of atoms in 3D space. 
For example, molecules are often represented as a set of atoms/nodes which contain information about the atom type (a scalar feature) and its 3D spatial position (the coordinates), as well as other geometric vector quantities such as velocity or forces.
Nodes are generally connected by edges using a predetermined radial cutoff distance $c$, such that the adjacency matrix is defined as $a_{ij} = 1 \text{ if } \Vert \gv{x}_i-\gv{x}_j \Vert_2 \leq c$, or $0$ otherwise, for all $a_{ij} \in \m{A}$.
In contrast, the 2D molecular graph representation from \Cref{sec:normal-graph} does not provide any information about the spatial location or geometric attributes of a molecule.
Conventional procedures for geometric graph creation beyond radial cutoffs are described in \Cref{subsec:input-rep}.

\textbf{Permutation and Euclidean symmetries}. 
As illustrated in \Cref{fig:mol-energy-forces}, the key factors distinguishing geometric graphs from standard graphs are the transformation behaviours of the geometric attributes under Euclidean symmetries.
Geometric attributes are symmetric under physical transformations of the system: translations, rotations, and sometimes reflections, while scalar features remain invariant or unchanged. The following symmetry groups are relevant for geometric graphs (see \Cref{fig:geom-graph-rot}):
\begin{itemize}
    \item \textbf{Permutation.} The permutation group over $n$ elements $\text{P}_n$ acts via a permutation matrix $\perm$ on the graph attributes as $\perm \graph{G} := (\perm \m{A} \perm^\top, \perm \m{S}, \perm \cgv{v}, \perm \cgv{x})$.
    This entails shuffling the ordering of rows of the feature tensors and follows directly from the fact that a graph has no canonical ordering of its nodes.
    \item \textbf{Rotation (and reflection)} The group of rotation $\text{SO}(d)$ or rotations and reflections $\text{O}(d)$, denoted interchangeably by $\group{G}$, acts via an orthogonal transformation matrix $\rot \in \group{G}$ on the vector feature $\cgv{v}$ and on the coordinates $\cgv{x}$ as $\rot \gG := (\m{A}, \m{S}, \cgv{v}\rot, \cgv{x}\rot)$.
    Vector features and coordinates are physical quantities measured from an arbitrary frame of reference, so rotating the frame of reference implies an equivalent rotation of these quantities.
    On the other hand, scalar features are generally denoting categorical information (such as the atom type of a node) that remains unchanged or invariant under rotations. Whether equivariance to reflections is desired or not depends on the application, as explained in \Cref{app:sec:groups}. 
    \item \textbf{Translation.} The group of translations $\text{T}(d)$ acts via a translation vector $\trans \in \text{T}(d)$ on the coordinates $\cgv{x}$ as $\gv{x}_i + \trans$ for all nodes $i$.
    The coordinates of each node are determined with respect to a single point in space (called the origin), so translating the origin leads to an equivalent translation of the coordinates.
    Importantly, translations do \emph{not} impact the vector features at each node as their values are always determined relative to the coordinate of that particular node.
\end{itemize}

\begin{tcolorbox}[enhanced,attach boxed title to top left={yshift=-2mm,yshifttext=-1mm,xshift = 10mm},
colback=cyan!3!white,colframe=cyan!75!black,colbacktitle=white, coltitle=black, title=Opinion,fonttitle=\bfseries,
boxed title style={size=small,colframe=cyan!75!black} ]
The interplay between discrete (permutation) and continuous (Euclidean) symmetries makes the modeling of geometric graph data very vibrant, bringing together mathematical ideas from graph theory, topology, geometry, functional analysis, and quantum mechanics. Since node permutation equivariance is handled in traditional GNNs, we focus on Euclidean symmetries (specific of 3D atomic systems) in this work. A short refresher on group theory with a focus on geometric graphs is provided in \Cref{app:sec:groups}. 
\end{tcolorbox}


\begin{figure}[h!]
    \centering
    \begin{subfigure}[b]{0.32\linewidth}
        \centering
        \includegraphics[width=\linewidth]{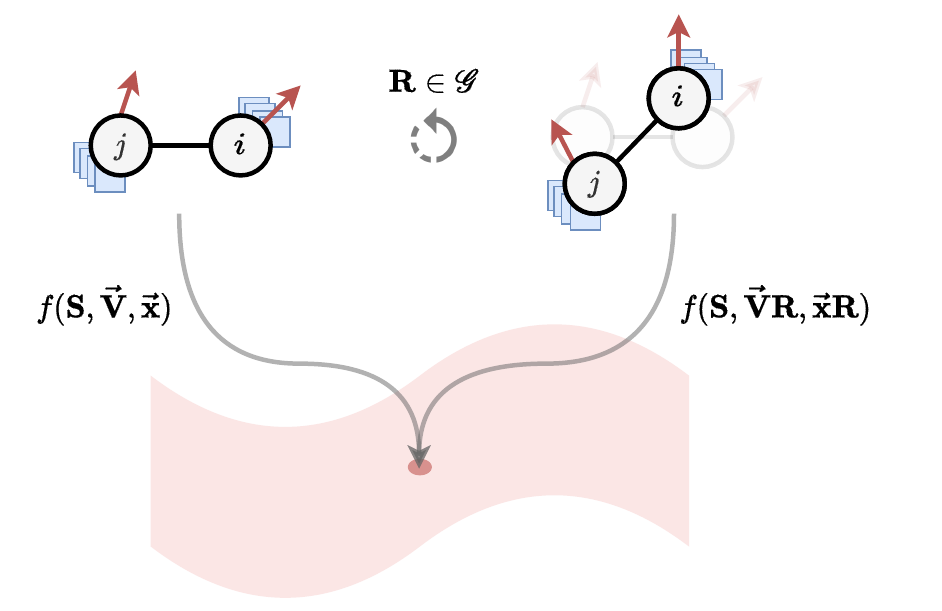}
        \caption{$\group{G}$-invariant function}
        \label{fig:invariant-fn}
    \end{subfigure}
    \hfill
    \begin{subfigure}[b]{0.32\linewidth}
        \centering
        \includegraphics[width=\linewidth]{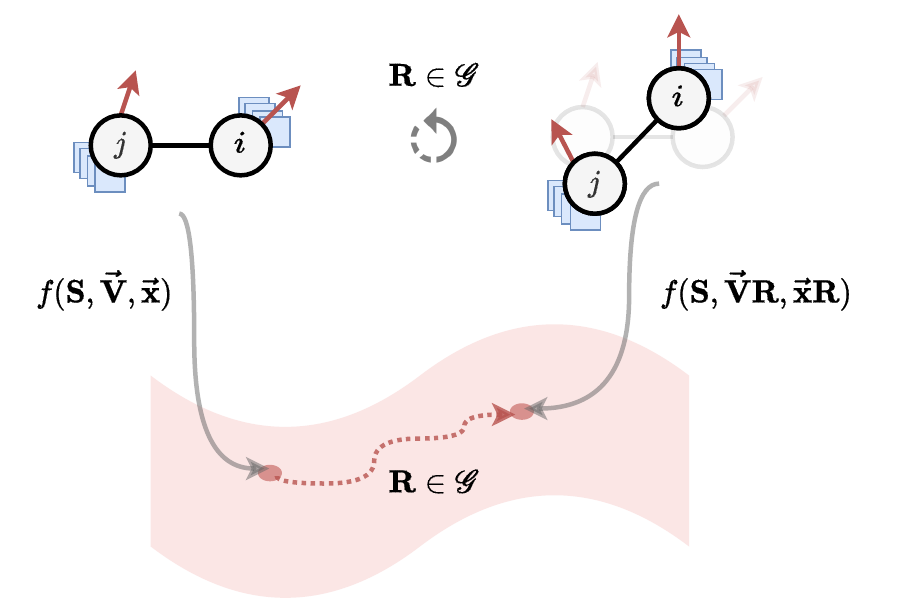}
        \caption{$\group{G}$-equivariant function}
        \label{fig:equivariant-fn}
    \end{subfigure}
    \hfill
    \begin{subfigure}[b]{0.32\linewidth}
        \centering
        \includegraphics[width=\linewidth]{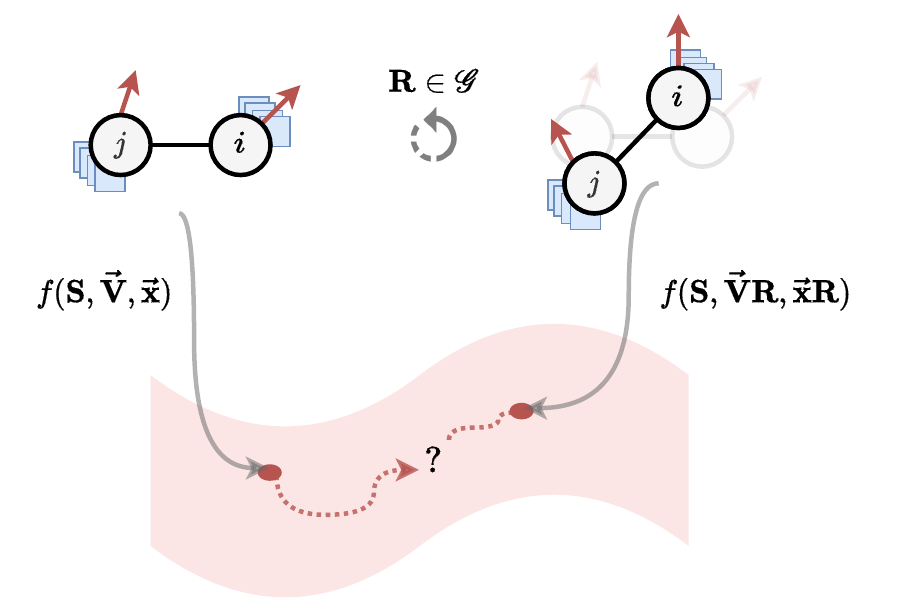}
        \caption{$\group{G}$-unconstrained function}
        \label{fig:nsp-fn}
    \end{subfigure}
    \caption{\textbf{Invariant, equivariant, and unconstrained functions.}
        The output of $\group{G}$-invariant functions remains unchanged regardless of transformations applied to the input. $\group{G}$-equivariant functions, on the other hand, exhibit transformations in the output that are equivalent to the transformations in the input. Finally, $\group{G}$-unconstrained functions do not have predictable or known transformations of the output when the input undergoes transformations.
    }
    \label{fig:invariant-equivariant}
\end{figure}


\textbf{Functions on geometric graphs. }
Before describing GNNs specialised for geometric graphs, we first define three classes of functions that are used to construct Geometric GNN layers.
Following the Geometric Deep Learning blueprint \citep{bronstein2021geometric}, we denote the action of a group $\group{G}$ on a space $X$ by $\gel{g} \cdot x$, for $\gel{g} \in \group{G}$ and $x \in X$.
If $\group{G}$ acts on spaces $X$ and $Y$, we say:
\begin{itemize} \label{def:equivariance}
    \item A function $f: X \to Y$ is $\group{G}$-\emph{invariant} if $f(\gel{g} \cdot x) = f(x)$, \textit{i.e.} the output remains unchanged under transformations of the input, as shown in \Cref{fig:invariant-fn}.
    \item A function $f: X \to Y$ is $\group{G}$-\emph{equivariant} if $f(\gel{g} \cdot x) = \gel{g} \cdot f(x)$, \textit{i.e.} a transformation of the input must result in the output transforming correspondingly
    , as shown in \Cref{fig:equivariant-fn}.
    \item A function $f: X \to Y$ which is not $\group{G}$-\emph{invariant} nor $\group{G}$-\emph{equivariant} is referred to as $\group{G}$-\emph{unconstrained}. The transformation of the input results in an unknown change in the output, as shown in \Cref{fig:nsp-fn}.
\end{itemize}



\clearpage


\section{From GNNs to Geometric GNNs}
\label{sec:geom-gnns}

\begin{figure}[t!]
    \centering
    \includegraphics[width=\linewidth]{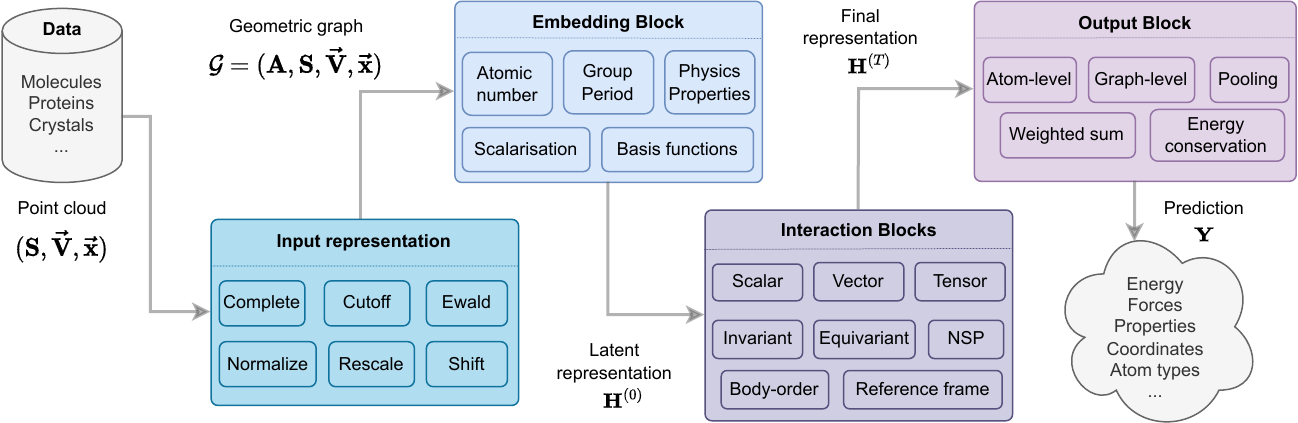}
    \caption{
        \textbf{Common Geometric GNN inference pipeline for 3D atomic systems.}
        The input representation phase (\Cref{subsec:input-rep}) involves the creation of a geometric graph $\graph{G}$ from the given point cloud $( \mS, \cgt{v}, \cgv{x} )$ data, which often depends on application-specific pre-processings (e.g. materials, small molecules, proteins). The Embedding block learns a representation for each node/edge of the graph,
        which is updated by Geometric GNN layers in repeated Interaction Blocks (\Cref{subsec:interaction-blocks}). Finally, the Output block (\Cref{subsec:output-block}) 
        computes node-level, edge-level or graph-level predictions. The key distinctions between Geometric GNNs essentially lie in the Interaction block, where the message passing scheme varies significantly, mainly depending on how data symmetries are enforced.
    }
    \label{fig:gnn-pipeline}
\end{figure}

Traditional Graph Neural Networks (GNNs) are not well-suited for tasks involving geometric graphs, primarily due to their inability to predict real-world quantities while adhering to physical symmetries. For example, the energy of an atomic system remains unchanged no matter how the 3D system is rotated or translated. In contrast, Geometric GNNs are designed to capitalize on the symmetries inherent in these systems, incorporating them into the core of the model. This can be seen as an \textit{inductive bias}\appref{app:sec:inductivebiases} that is built into the model architectures. In general, by confining the scope of learnable functions to desirable ones, these models ensure predictions align with the principles of physics, which in turn enhances generalization and data efficiency throughout the learning process.

Typically, making predictions on 3D atomic systems using Geometric GNNs involves a specific way (1) to represent the problem, (2) to learn meaningful atom embeddings, and (3) to predict desired physical quantities from the learned representations. In subsequent sections, we describe each part of the pipeline, represented in \Cref{fig:gnn-pipeline}.


\subsection{Input preparation}
\label{subsec:input-rep}

The minimal `raw' data for an atomic system is typically a set of atom types ($\m{S}$) and positions in 3D space $(\cgv{x})$, i.e. a 3D point cloud $(\m{S}, \cgv{x})$.
The geometric graph is constructed from the underlying point cloud to model pairwise interactions among the atoms, and other physical descriptors and attributes are attached to the nodes and edges to prepare the input representation into Geometric GNNs.

Note that this survey uses the terms `atoms' in an atomic system and `nodes' in the corresponding geometric graph interchangeably.
Geometric GNNs typically operate on a subset of all input atoms.
For instance, Hydrogen atoms are generally ignored for computational efficiency when modeling small molecules as well as biomolecules.
For larger systems such as proteins and nucleic acids, models may operate at different levels of granularity, such as only using the alpha Carbon atoms to represent an entire residue.

\begin{figure}
    \centering
    \includegraphics[width=\textwidth]{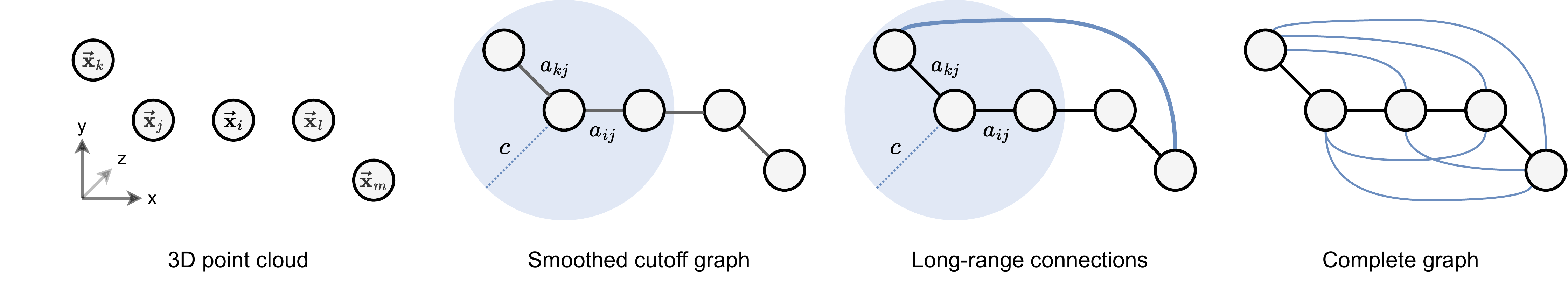}
    \caption{\textbf{From point clouds to geometric graphs}. The original 3D point cloud is transformed into a representative geometric graph. Examples include cutoff graphs, long-range connections, and complete graphs.}
    \label{fig:input-rep}
\end{figure}

\textbf{Graph representation}. Given the pivotal role of atomic interactions in determining a system's properties, constructing a geometric graph, i.e. an adjacency matrix $\mathbf{A}$, from the 3D point cloud becomes a natural direction to pursue. Indeed, this approach then enables the design of a Geometric GNN that effectively captures both the topological and feature-related information of the system. Various strategies to construct the initial geometric graph have been explored in the literature, illustrated in \Cref{fig:input-rep}:

\begin{itemize}
    \item \textbf{Via a complete graph} where every atom is connected to every other atom, capturing all potential atomic interactions including pairwise dependencies and potential long-range effects. This representation is motivated by the physical principle that atoms in a system can interact with each other to some degree. 
    Using a complete graph (with pairwise distances as edge weights) allows for a comprehensive analysis of an atomic system and has been the preferred solution on small molecules (MD17 \citep{duvenaud2015convolutional}, QM9 \citep{ramakrishnan2014quantum}). However, it is computationally demanding and leads to unnecessary complexity, especially in large systems like proteins, which is why the approaches below are often preferred.
    It should be noted however that transformer-based architectures exist which are capable of handling large graphs \citep{brehmer2023geometric}.
    \item \textbf{Via a cutoff graph}, where there exists an edge between any two atoms if their relative distance $d_{ij} = || \gv{x}_i - \gv{x}_j ||$ is below a certain threshold ``cutoff'' distance $c$, expressed in Angstrom (e.g., $c=6\textup{\AA}$) \citep{schutt2017schnet, gasteiger2021gemnet, thomas2018tensor}. 
    \begin{align}
              \mA_{ij} & =
              \begin{cases}
                  1 \text{ if } d_{ij} \leq c \\
                  0 \text{ otherwise.}
              \end{cases}
    \end{align}
    By focusing on local interactions, the cutoff graph facilitates a deeper understanding of the system's behaviour while reducing the computational overhead. 
    This approach aligns with physical and chemical constraints, explicitly enforcing locality as an inductive bias since atoms that are too far apart generally have negligible interactions. Stacking several GNN layers may enable to capture long-range dependencies. Cutoff graphs is the most widespread approach at present. It is sometimes combined with $k$ nearest neighbours techniques to ensure that nodes have the same degree, constructing a regular graph.
    \item \textbf{Via a smooth cutoff graph} to ensure a smooth energy landscape and well-behaved force predictions, particularly for molecular dynamics (MD) simulations. Using a traditional cutoff graph for MD would imply that a small change in the position of a single atom could result in a large change in energy prediction, i.e. a very steep gradient in the energy landscape. This happens because from one frame to another, an atom can move from outside the cutoff to inside the cutoff, most likely breaking simulations since forces would be unbounded. To alleviate these jumps in the regression landscape, \citet{unke2019physnet} proposed using a smoothed cutoff graph using the cosine function, where distances $d_{ij} = || \gv{x}_i - \gv{x}_j ||$ are smoothed out in the following way:
          \begin{align}
              \mA_{ij} & =
              \begin{cases}
                  \dfrac{1}{2} \big( \cos \big( \frac{\pi d_{ij}}{c}\big) +1 \big) \text{ if } d_{ij} \leq c \\
                  0 \text{ otherwise.}
              \end{cases}
          \end{align}
    Note that the adjacency matrix is no longer discrete and each edge's value is utilised inside the message passing scheme to weight atoms' contributions.
    \item \textbf{Long-range connections}. While cutoff graphs leverage locality as a useful inductive bias, this impedes learning long-range interactions such as electrostatics and van der Waals forces\appref{app:eq:energy-decomposition}. To address this drawback, in addition to short-range interactions modelled by cutoff distance, \citep{kosmala2023ewald} propose to incorporate long-range interactions using a non-local Fourier space scheme limiting interactions via a cutoff on frequency. It is particularly useful for systems with charged particles where the electrostatic interactions need to be taken into account, as well as for periodic structures containing diverse atoms.
    Alternate ways to incorporatie long-range interactions include sampling random connections weighted by a heuristically determined probability, e.g. the inverse of the distance \citep{ingraham2022illuminating} which has proven effective when used within generative models.

\end{itemize}

\begin{figure}[t!]
    \centering
    \hfill
    \begin{subfigure}[b]{0.50\linewidth}
        \centering
        \includegraphics[width=\linewidth]{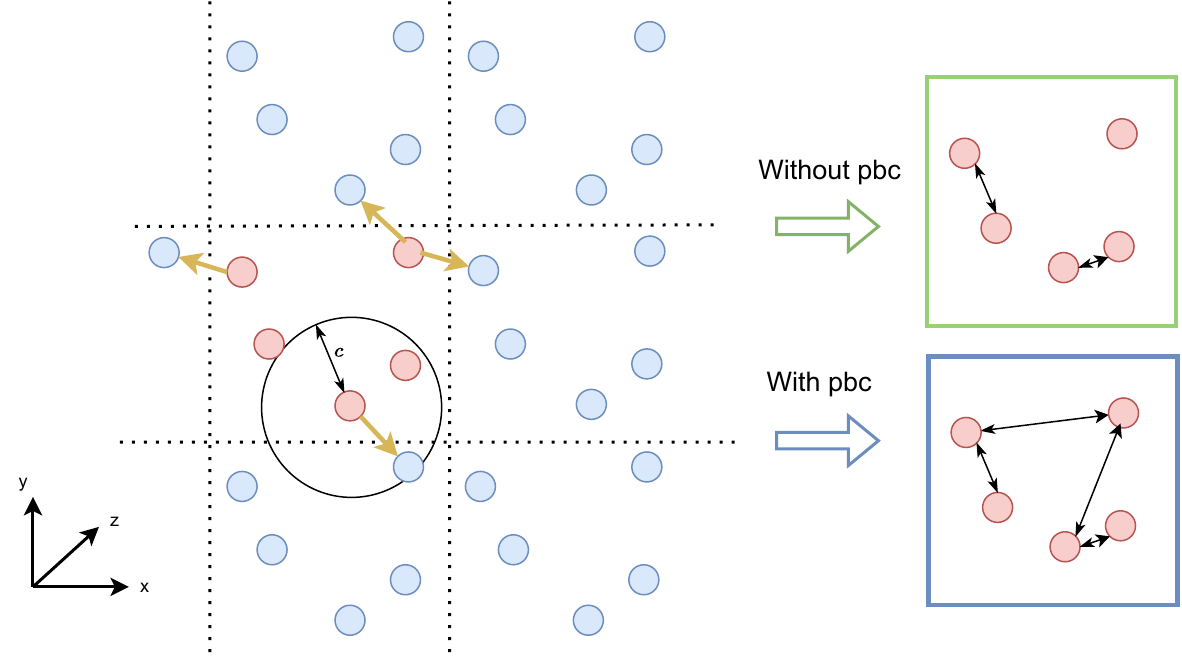}
    \end{subfigure}
    \hfill
    \begin{subfigure}[b]{0.40\linewidth}
        \centering
        \includegraphics[width=\linewidth]{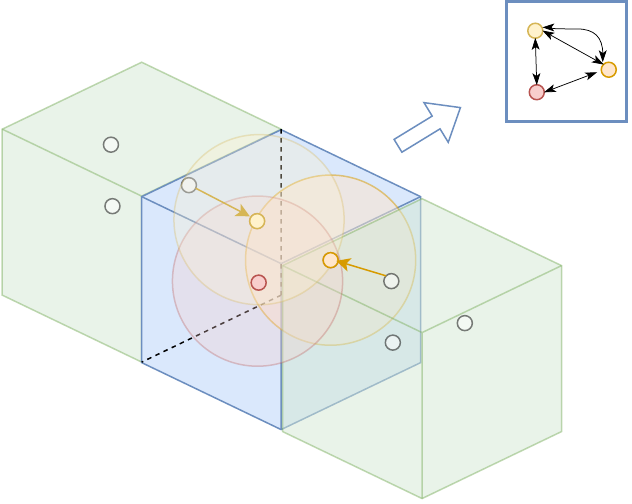} %
    \end{subfigure}
    \hfill
    \caption{\textbf{Cutoff graph with periodic boundary conditions}. Without considering the repetition of the atomic pattern, one would neglect meaningful atomic interactions. With pbc, on the other hand, we notice that the bottom-most atom and top-right atom (in a single cell, red for instance) are, in fact, close enough to be connected \textit{through} the boundary.}
    \label{fig:pbc}
\end{figure}

\begin{tcolorbox}[enhanced,attach boxed title to top left={yshift=-2mm,yshifttext=-1mm,xshift = 10mm},
colback=cyan!3!white,colframe=cyan!75!black,colbacktitle=white, coltitle=black, title=Opinion,fonttitle=\bfseries,
boxed title style={size=small,colframe=cyan!75!black} ]
While modeling atomic systems using a complete graph seems to be the most faithful representation of reality, the (smooth) local cutoff is a powerful inductive bias for modeling intermolecular interactions, which are mostly localised. In this case, long-range interactions may be captured by stacking several message passing layers. Alternatively, manually adding long-range dependencies to the cutoff graph reduces complexity (i.e. fewer layers needed) and may mitigate potential \textit{over-squashing}\appref{app:subsec:message-passing} issues. 
\end{tcolorbox}


\textbf{Periodic boundary conditions in crystals}. While molecules simply consist of a set of 3D points in space, easily representable using a finite graph, \textit{crystals}\appref{app:subsec:crystals} are modelled to be infinite periodic structures whose repeating pattern is called a \textit{unit cell}. To account for these infinite repetitions of the same substructures, we represent a single unit cell but take into consideration adjacent cells in all directions using \textit{periodic boundary conditions}\appref{app:sec:pbc} (PBC), depicted in \Cref{fig:pbc}. In short, distances under PBC are expressed as $d_{ij} = || \gv{x}_{ij} + \gv{o}_{ij} \cdot \cgv{c}||$, where $\cgv{c}$ is the 3D unit cell and $\gv{o}_{ij}$ is the cell offset parameter specifying pairwise atom proximity in neighbouring cells (i.e. a 3D vector in $\{0, 1, -1\}^3$ associated with edge $(i,j)$).
Whether periodic boundary conditions are sufficient to handle crystal systems, or if alternative approaches are worth exploring, is an active area of research \citep{kaba2022equivariant, yan2022periodic}.

\textbf{Data pre-processing and augmentation}. In addition to the graph creation, performing data pre-processing steps is sometimes useful to ensure accurate and efficient training. This includes standard techniques such as feature normalization, centering the coordinates to the origin, or target rescaling; as well as more domain-specific techniques like atom-type rescaling, defined in \Cref{app:subsec:data-preproc}. 
Finally, to improve the robustness of models, practitioners often consider multiple representations of the same sample during training, e.g. various Euclidean transformations \citep{hu2021forcenet} or corrupted versions where noise have been added to atom positions \citep{godwin2021simple}.
For biomolecules, adding random Gaussian noise may improve model robustness to small changes in atomic coordinates due to crystallography artefacts \citep{dauparas2022robust}.



\subsection{Learning representations of atoms}
\label{subsec:interaction-blocks}

Once the geometric graph is defined, the main objective is to learn meaningful atom representations. This is handled by the Embedding and Interaction blocks, which respectively initialise learnable latent representations of each atom and iteratively update them using Geometric GNN layers. 

\subsubsection{Embedding block: initialising latent representations}

The Embedding block incorporates \emph{independent} information about each atom in the geometric graph (without considering who it is connected to). 
This is typically accomplished by learning distinct embeddings $\cs{s}_z$ for each chemical element since the atomic number is always provided as part of the scalar feature matrix $\mathbf{S}$ of the geometric graph. 

Especially for quantum chemistry tasks, \citet{hu2021forcenet}  and \citet{duval2022phast} demonstrated the advantage of learning additional embeddings for the group/period of each atom ($\cs{s}_p$, $\cs{s}_g$) as well as incorporating known physical properties ($\cs{s}_f$, e.g., atomic mass, electronegativity). Such information can be easily derived from the atomic number of each atom. 
For biomolecules, practitioners typically consider only the alpha Carbon atoms to represent an entire residue for efficiency, and pre-compute geometric quantities such as displacement vectors and torsion angles to initialise node representations \citep{jamasb2023evaluating}.

The embedding block creates learnable atom representations, where the different embeddings are concatenated together to form the initial scalar representation $\mathbf{s}^{(0)}$ at layer $0$ for each atom:
\begin{equation}
\cs{s}^{(0)} = \cs{s}_z \Vert \cs{s}_p \Vert \cs{s}_g \Vert \cs{s}_f.
\end{equation}

\subsubsection{Interaction blocks: learning geometric and relational features}


\begin{figure}[t!]
  \centering
  \includegraphics[width=0.8\linewidth]{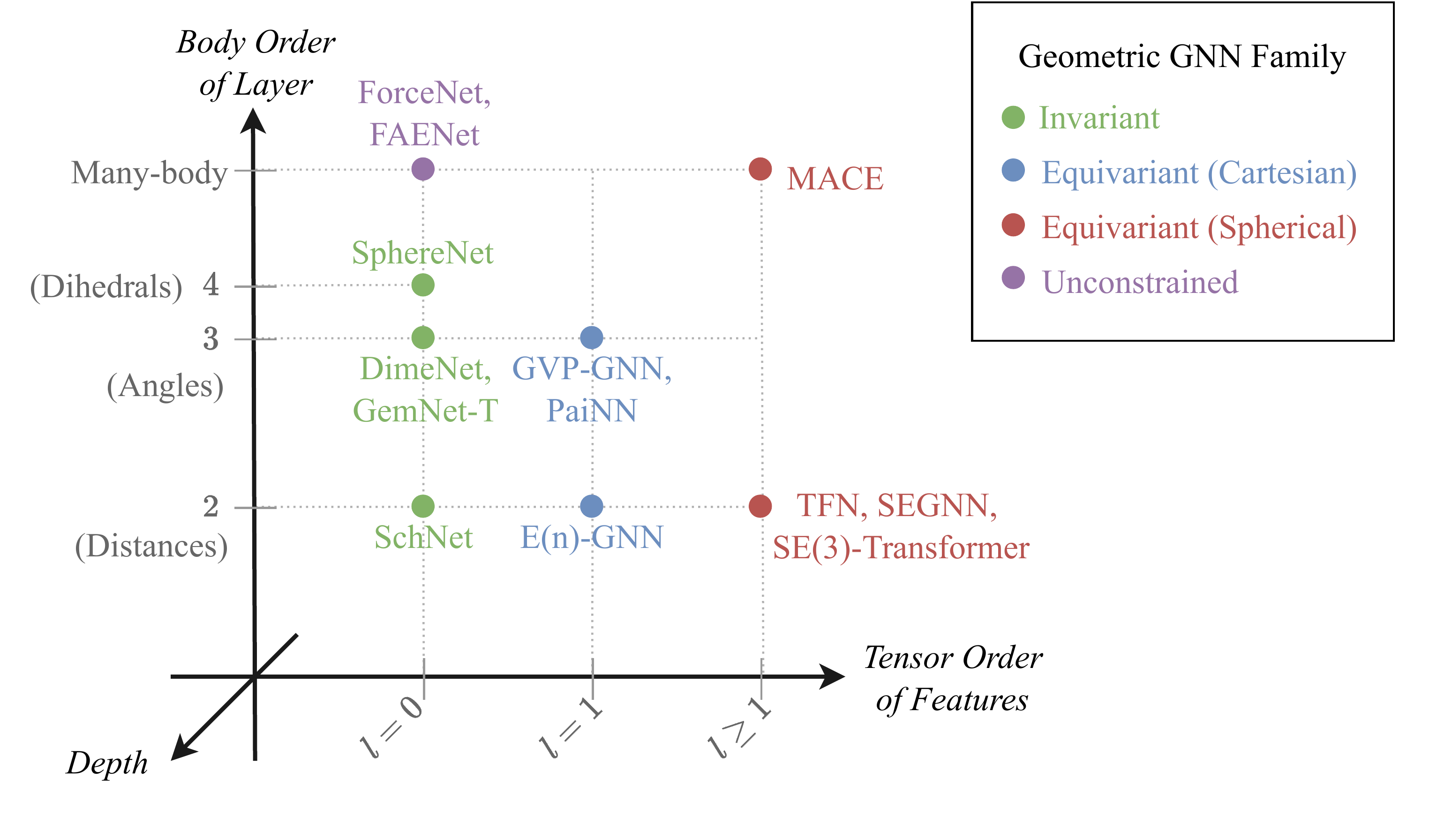}
  \caption{
  \textbf{Geometric GNN families, categorised by measures of expressive power} in terms of representing geometric (sub-)graphs: (1) \emph{Body order} refers to the number of atoms involved in constructing local neighbourhood features; (2) \emph{Tensor order} determine the granularity of capturing directional information as well as invariance or equivariance properties, and (3) \emph{Depth} controls the extent to which geometric information is propagated beyond local neighbourhoods.
  Provably enforcing symmetries acts as a constraint on the expressivity of Geometric GNNs. Unconstrained Geometric GNNs `break' the associated expressivity bottleneck of the axes by not being bound to enforce strict equivariance of intermediate features.
  }
  \label{fig:axes}
\end{figure}


Incorporating geometric and relational information about how the atoms in a system interact with one another is the critical function of Geometric GNNs. 
As previously discussed, directly treating atomic coordinates as scalar quantities in standard GNNs would violate the equivariance of the model's features under Euclidean transformations of the input system.
Therefore, the primary focus of Geometric GNNs has been to devise efficient and expressive approaches for encoding and processing geometric graphs while respecting physical symmetries. 
Most architectures explicitly incorporate these constraints into the model architecture, thereby constraining the GNN's function space to symmetry-preserving functions \citep{schutt2017schnet, satorras2021en, thomas2018tensor}. Recent approaches have also explored alternative methods for handling symmetries, such as data augmentation \citep{hu2021forcenet} or frame-based techniques \citep{duval2023faenet}. 


Interaction blocks of Geometric GNN are iteratively applied to the initial atom representations to build latent representations that capture geometric information from the local neighbourhood of each atom via message passing.
At a high level, the functioning of Interaction blocks can be broadly expressed as follows. They update scalar (and vector) features from layer $t$ to $t+1$ via learnable message and update functions, $\l{\textsc{Msg}}$ and $\l{\textsc{Upd}}$, respectively, as well as a fixed permutation-invariant operator $\bigoplus$ (e.g. mean, sum):
\begin{align}
    \label{eq:gnn-equiv}
    \vs_i^{(t+1)}, \cgv{v}_i^{(t+1)} & \defeq \l{\textsc{Upd}} \bigg( \big(\vs_i^{(t)}, \cgv{v}_i^{(t)}\big) \ , \ \underset{j \in \nei_i}{\bigoplus} \l{\textsc{Msg}} \left(  \vs_i^{(t)}, \vs_j^{(t)},  \cgv{v}_i^{(t)},  \cgv{v}_j^{(t)}, \gv{x}_{ij}\right) \bigg),
\end{align}
where $\gv{x}_{ij} = \gv{x}_{i} - \gv{x}_{j}$ denote relative position vectors. Note that \emph{$\group{G}$-invariant GNN layers} do not update vector features and only aggregate scalar quantities from local neighbourhoods. For instance, invariant GNNs may consider updates of the following form: 
\begin{align}
    \label{eq:gnn-inv}
    \vs_i^{(t+1)} & \defeq \l{\textsc{Upd}} \left( \vs_i^{(t)} \ , \ \underset{j \in \nei_i}{\bigoplus} \left( \l{\textsc{Msg}} (\vs_i^{(t)}, \vs_j^{(t)}, \ \gv{x}_{ij}) \right) \right).
\end{align}

One of the key contributions of this survey is to categorise Geometric GNN architectures into four distinct families: (1) Invariant, (2) Equivariant in Cartesian basis, (3) Equivariant in spherical basis, and (4) Unconstrained models\footnote{
Unconstrained GNNs do not strictly enforce symmetries via their architecture, but generally attempt to learn approximate symmetries via data augmentation or canonicalization.
}. We will describe each architecture family in subsequent sections, and provide a concise opinionated history of methods in \Cref{sec:history-of-methods}.



\subsection{Output block: making predictions}
\label{subsec:output-block}

Once we have obtained meaningful latent representations of each atom using the Embedding and Interaction blocks, the Output block is used to make task-specific predictions for which the model is trained. 
In general, most predictive tasks involve outputs at the node level (e.g. forces acting on each atom) or global graph level (e.g. potential energy of the entire system). 
Moreover, node-level outputs can be either invariant or equivariant to physical transformations of the system, while global outputs are generally invariant.
Therefore, Geometric GNN pipelines have different task-specific heads for each prediction level (graph, node, edge) and prediction type (invariant, equivariant). 

Node-level predictions are obtained from the final node representations from the interaction blocks by passing the individual representations through multi-layer perceptrons (MLP) for invariant tasks, and equivariant variations of MLPs \citep{jing2020learning, schutt2021equivariant} for equivariant tasks. 
When graph-level predictions are required, the node representations from the interaction blocks are mapped to graph-level predictions via a permutation-invariant readout function $f: \reals^{n \times a} \rightarrow \reals^{b}$. 
This is usually a simple sum or averaging operation ($b=a$), but weighted averaging or hierarchical pooling approaches may also be effective \citep{duval2022phast}.
Particularly in molecular dynamics tasks, it is common to concatenate all intermediate node representations when making predictions instead of considering only the final features \citep{schutt2017schnet, batatia2022design}.

\textbf{Energy conservation\appref{app:sec:energy-conservation}}. 
It is worth highlighting how atomic forces\footnote{Atom-wise 3D vectors representing the forces currently applied on each atom by the rest of the system.} are predicted by Geometric GNNs for molecular dynamics. 
There are two main approaches in the literature: 
(1) computing atomic forces as the negative gradient of the energy with respect to atom positions $\vec{F}_i = - \frac{\partial E}{\partial \gv{x}_i}$ (i.e. following the formal definition from physics), or 
(2) using a separate neural network to predict the forces directly from atom representations. 
Computing forces as the gradient of the energy guarantees energy-conserving forces, which is a highly desirable feature when using Geometric GNNs to run molecular simulations since it ensures the stability of simulations and retains the ability to reach local minima \citep{chmiela2017machine}.
However, \citet{kolluru2022open} demonstrated the significant computational burden associated with this approach.  Compared to training a separate force prediction head, energy conservation increases memory usage by 2-4$\times$ and leads to a drop in modeling performance on several datasets (particularly the large-scale OC20 and OC22 datasets \citep{chanussot2021open, tran2023open}).
Whether models should enforce energy conservation is an open research question which probably depends on the task at hand \citep{fu2023forces}.
We will expand on this discussion in \Cref{sec:discussion} where we describe promising future directions.


\clearpage


\section{Invariant Geometric GNNs}
\label{sec:invariant-gnns}


\begin{tcolorbox}[enhanced,attach boxed title to top left={yshift=-2mm,yshifttext=-1mm,xshift = 10mm},
colback=cyan!3!white,colframe=cyan!75!black,colbacktitle=white, coltitle=black, title=Key idea,fonttitle=\bfseries,
boxed title style={size=small,colframe=cyan!75!black} ]
Invariant GNNs leverage 3D geometric information by pre-computing informative scalar quantities between atoms, such as pairwise distances, triplet-wise angles, and quadruplet-wise torsion angles, and using learned latent representations of these quantities during message passing.
Since these input scalar quantities are invariant to Euclidean transformations, the intermediate representations and predictions of these models are guaranteed to be invariant.
\end{tcolorbox}

\textbf{Overview. }
Invariant GNNs aim to learn atomic representations and make predictions that are guaranteed to be invariant to 3D Euclidean transformations of the system, including the group of rotation $\text{SO}(3)$ or rotations and reflections $\text{O}(3)$---denoted interchangeably by $\group{G}$---and the translation group $\text{T}(3)$.
Enforcing translation invariance is straightforwardly done by:
(1) centring input point clouds to the origin by subtracting the centre of mass from each atom coordinate;
and 
(2) operating on relative displacements instead of raw coordinates.

One way to enforce $\group{G}$-invariance is to avoid directly processing quantities that depend on the frame of reference\appref{app:sec:geom-voc}. The reason for that is intuitive: if we let GNNs freely process any geometric quantity like a standard scalar quantity, every time our GNN is given as input a slightly transformed version of the same system, it may make a distinct prediction whereas the system's underlying properties are unchanged. 
Hence, $\group{G}$-invariant GNN layers extract and aggregate invariant scalar quantities from atomic coordinates. 
These quantities are computed by \textit{scalarising}\footnote{A term used colloquially in the community for extracting scalar (i.e. invariant) components from a combination of geometric vector or tensor quantities.} geometric quantities that are guaranteed to not change with Euclidean transformations of atomic systems. 
For instance, computing relative distances between atoms is a scalarisation of the geometric information $\cgv{x}$, and is invariant to translations, rotations and reflections.

These scalar features $\m{S}$ are updated from iteration $t$ to $t+1$ via learnable message ($\l{\textsc{Msg}}$) and update ($\l{\textsc{Upd}}$) functions as part of a standard message passing framework similar to \Cref{eq:standard-mpnn}:
\begin{align}
    \label{eq:gnn-inv-2body}
    \vs_i^{(t+1)} & \defeq \l{\textsc{Upd}} \left( \vs_i^{(t)} \ , \ \underset{j \in \nei_i}{\bigoplus} \left( \l{\textsc{Msg}} (\vs_i^{(t)}, \vs_j^{(t)}, \ \gv{x}_{ij}) \right) \right).
\end{align}

Depending on the task and implementation, the message from node $i$ to node $j$ may contain arbitrarily long dependencies through the graph. For instance, it could contain an aggregation over neighbours of $j$. 
Thus, in the more general form of \Cref{eq:gnn-inv-2body} provided below, $\l{\textsc{Msg}}$ takes $\cgv{x}$\footnote{Remember: $\m{a}$ is computed from atomic positions so $\cgv{x}$ contains the information about adjacency.} and $\m{s}^{(t)}$ as arguments.
\begin{align}
    \label{eq:gnn-inv-app}
    \vs_i^{(t+1)} & \defeq \l{\textsc{Upd}} \Big( \vs_i^{(t)} \ , \ \underset{j \in \nei_i}{\bigoplus} \l{\textsc{Msg}} \big(\m{s}^{(t)}, \ \cgv{x}, \ i, \ j\big) \Big).
\end{align}

\begin{figure}[t!]
    \centering
    \begin{subfigure}[b]{0.27\linewidth}
        \centering
        \includegraphics[width=\linewidth]{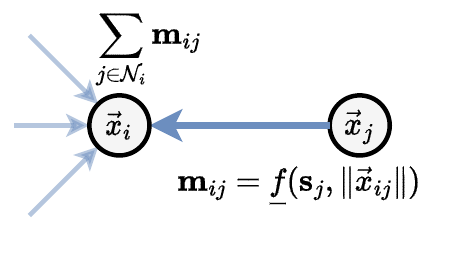}
        \caption{SchNet}
        \label{fig:schnet}
    \end{subfigure}
    \hfill
    \begin{subfigure}[b]{0.30\linewidth}
        \centering
        \includegraphics[width=\linewidth]{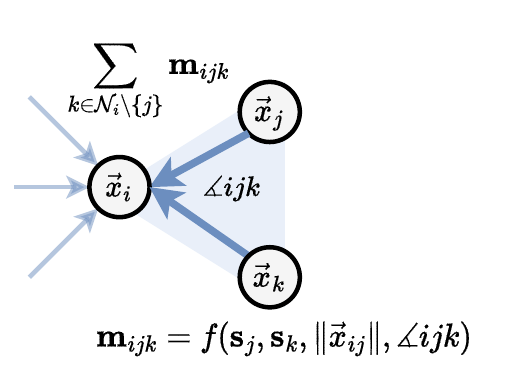}
        \caption{DimeNet}
        \label{fig:dimenet}
    \end{subfigure}
    \hfill
    \begin{subfigure}[b]{0.35\linewidth}
        \centering
        \includegraphics[width=\linewidth]{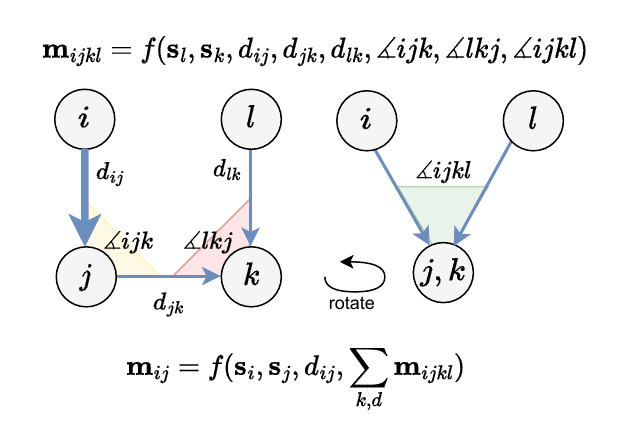}
        \caption{GemNet}
        \label{fig:gemnet}
    \end{subfigure}
    \caption{\textbf{Invariant GNN message passing. }
        $\group{G}$-invariant layers extract and propagate local scalar geometric quantities such as distances (SchNet), bond angles (DimeNet), and torsion angles (GemNet) which are guaranteed to be invariant to Euclidean transformations.
    }
    \label{fig:schnet-dimenet-gemnet}
\end{figure}

\textbf{Distance-based invariant GNNs. }
SchNet \citep{schutt2018schnet} was one of the first invariant GNN models and uses relative distances $\Vert \gv{x}_{ij} \Vert$ between pairs of nodes, encoded by a learnable Radial Basis Functions\appref{app:subsec:basis-functions} ($\l{\basis}$, i.e. an RBF with a two-layer MLP), to encode local geometric information, as shown in \Cref{fig:schnet}. 
Each SchNet layer performs a continuous convolution\appref{app:subsec:examples-archi} to combine the encoded distance information (i.e. the filter) with neighbouring atom representations (via element-wise multiplication $\odot$). This creates a message which is propagated along graph edges, enabling SchNet to effectively capture and integrate local structural features in molecular systems:
\begin{align}
    \label{eq:schnet}
    \v{s}_{i}^{(t+1)} & \defeq \v{s}_{i}^{(t)} + \sum_{j \in \nei_{i}} \lv{f}_1 \left( \v{s}_{j}^{(t)} , \Vert \gv{x}_{i j} \Vert \right) \nonumber                                     \\
                  & \defeq \v{s}_{i}^{(t)} + \sum_{j \in \nei_{i}}  \v{s}_{j}^{(t)} \odot \l{\basis}(\Vert \gv{x}_{i j} \Vert)
\end{align}
As a result, SchNet efficiently utilizes both atom-identity-related and geometric information during message passing, making it an efficient and simple-to-understand tool for processing geometric graphs.
Other architectures which pioneered the use of GNNs for 3D atomic systems also relied on distance-based invariant message passing, including CGCNN \citep{xie2018cgcnn} and PhysNet \citep{unke2019physnet}.

However, distance-based invariant GNNs are not sufficiently expressive at modeling higher-order geometric invariants. 
As SchNet relies on atom distances within a cutoff value, it cannot differentiate between atomic systems that have the same set of atoms and pairwise distances among them but differ in higher-order geometric quantities such as bond angles (refer to \Cref{app:sec:expressivity}).
This well-known limitation of low \emph{body order}\appref{app:lexicon} invariant descriptors of atomic representations is well known in the broader molecular modeling community \citep{bartok2013representing}, and continues to inform improvements to Geometric GNN architecture design \citep{pozdnyakov2022incompleteness, joshi2022expressive}.


\textbf{Going beyond distances with many-body scalars. }
To address the lack of geometric expressivity of distance-based message passing, recent invariant GNNs \citep{shuaibi2021rotation, wang2022graph, wang2022comenet, wang2023learning} focus on incorporating higher body order scalar quantities beyond pairwise interactions (e.g. from triplets and quadruplets of atoms). 
Two pedagogical examples of architectures following this trend are explained subsequently.

DimeNet \citep{klicpera2020directional}, as illustrated in \Cref{fig:dimenet}, employs continuous convolutions whose filter combines both pairwise distances $d_{ij}=\Vert \gv{x}_{ij} \Vert$ and bond angles $\measuredangle ijk = \measuredangle (\gv{x}_{ij}, \gv{x}_{ik})$
between triplets of atoms (i.e. 3-body order). 
DimeNet operates on local reference frames $(i, j \in \nei_i, k \in \nei_j \backslash \{i\})$ defined at each atom, which enables computing spatial angles between pairs of neighbours: 
\begin{align}
    \vs_i^{(t+1)} & \defeq \sum_{j \in \nei_i} \l{f_1} \Big(\vs_i^{(t)} , \ \vs_j^{(t)} , d_{ij}, \sum_{k \in \nei_j \backslash \{i\}} \l{f_2} \left( \vs_j^{(t)} , \ \vs_k^{(t)} , \ d_{ij} , \ \measuredangle ijk \right) \Big) \label{eq:dimenet}
\end{align}
The updated scalar features are $\group{G}$-invariant since geometric information is only exploited via relative distances and angles\footnote{Note that the distances and angular information are computed only once as a data pre-processing step. They are then leveraged inside each message passing layer.}, both of which remain unchanged under the action of $\group{G}$. 
Nonetheless, there exist well-known edge cases of pairs of point clouds which are the same up to distances and angles \citep{pozdnyakov2020incompleteness}.

Thus, GemNet \citep{gasteiger2021gemnet} turned to 4-body order scalarization, additionally extracting torsion angles between groups of four atoms using local reference frames, denoted $\measuredangle ijkl = \measuredangle (\gv{x}_{ij}, \gv{x}_{kd}) \perp \hat{x}_{jk}$ (see \Cref{fig:gemnet}). 
However, moving to higher body order scalarization of geometric information becomes computationally expensive. For each atom $i$, GemNet message passing must consider all direct neighbours $j \in \nei_i$, 2-hop neighbours $k \in \nei_j$ and 
 3-hop neighbours $l \in \nei_k$:
\begin{align}
    \vs_i^{(t+1)} & \defeq \sum_{j \in \nei_i} \l{f_1} \bigg( \vs_i^{(t)}, \vs_j^{(t)}, d_{ij}, \sum_{\substack{{k \in \nei_j \setminus \{i\}}, \\ l \in \nei_k \setminus \{i,j\}}} \l{f_2}(\vs_{k}, \vs_{l}, d_{kl}, d_{ij}, d_{jk}, 
    \measuredangle ijk, \measuredangle jkl, \measuredangle ijkl \bigg).
\end{align}

To improve scalability, practical GemNet variants such as GemNet-OC \citep{gasteiger2022gemnet} are often restricted to 3-body scalars. 
SphereNet~\citep{liu2022spherical} and ComENet \citep{wang2022comenet} introduced efficient method for extracting 4-body angles within local neighbourhoods, avoiding the need to loop through all 3-hop neighbours. However, as noted in the SphereNet paper, this localised approach has known failure cases where local scalars up to 4-body angles are the same across two geometric graphs, but the systems differ in terms of non-local, higher-order scalars such as dihedral angles.
Thus, the precise body order of scalars at which all geometric graphs can be uniquely identified remains an open question \citep{joshi2022expressive}.

\begin{tcolorbox}[enhanced,attach boxed title to top left={yshift=-2mm,yshifttext=-1mm,xshift = 10mm},
colback=cyan!3!white,colframe=cyan!75!black,colbacktitle=white, coltitle=black, title=Opinion,fonttitle=\bfseries,
boxed title style={size=small,colframe=cyan!75!black} ]

The GemNet paper includes a theoretical section, in which it is stated that GNNs with directed edge embeddings and two-hop message passing can universally approximate predictions that are invariant to translation, and equivariant to permutation and rotation. This statement needs careful reading. It is important to note that the universality claim requires conditions like an infinite cut-off (i.e. a fully connected graph) and appropriate discretization, as it builds upon a previous proof by \citet{dym2020universality} which showed that Tensor Field Networks are universal when operating on full graphs and using infinite tensor rank for equivariant features. As highlighted in the paper, the choice of discretization scheme can affect the universality of the approximation, and depending on the discretization scheme the resulting mesh might not provide a universal approximation guarantee. How to relax these two requirements and construct sufficient geometric conditions for universality is still an open research question and emphasized in Section 5.9 in \citep{gasteiger2023convergence}. \\
In particular, this means that while the theoretical model in the GemNet paper can be universal, the practical final architecture is not. The 4-body message passing in GemNet-Q sacrifices universality guarantees by operating on a discretization of representations in the directions of each atom's neighbours. 
Additionally, GemNet-T, the more efficient version of GemNet, performs 3-body message passing similar to DimeNet on radial cutoff graphs, which is not universal due to known counterexamples \citep{pozdnyakov2020incompleteness}. 
The fact that the universality proof does not necessarily carry over to the final GemNet architecture was also emphasised by the authors of GemNet in \href{https://twitter.com/gasteigerjo/status/1469692491960102913}{this thread}. \\
In summary, in the GemNet paper it is key to distinguish between the \emph{theoretical model}, which can be universal, and the \emph{final architecture}, which is not. We highlight this point here to avoid a misconception in the community that invariant architectures operating on distances, angles, and torsions angles are guaranteed to be universal or complete. Developing a universal geometric GNN in the general case, for sparse graphs and using finite tensor rank, remains an open question which we discuss in \Cref{sec:discussion}.

\end{tcolorbox}

\textbf{Canonical frame-based invariant GNNs. }
Canonical frame-based GNNs \citep{liu2022spherical, wang2022comenet} use a local or global frame of reference to scalarise geometric quantities into invariant features which are used for message passing, offering an alternative technique when canonical reference frames can be defined. 
Most notably, the Invariant Point Attention layer (IPA) from AlphaFold2 \citep{jumper2021highly} defines canonical local reference frames at each residue in the protein backbone centred at the alpha Carbon atom and using the Nitrogen and adjacent Carbon atoms.
Other invariant GNNs for protein structure modelling also process similar local reference frames \citep{ingraham2019generative, wang2023learning}.
IPA is an invariant message passing layer operating on an all-to-all graph of protein residues.
In each IPA layer, each node creates a geometric feature (position) in its local reference frame via a learnable linear transformation of its invariant features.
To aggregate features from neighbours, neighbouring nodes' positions are first rotated into a global reference frame where they can be composed with their invariant features (via an invariant attention mechanism), followed by rotating the aggregated features back into local reference frames at each node and projecting back to update the invariant features.

\textbf{Summary. }
Overall, invariant GNNs' reliance on a precomputed procedure to scalarise geometric information is both a blessing and a curse.
Models such as SchNet or DimeNet can be very efficient baselines for modeling 3D atomic systems.
However, improving the expressivity of invariant GNNs results in increasingly complex architectures as incorporating higher body order invariants necessitates expensive accounting of higher-order tuples \citep{li2023distance}. 


\clearpage


\section{Equivariant Geometric GNNs}
\label{ssec:equivariant_gnns}


\begin{figure}[h!]
    \centering
    \includegraphics[width=0.6\linewidth]{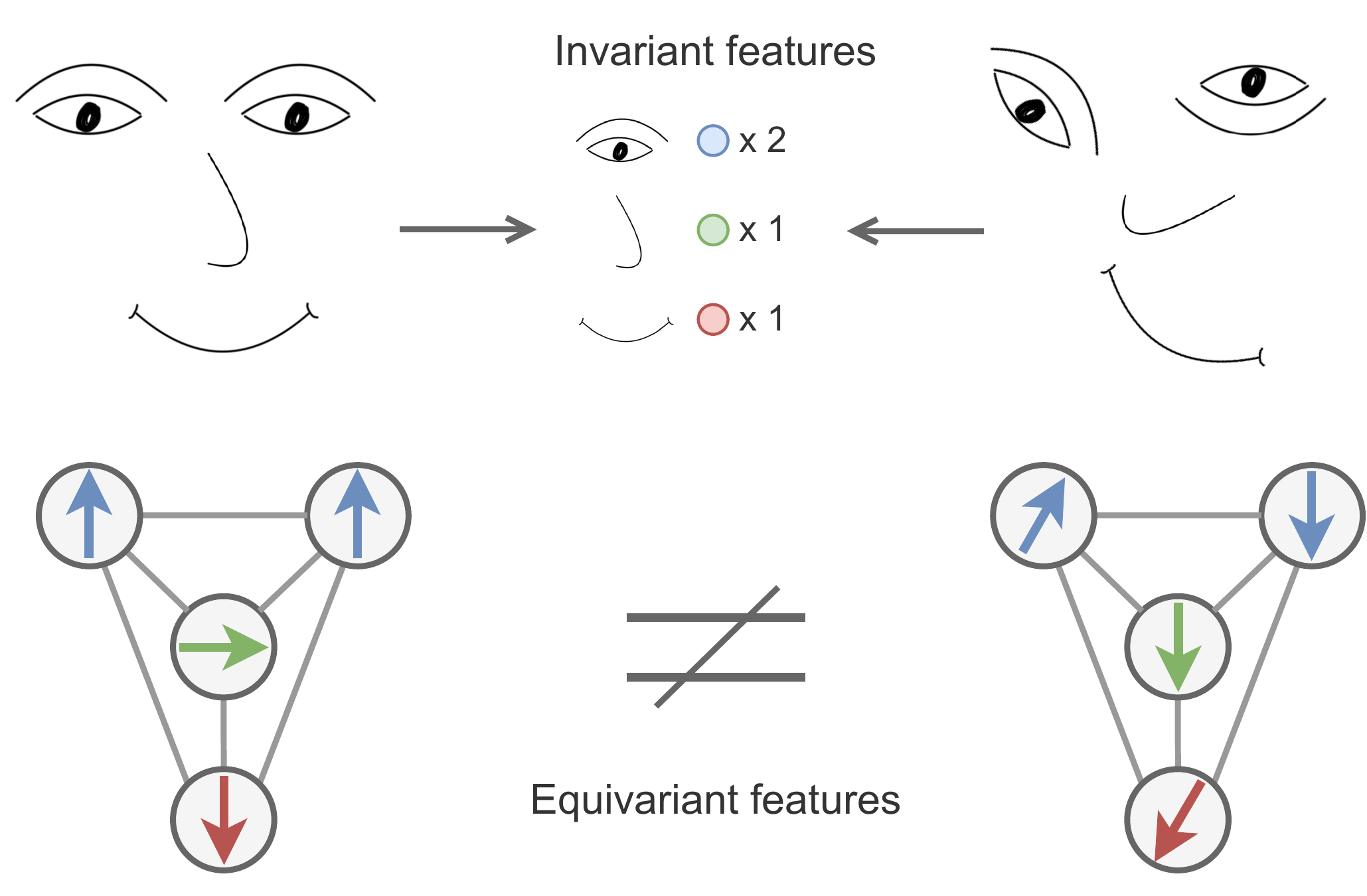}
    \caption{\textbf{The \emph{Picasso Problem} and the need for equivariant representations.} Identifying a face composed of a set of eyes, a nose, and a mouth requires understanding the relative orientation of the parts with respect to one another (equivariant information), and not just detecting their presence (invariant features) \citep{hinton2021represent}.
    For 3D atomic systems, making invariant predictions such as functional properties of molecules may still necessitate solving equivariant sub-tasks, such as composing how sub-graphs and motifs interact with one-another geometrically. 
    Intuitively, equivariant representations in geometric GNNs allow the network to \emph{learn} a set of invariants beyond local neighbourhoods.
    }
    \label{fig:picasso}
\end{figure}

\textbf{Overview. }
As explained in the previous section, invariant GNNs pre-compute a set of \emph{local invariants} in each neighbourhood before performing message passing. While this can be efficient, it is also restrictive. A key limitation of invariant GNNs is that the set of local invariants is fixed and has to be determined prior to message passing.

How could we overcome this limitation and instead allow the network to \emph{learn} its set of invariants which could (1) be more suited to the task at hand, and (2) whose complexity could be controlled by performing more or fewer message passing steps?

In this section, we investigate a family of GNNs, which we will refer to as \emph{equivariant GNNs} (EGNNs), that fulfil these two goals and additionally allow the prediction of equivariant quantities. These models do not pre-compute local invariants but instead perform message passing in a way such that the hidden features at each layer are equivariant to symmetry transformations of the input. 
In other words, if we were to, say, rotate the input graph, the hidden features at each layer would also be rotated correspondingly. 
This is in contrast to invariant GNNs, where the hidden features at each layer would remain the same. We will see that by using equivariant GNN layers, the network can build up its own set of invariants \emph{on the fly}, as it performs message passing, and more message-passing layers can lead to more complex invariants which contain information from multiple, possibly non-local neighbourhoods.

The crux to make equivariant GNNs work is to perform \emph{diligent accounting} of how each hidden feature in each layer has to transform in order to remain equivariant. This accounting amounts to the following basic intuitions\footnote{These intuitions are formalised in the well-studied mathematical sub-field of representation theory. There are many great treatises on representation theory available (e.g. \citep{zee2016group}, chapter 2). Instead of repeating them and giving you the theory ``top-down'', we follow a different route and try to help you build an intuition as to why this is relevant in machine learning and how you would arrive at some basic tenants of representation theory from the ``bottom-up''. For those interested in exploring formal mathematics, we will provide the official mathematical terms for key concepts we discuss in footnotes so that you may look them up in standard references in your own time. For all mathematicians, we note that we will assume all representations we are working with are orthogonal.}:

\begin{enumerate}
    \item \textbf{Types:} Each feature is associated with a \emph{type} which tells us how that feature changes under symmetry transformations\footnote{Mathematically, the \emph{type} of a feature tells us which linear representation of the symmetry group the feature vector transforms in. The possible types depend on the symmetry group, e.g. rotations in 3D, and classifying all possible types of a given group is one of the main tasks in representation theory. Luckily for us, all types for rotations in 3D are well known, but this is not the case for some other groups.}.
    \item \textbf{Addition of types:} When adding features through component-wise addition of lists of numbers that represent them, we need to make sure that we only add features of the same type. This ensures that the sum of the features has the same type.
    \item \textbf{Multiplication of types}: When we multiply features with each other, we can multiply features of \emph{different types}, but we need to keep track of how the product transforms. As we shall see, the complication is that the naive component-wise product of two lists of numbers which represent two features, each of which transforms equivariantly individually, will generally transform differently than each of its factors\footnote{For instance, $ \rot (\cgv{u} \odot \cgv{v}) \neq \rot \cgv{u} \odot \rot\cgv{v} $ in general.}. To take this into account, we cannot just perform component-wise multiplication of equivariant feature types. Instead, we use a special product called the \emph{tensor product}\appref{app:sec:geom-voc}, which, for our purposes, can loosely be seen as a generalization of multiplication that takes into account how the product transforms.
    \item \textbf{Non-linear operations on types:} Non-linear operations need to be handled with care. Why is this the case?
    For the relevant intuition, you may think of non-linear operations in terms of their Taylor expansion, which incorporates products of features. If higher-order products of features yield different types (as described in the previous point (3) on the multiplication rule), we cannot add them together by the addition rule in point (2). Importantly, the nonlinearities in machine learning are usually understood to be applied component-wise. But this often breaks equivariance for all but scalar representations. For this reason, nonlinearities are most often applied to scalar-type features only.
\end{enumerate}

How then should we define a \emph{type}? And which multiplication rules between these types do we then need to follow to ensure the multiplication rule (3)?

There are many ways of choosing types (which in turn determine the multiplication rules). Below, we discuss two of them: (1) using Cartesian coordinates and (2) using spherical coordinates\footnote{This corresponds to an example of dealing with reducible and irreducible representations respectively. These terms will become clearer when we talk about the relations between Cartesian and spherical tensors.}. We will start with the Cartesian formulation, which we expect to be more intuitive for most readers and then transition to the spherical formulation which has a natural relationship to rotations and reflections in 3D.

\subsection{Equivariant GNNs with Cartesian tensors}
\label{sssec:cartesian_equivariant_gnns}

\begin{tcolorbox}[enhanced,attach boxed title to top left={yshift=-2mm,yshifttext=-1mm,xshift = 10mm},
colback=cyan!3!white,colframe=cyan!75!black,colbacktitle=white, coltitle=black, title=Key idea,fonttitle=\bfseries,
boxed title style={size=small,colframe=cyan!75!black} ]
Cartesian EGNNs model atomic interactions in Cartesian coordinates and restrict the set of possible operations on geometric features to preserve equivariance. They often update (and combine) both scalar and vector messages in parallel. 
\end{tcolorbox}

\paragraph{Scalar-Vector GNNs.} Let us start simple and consider two familiar types of features only: \emph{scalars} and \emph{vectors}. An example of a scalar is the distance between two nodes or the angle between two vectors: it is an object that does not change under transformations in the symmetry group \group{G}, in our case under rotations, reflections and translations. A vector (e.g. atomic forces), in contrast, transforms under these operations in the familiar way: $\gv{v} \mapsto \rot \gv{v} + \trans$ where $\rot$ is a rotation or reflection matrix and $\trans$ is a translation vector. In the didactic discussion below, we will ignore translations and reflections and focus on the group of rotations only. Reflections and translations do not require novel insights and can be dealt with fairly easily once we understand how to deal with rotations\footnote{Global translations can be dealt with easily because we are normally only interested in invariant features with respect to translations. This can be achieved for example through zero-centring the point cloud by subtracting the centre of mass, or by working exclusively with displacement vectors between nodes. In both cases, a global translation $\gv{t}$ drops out in the subtraction $\gv{v}_1 - \gv{v_2} \mapsto (\gv{v}_1 + \gv{t}) - (\gv{v}_2 + \gv{t}) = \gv{v}_1 - \gv{v}_2$. We will speak more about reflections in \Cref{sssec:equivariant_gnns_spherical_tensors}.
}. 

What multiplication rules make sense, given we have these two types of features? Let us think about multiplication operations between scalars and vectors that we are already familiar with. We know that the product of two scalar-types will give us a scalar-type again and that the product of a scalar-type and a vector-type will simply give us a scaled vector-type\footnote{While the notations ``scalar-type'' and ``vector-type'' may seem like an unnecessary burden at this point, we want to emphasize that thinking constantly about the types of the objects being manipulated is paramount in this section.}. Finally, we can `multiply' two vector-types via the dot-product to get a scalar-type\footnote{You might rightly wonder about the cross-product here. Wouldn't that give us a valid type too? In fact, the cross-product of two vectors would yield a pseudo-vector, which transforms differently under point reflections at the origin than a vector would and is, therefore, a different type. A vector would flip sign under point reflections while the cross product's sign remains unchanged (try it for yourself!). Hence, it is not in our multiplication table. We will return to products such as the cross-product slightly later and see that it effectively corresponds to a special case of the tensor product.}. Notice that taking the norm $\vert\vert \gv{v} \vert \vert$ of a vector is just a special case of the dot-product: it amounts to taking the dot-product of a vector with itself (followed by a square root).


So, given scalar and vector feature types, as well as the three multiplication rules above, which GNNs can we build?
The messages and update rules would have to be of the form:
\begin{align}
    \label{eq:gnn-equiv-app}
    \cs{m}_{ij}^{(t)}, \cgv{m}_{ij}^{(t)}     & \defeq \l{\textsc{Msg}} \big(\vs_i^{(t)}, \c{s}_j^{(t)}, \cgv{v}_i^{(t)}, \cgv{v}_j^{(t)}, \gv{x}_{ij}\big) & \text{(Message)} \\
    \vs_i^{(t+1)}, \cgv{v}_i^{(t+1)} & \defeq \l{\textsc{Upd}} \bigg( (\vs_i^{(t)}, \cgv{v}_i^{(t)}) \ , \ \underset{j \in \nei_i}{\bigoplus}(\vm_{ij}^{(t)}, \cgv{m}_{ij}^{(t)}) \bigg)                                                         & \text{(Update)}
\end{align}
with a scalar message $\vm_{ij}^{(t)}$ and a vector message $\cgv{m}_{ij}^{(t)}$. The most general messages with these operations could then be constructed by the following operations (suppressing the superscript $(t)$ for readability and letting $\v{m}_i$, $\cgv{m}_i$ denote the aggregated message for node $i$ from $\nei_i$, right before the actual update):
\begin{align}
    \label{eq:general-message-s}
    \cs{m}_i       & \defeq &
        \l{f_1}(
            \cs{s}_i,
            \norm{\cgv{v_i}}  
        )
        & + \sum_{j \in \nei_i} \l{f_2} \left(
            \cs{s}_i,
            \cs{s}_j ,
            \norm{ \gv{x}_{ij} },
            \norm{ \vec{\vv_j} },
            \gv{x}_{ij} \cdot \cgv{v}_j,
            \gv{x}_{ij} \cdot \cgv{v}_i,
            \cgv{v}_i \cdot \cgv{v}_j
    \right)                                                                                                                                      \\
    \label{eq:general-message-v}
    \cgv{m}_i       & \defeq &
        \l{f_3}(
            \cs{s}_i,
            \norm{\cgv{v_i}}  
        ) \odot \cgv{v}_i 
        & +  \sum_{j \in \nei_i} 
        \l{f_4} \left( 
            \cs{s}_i,
            \cs{s}_j ,
            \norm{ \gv{x}_{ij} },
            \norm{ \vec{\vv_j} },
            \gv{x}_{ij} \cdot \cgv{v}_j,
            \gv{x}_{ij} \cdot \cgv{v}_i,
            \cgv{v}_i \cdot \cgv{v}_j
        \right) \odot \cgv{v}_j \nonumber
        \\ & & &
        + \sum_{j \in \nei_i} 
        \l{f_5} \left( 
            \cs{s}_i,
            \cs{s}_j ,
            \norm{ \gv{x}_{ij} },
            \norm{ \vec{\vv_j} },
            \gv{x}_{ij} \cdot \cgv{v}_j,
            \gv{x}_{ij} \cdot \cgv{v}_i,
            \cgv{v}_i \cdot \cgv{v}_j
        \right) \odot \gv{x}_{ij},
\end{align}
where $\l{f_1}$ to $\l{f_5}$ are learnable, possibly non-linear, functions and $\odot$ denotes element-wise multiplication\footnote{If it is unclear to you why this is the most general form, think of it this way: first, as stated above, component-wise non-linearities ($\l{f_i}$) can only be applied to scalars so they can only work from all possible scalar quantities at our disposal, namely scalar features and dot products; second we decompose self-interactions and neighbourhood interactions; lastly in the case of $\cgv{m}_i$ we compose those non-linearities with all the \textit{vector} features at our disposal. Note that we do indeed maintain equivariance of vector features because of the distributivity of the matrix-vector product: \rot \gv{v} + \rot \gv{u} = \rot (\gv{u} + \gv{v}).}.
Special cases of the ``most general'' equations above give rise to a whole host of published architectures \citep{jing2020learning, schutt2021equivariant, satorras2021n, tholke2022torchmd, du2022se, le2022representation, morehead2022geometry}. 

\begin{figure}[t!]
    \centering
    \begin{subfigure}[b]{0.45\linewidth}
        \centering
        \includegraphics[width=0.7\linewidth]{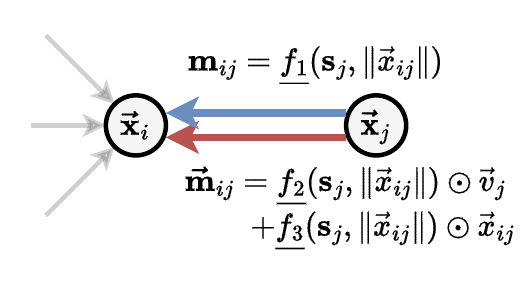}
        \caption{PaiNN}
        \label{fig:painn}
    \end{subfigure}
    \begin{subfigure}[b]{0.45\linewidth}
        \centering
        \includegraphics[width=0.7\linewidth]{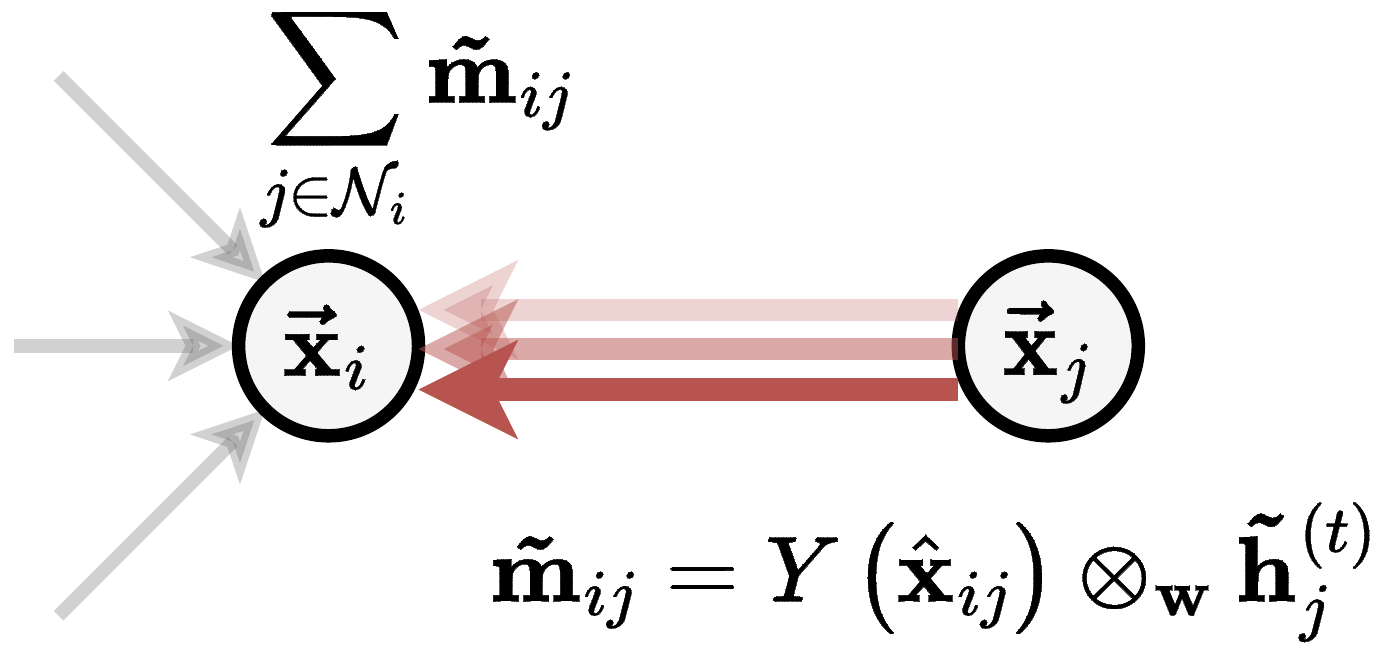}
        \caption{TFN}
        \label{fig:tfn}
    \end{subfigure}
    \caption{\textbf{Equivariant message passing. }
        $\group{G}$-equivariant layers such as PaiNN and TFN propagated geometric quantities such as vectors, relative positions, or tensors.
    }
    \label{fig:painn-tfn}
\end{figure}

For example, in PaiNN \citep{schutt2021equivariantmp} interaction layers aggregate scalar and vector features via learnt filters conditioned on the relative distance, as shown in \Cref{fig:painn}:
\begin{align}
    \label{eq:painn-s}
    \vm_i^{(t)}       & \defeq \vs_i^{(t)} + \sum_{j \in \nei_i} \l{f_1} \left( \vs_j^{(t)} , \ \Vert \gv{x}_{ij} \Vert \right)                                                                                                                                               \\
    \label{eq:painn-v}
    \cgv{m}_i^{(t)} & \defeq \cgv{v}_i^{(t)} + \sum_{j \in \nei_i} \l{f_2} \left( \vs_j^{(t)} , \ \Vert \gv{x}_{ij} \Vert \right) \odot \cgv{v}_j^{(t)} + \sum_{j \in \nei_i} \l{f_3} \left( \vs_j^{(t)} , \ \Vert \gv{x}_{ij} \Vert \right) \odot \gv{x}_{ij}.
\end{align}


TorchMD-Net \citep{tholke2022torchmd}, an equivariant transformer-based GNN, extends the above message passing layer to attention by choosing the $\l{f}$'s appropriately. E-GNN \citep{satorras2021n} and GVP-GNN \citep{jing2020learning} also fall within this paradigm. The update step applies a gated non-linearity \citep{weiler20183d} on the vector features, which learns to scale their magnitude using their norm concatenated with the scalar features:
\begin{align}
    \label{eq:painn-u}
    \vs_i^{(t+1)} \defeq \vm_i^{(t)} + \l{f}_4 \left( \vm_i^{(t)}, \Vert \cgv{m}_i^{(t)} \Vert \right) , \quad\quad
    \cgv{v}_i^{(t+1)} \defeq \cgv{m}_i^{(t)} + \l{f}_5 \left( \vm_i^{(t)}, \Vert \cgv{m}_i^{(t)} \Vert \right) \odot \cgv{m}_i^{(t)} .
\end{align}
The updated scalar features are both $\group{G}$-invariant and $T(d)$-invariant because the only geometric information used is the relative distances, while the updated vector features are $\group{G}$-equivariant and $T(d)$-invariant as they aggregate $\group{G}$-equivariant, $T(d)$-invariant vector quantities from the neighbours.

These \emph{scalar-vector} GNNs achieve good performance and are relatively fast by avoiding expensive operations. Obtaining them required us to manually define and exploit the multiplication rules between scalars and vectors that we already knew. But are these all the possible multiplication-like operations that exist in geometric operations? As we shall see, these 
scalar-vector GNNs are but special cases of a much broader possible design space\footnote{What we call scalar-vector-based models here are sometimes referred to as low tensor rank models elsewhere, for reasons that will become apparent shortly.}.

\paragraph{Higher tensor-types and the tensor product.} 
We just saw that we can understand many of the published equivariant GNNs as special cases of scalar-vector type GNNs in which we restrict ourselves to only two types of features: scalars and vectors. Why stop there? We could also create other types of features which transform differently. For example, if we have two (possibly identical) vectors $\gv{v}$ and $\gv{w}$, we could create a matrix from them by assigning the components $\cm{M}_{ij} = \gv{v}_i \gv{w}_j$. Under a global rotation $\rot$, this matrix transforms as $\m{M} \mapsto \rot \m{M} \rot^\top$, or in component notation\footnote{It is a good exercise to verify this by hand once.}:
$$\cm{M}_{ij} \mapsto  \gs{m} \gs{n} \rot_{im} \rot_{jn} \m{M}_{mn}.$$ 
 Importantly, if we construct two matrices this way, any sum or indeed linear combination of so constructed matrices will continue to transform as stated above. Further, as we have just seen, this matrix-like transformation is a different transformation than that of the vector types $\gv{v}$ or $\gv{w}$, which each transforms as $\gv{v} \mapsto \rot \gv{v}$. We have therefore discovered a \emph{new type}\footnote{We use the word \emph{type} here as we imagine our readers to mostly be computer scientists to whom this term will be more familiar. Mathematically, our types correspond to different representations of the relevant group, i.e. $O(3)$ or $SO(3)$ here.}! 

 \begin{figure}
     \centering
     \includegraphics[width=0.9\textwidth]{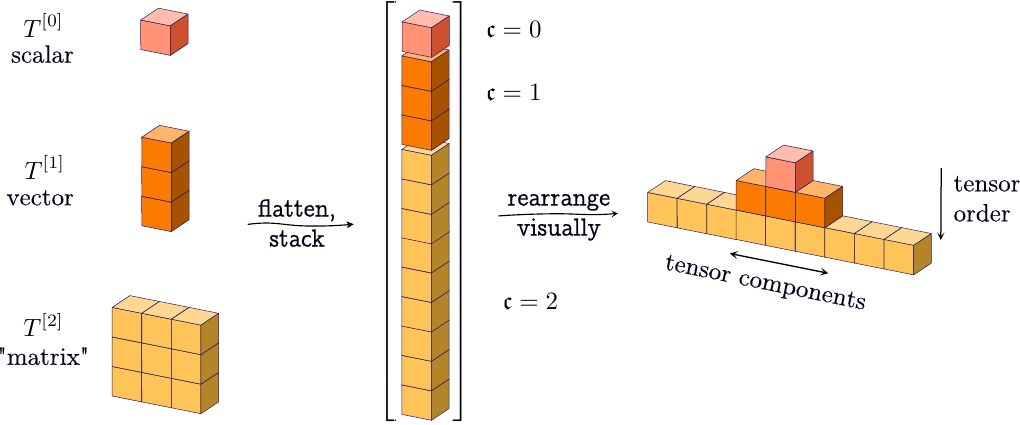}
     \caption{Cartesian Tensors, represented visually. The left column shows a few familiar objects that behave as Cartesian tensors do under rotations: scalars, vectors, matrices. When we flatten them (central column), it becomes apparent that higher-order Cartesian tensors can be re-interpreted as vectors in a higher-dimensional linear space. A matrix, for example, can be flattened into a vector in a 9-dimensional space. On the right, we group the tensors together into a pyramid to visually disentangle the tensor component and tensor order axis.}
     \label{fig:cartesian_tensors_intro}
 \end{figure}
 
 In the same way, we created the ``matrix'' type above, we can also create objects of yet other types, which transform differently than matrices. For example, the object comprised of the components\footnote{You may think of $\mT$ as a $3 \times 3 \times 3$ cube.} 
 $\mT_{ijk} = \gv{v}_i \gv{v}_j \gv{w}_k$ would transform as 
 $$\mT_{ijk} = \gs{l}\gs{m}\gs{n} \rot_{il} \rot_{jm} \rot_{kn} \mT_{lmn}.$$ 
 By its definition, this object transforms differently than a vector-type $\gv{v}$ or matrix-type $\m{M}$. It transforms with the help of 3 copies of a rotation matrix and has $3^3=27$ components.

\begin{figure}
     \centering
     \includegraphics[width=0.9\textwidth]{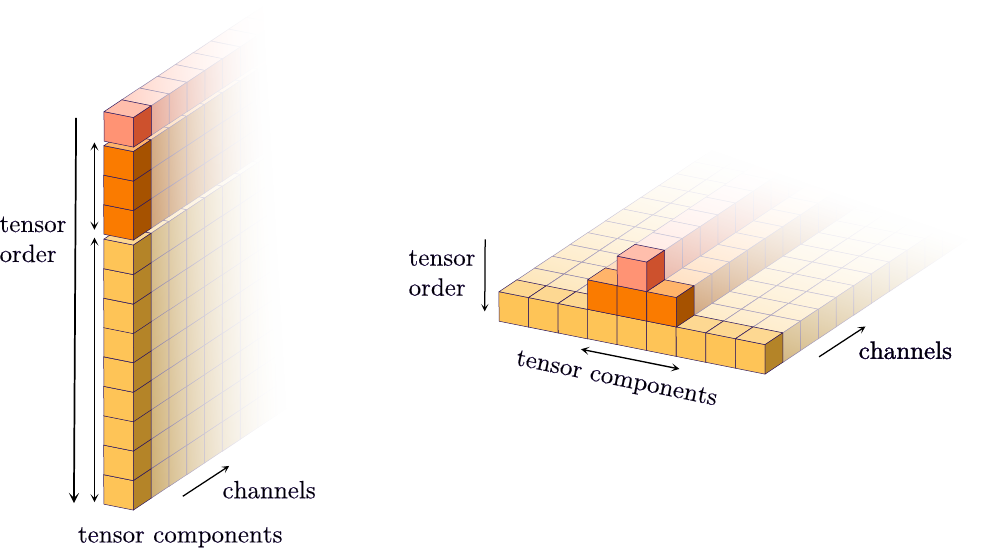}
     \caption{For simplicity we omit the channel axis in most of our illustrations and the channel index in our equations. In all contemporary machine learning models, you would carry an implicit channel index. The channel axis has no bearing on the transformation behaviour, which is why we omit it without loss of generality.}
     \label{fig:cartesian_tensors_with_channels}
 \end{figure}

We shall call \emph{Cartesian tensors} $\gt{T}^\ucart$ the new types that we can build by multiplying each combination of components of $\cart$ vectors, illustrated in \Cref{fig:cartesian_tensors_intro}. We assign $\cart$ indices to enumerate them and call this number the rank of the tensor. Channel indices are ignored without loss of generality, as shown in \Cref{fig:cartesian_tensors_with_channels}. The matrix $\cm{M}_{ij}$ that we constructed above is therefore a rank-2 type Cartesian tensor $\gt{M}^{[2]}$ and the object $\mT_{ijk}$ is a rank-3 type Cartesian tensor $\gt{T}^{[3]}$. A vector $\gv{v} = \gt{V}^{[1]}$ could be considered a rank-1 type Cartesian tensor, although for consistency and ease of notation, we will continue to refer to vectors via the simpler notation $\gv{v}$. 

Mathematically, the above way of creating new types with more indices from existing ones is called taking the \emph{tensor product}\footnote{The tensor product can also be defined abstractly as a map from two input linear spaces to a third one (the tensor product space) that satisfies what is called the \emph{universal property}, which loosely says that for every bilinear map on the original two linear spaces there is a unique, corresponding linear map in the third linear space. It turns out this construction is unique up to a (unique) isomorphism and its practical instantiation gives the tensor product we ``re-discovered'' above.} (denoted as $\otimes$) of types we had previously, which is why we called these new types tensors. As we just saw, it gives us a way to build tensor types of higher and higher rank, i.e. more and more indices. We can naturally apply the tensor product not just to vectors but any other Cartesian tensors\footnote{Mathematically, these are simply vectors in a higher-dimensional linear space that is the tensor product of two lower dimensional linear spaces. The reason we call them tensors is mostly because we keep multiple indices for them around, because this makes it easy to address how they transform via standard rotation matrices, but there are many different but equivalent ``viewpoints'' you can take, illustrated in \Cref{fig:cartesian_tensors_intro}. } as well. In the new notation:
\begin{align}
    \gt{M}^{[2]} &= \gv{v} \otimes \gv{w} 
    &\text{component wise:\hspace{0.1cm}}& \gt{M}^{[2]}_{ij} = \gv{v}_i \gv{w}_j \\
    \gt{T}^{[3]} &= \gv{u} \otimes \gv{v} \otimes \gv{w} 
    &\text{component wise:\hspace{0.1cm}}& \gt{T}^{[3]}_{ijk} = \gv{u}_i \gv{v}_j \gv{w}_k \\
    \gt{U}^{[5]} &= \gv{x} \otimes \gv{y} \otimes \gt{T}^{[3]} 
    &\text{component wise:\hspace{0.1cm}}& \gt{U}^{[5]}_{ijklm} = \gv{x}_i \gv{y}_j \gt{T}^{[3]}_{klm} 
\end{align}

A Cartesian tensor of a given rank $\cart$ is a well-defined type according to the type intuition (1), because it consistently transforms in a predictable way under rotations and reflections, and linear combinations of tensors of the same type also yield a tensor of that type\footnote{Mathematically, these two properties mean that the tensor product linear space satisfies the definition a linear representation of the rotation and reflection group $O(3)$.}. 

To write down these transformation rules and deal with tensors more generally, it is convenient to use a notational convention called \emph{Einstein summation}. In \emph{Einstein summation} notation, repeated indices are understood to be summed over and we therefore drop the explicit summation symbol. Let's see a few examples and write down the ways the Cartesian tensors we constructed above transform:
\begin{align}
    &\text{Full notation:} & \text{Einstein summation notation:}& \\
    &\gs{i} \gv{v}_i \gv{w}_i & \gv{v}_i \gv{w}_i& \\
    &\gs{m} \gs{n} \rot_{im} \rot_{jn} \gt{M}^{[2]}_{mn} & \rot_{im} \rot_{jn} \gt{M}^{[2]}_{mn}& \\
    & \gs{l}\gs{m}\gs{n} \rot_{il} \rot_{jm} \rot_{kn} \gt{T}^{[3]}_{lmn} & \rot_{il} \rot_{jm} \rot_{kn} \gt{T}^{[3]}_{lmn} &\\
    & \gs{i}\gs{j}\gs{k}\gs{l}\gs{m} \rot_{ri} \rot_{sj} \rot_{tk} \rot_{ul} \rot_{vm} \gt{U}^{[5]}_{ijklm} &  \rot_{ri} \rot_{sj} \rot_{tk} \rot_{ul} \rot_{vm} \gt{U}^{[5]}_{ijklm}
\end{align}
As we can see, Einstein summation helps express the transformation rules more concisely. Unless otherwise specified and until the end of this section, \textit{we will assume Einstein summation}.

\begin{figure}
    \centering
    \includegraphics[width=0.5 \linewidth]{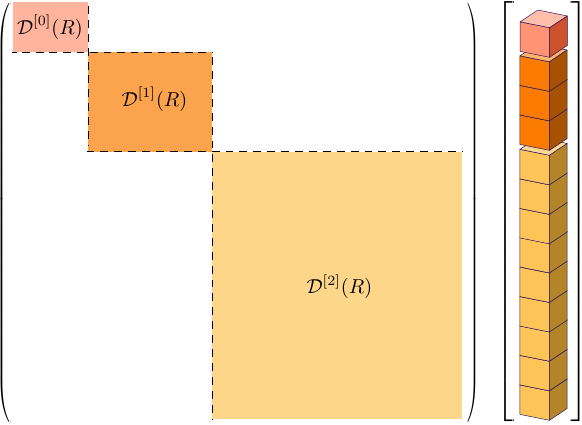}
    \caption{Illustration of the transformation of a compound tensor composed of $\mathfrak{c}=0,1,2$ components when seen as a column vector in a higher dimensional space. The transformation matrix is block-diagonal and the blocks $\mathcal{D}^{[0]}(R)=1$ and $\mathcal{D}^{[1]}(R)=R$ are familiar. $\mathcal{D}^{[2]}(R)$ corresponds to a transformation matrix that performs $\cdot \mapsto R \cdot R^\top$ on the flattened, $9$ dimensional tensor $T^{[2]}$. These matrices represent rotations on rank $\cart$ Cartesian tensors.}
    \label{fig:cartesian_trafo_rule}
\end{figure}

Just as above, we can now investigate which multiplication rules are allowed, in the sense that they allow us to do the diligent accounting of multiplication rule (3) given a certain set of tensor types that we wish to use in our model. With the tensor product, we have found a way to multiply two Cartesian tensors of ranks $\cart_1$ and $\cart_2$ and obtain a new Cartesian tensor of rank $\cart_1 + \cart_2$, with $3^{(\cart_1 + \cart_2)}$ components. But could we also ``multiply'' two higher-order types together to get a type of the same or lower-order rank? 

We already know one operation that goes from two higher-rank tensors to a lower rank: the dot product, which turns two vectors into a scalar: 
\begin{align*}
    \gt{V}^{[0]} = \gt{T}^{[1]}_{\mathbf{n}} \gt{U}^{[1]}_{\mathbf{n}} &\hspace{0.5cm}\text{(Einstein summation!).}
\end{align*}
We can see that dimensions $1$ and $2$ in $\gt{T}^{[1]}_{\mathbf{n}} \gt{U}^{[1]}_{\mathbf{n}}$ have been \textit{contracted} into dimension $0$ of $\gt{V}^{[0]}$.

It turns out that performing a generalised dot product-like operation called \emph{contraction} gives us a way to equivariantly generate tensors of lower ranks from higher ranks by summing over pairs of dimensions. 

Imagine we have a rank-3 Cartesian tensor $\gt{T}^{[3]}$ and a rank-5 tensor Cartesian tensor $\gt{U}^{[5]}$, then the \emph{contracted} combination along the pairs of dimensions $(1:3)$ and $(2:5)$:
\begin{align*}
    \gt{V}^{[4]}_{ijkl} = \gt{T}^{[3]}_{\mathbf{nmn}} \gt{U}^{[5]}_{i\mathbf{m}jkl}
\end{align*}
will transform as a rank-4 Cartesian tensor. To see that $\gt{V}^{[4]}$ will indeed transform predictably under rotations, we use the fact that rotation and translation matrices are orthonormal: $(\rot_{ia} \rot_{ja})_{ij} = (\rot_{ia} \rot^\top_{aj})_{ij} = (\rot \rot^\top)_{ij} = (\mathds{1})_{ij} = \delta_{ij}$, with $\mathds{1}$ the identity matrix, $\delta_{ij}$ the Kronecker delta\footnote{The Kronecker delta is 1 if $i=j$ and 0 otherwise.} and using Einstein notation. Using this identity and the transformation rules for $\gt{T}^{[3]} \otimes \gt{U}^{[5]}$ from above, $\gt{V}^{[4]}$ transforms as:

\begin{align*}
    \gt{V}^{[4]}_{ijkl} &= \gt{T}^{[3]}_{nmn} \gt{U}^{[5]}_{imjkl} = \delta_{no} \delta_{mp} \gt{T}^{[3]}_{nmo} \gt{U}^{[5]}_{ipjkl} \hspace{0.7cm}\text{\small{(Replacing repeated indices by a }} \delta \text{\small{ and a new index)}} \\
    &\mapsto \left( \rot_{n^\prime n} \rot_{o^\prime o} \delta_{no} \right) \left( \rot_{m^\prime m} \rot_{p^\prime p}\delta_{mp} \right) \left( 
    \rot_{n^\prime n} \rot_{o^\prime o}
    \rot_{m^\prime m} \rot_{p^\prime p}
    \rot_{i^\prime i} \rot_{j^\prime j} \rot_{k^\prime k} \rot_{l^\prime l}
    \gt{T}^{[3]}_{nmo} \gt{U}^{[5]}_{ipjkl} \right) \\
    &= 
    \left( \rot_{n^\prime n} \rot_{o^\prime n}\right) \left( \rot_{m^\prime m} \rot_{p^\prime m} \right) \left( 
    \rot_{i^\prime i} \rot_{j^\prime j} 
    \rot_{k^\prime k} \rot_{l^\prime l} 
    \gt{T}^{[3]}_{n^\prime m^\prime o^\prime } \gt{U}^{[5]}_{ip^\prime jkl} 
    \right) \\
    &=
    \delta_{n^\prime o^\prime}
    \delta_{m^\prime p^\prime}
    \rot_{i^\prime i} \rot_{j^\prime j} 
    \rot_{k^\prime k} \rot_{l^\prime l}
    \gt{T}^{[3]}_{n^\prime m^\prime o^\prime } \gt{U}^{[5]}_{ip^\prime jkl}  \\
    &= 
    \rot_{i^\prime i} \rot_{j^\prime j} 
    \rot_{k^\prime k} \rot_{l^\prime l}
    \gt{T}^{[3]}_{n^\prime m^\prime n^\prime} 
    \gt{U}^{[5]}_{i m^\prime jkl} \\
    &= \rot_{i^\prime i} \rot_{j^\prime j} \rot_{k^\prime k} \rot_{l^\prime l} \gt{V}^{[4]}_{ijkl}.
\end{align*}


This is the transformation of a rank 4 Cartesian tensor, so $\gt{V}^{[4]}$ is indeed a rank 4 Cartesian tensor! We have thus found a consistent way to generate lower-rank Cartesian tensors by contracting higher-rank tensors, for example from the tensor product of Cartesian tensors. We denote the contraction operation as $\mathfrak{C}$ and write 
\begin{align}
    \gt{V}^{[0]} &&=&& \mathfrak{C}_{(1:2)} [\gt{T}^{[1]} \otimes \gt{U}^{[1]}] &&:=&& \gt{T}^{[1]} \otimes_{(1:2)} \gt{U}^{[1]},\\
    \gt{V}^{[4]} &&=&& \mathfrak{C}_{(1:3)(2:5)} [\gt{T}^{[3]} \otimes \gt{U}^{[5]}] &&:=&& \gt{T}^{[3]} \otimes_{(1:3)(2:5)} \gt{U}^{[5]},
\end{align}
to signify the contraction of dimension $1$ with $3$ ($n$ with $n$ in $\gt{T}^{[3]}$), and $2$ with $5$ ($m$ in $\gt{T}^{[3]}$ with $m$ in $\gt{U}^{[5]}$) of the intermediate rank $8$ Cartesian tensor $\gt{T}^{[3]} \otimes \gt{U}^{[5]}$ that results from the tensor product. 

Let us step back and reflect on what we have learned. We saw that scalars and vectors are not the only types with which information can transform consistently with rotations and reflections. Indeed, we found many examples of Cartesian tensors of different ranks that transform differently to scalars and vectors. And with the (contracted) tensor product and the notion of Cartesian tensors, we now have the tools to build new, higher-rank types and contract them to lower-rank types which transform consistently under rotations and translations. By keeping with the rules of diligent accounting set out at the start of \Cref{ssec:equivariant_gnns}, we can use these Cartesian tensor types and the above operations to build equivariant GNNs with higher-order tensor types\footnote{The creation of higher-order Cartesian tensors and their contraction to lower order tensors here mostly serves didactic purposes. If one were to build a Cartesian tensor-based GNN one would need to include asymmetric contractions, in addition to the symmetric contractions introduced in the main text, to reach generality. We omit this here and instead make the link to spherical tensors and harmonics, which form the backbone of many current equivariant GNNs.}. However, in the literature higher-order Cartesian tensor GNNs are not common, mostly because of the exponential cost in memory that comes with creating and storing many tensors of higher ranks\footnote{A rank $\cart$ Cartesian tensor in 3D has $3^\cart$ components}. 
A recent exception is the TensorNet model \citep{simeon2023tensornet} which uses Cartesian tensors up to rank 2.


Instead of building a Cartesian GNN, let us turn to two natural questions that arise with the tools to build and contract Cartesian tensors at hand. Discussing these questions will lead us to the common class of equivariant GNNs with higher tensor types used in the current machine learning literature.
\begin{enumerate}
    \item \textbf{Exhaustiveness}: How do we know these Cartesian tensor types capture all possible types of transformation for rotational symmetries, or could there be some types that are constructed in yet a different way?
    \item \textbf{Usefulness}: Does using higher-rank tensors and contracting them down to scalars (sometimes colloquially referred to as \emph{scalarisation}) yield any new information about our point cloud? In particular, can we obtain any new invariants that cannot be built from scalars and vectors via the interactions in the scalar-vector GNNs we saw above?
\end{enumerate}

To answer these questions, let's switch gears and utilize a well-established mathematical result regarding the representation theory of the rotation and rotation group $SO(3)$. This will lead us to the concept of spherical tensors, which are tensor types corresponding to the irreducible representations of $SO(3)$. More details will be explained in the next section.

\newpage
\subsection{From Cartesian tensors to spherical tensors -- reducible to irreducible representations}
\label{sec:cartesian_to_spherical}


For the rotation group $SO(3)$, the fundamental building blocks into which all types can be decomposed are known. Such \textsc{LEGO}-block-like types are called irreducible representations, or \emph{irreps} for short. In our language, irreps correspond to tensor types that can be combined\footnote{By combined we here mean taking the direct sum $\oplus$, the mathematical definition of concatenation, and performing a change of basis.} to create any other possible tensor type -- including any Cartesian tensor that we can come up with via the rules in \Cref{sssec:cartesian_equivariant_gnns}. Conversely, any other type that is not an irrep can be decomposed into its irrep components.
The irreps of $SO(3)$, can be numbered by non-negative integers $l=0,1,2,\dots$ and we will refer to them as \emph{spherical tensor} types $\gt{T}^{(l)}$\footnote{Notice that we use parentheses $()$ to explicitly denote spherical tensor types as opposed to $[]$, which denoted Cartesian tensor types.}. A spherical tensor of type $l$ is $2l+1$-dimensional, that is it can be represented as a list of $2l+1$ components. All other (finite-dimensional)\footnote{This decomposition result also holds for many infinite dimensional representations, e.g. for $L^2(SO(3))$ via the Peter-Weyl Theorem \citep{zee2016group}.} representations of $SO(3)$ are fully reducible into these irreps\footnote{Representations of compact Lie groups, such as $SO(3)$ or $O(3)$, are fully reducible \citep{zee2016group}.}. These results are well known in the mathematical subfield of representation theory, but their proofs are beyond the scope of this hitch-hike\footnote{\citet{zee2016group}, chp.~4 is a good textbook reference.}. 

\begin{figure}
    \centering
    \includegraphics[width=0.75\textwidth]{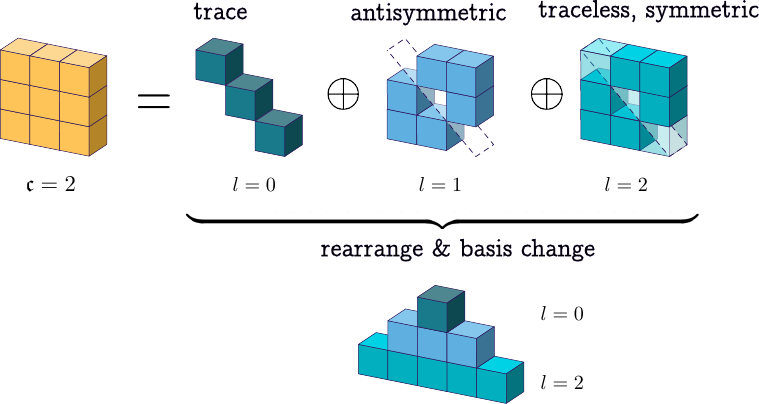}
    \caption{Visual illustration of the decomposition of a $\mathfrak{c}=2$ Cartesian tensor into spherical tensors. The $9$ dimensional Cartesian vector decomposes into a $1$ dimensional $l=0$ (scalar), a $3$ dimensional $l=1$ (vector) and a $5$ dimensional $l=2$ component. This decomposition corresponds to a change of basis in the $9$ dimensional space in which $\gt{T}^{[2]}$ lives and the change of basis equations are given in the main text.}
    \label{fig:cartesian_to_spherical_decomposition}
\end{figure}

Rather than provide proofs, we will continue with examples to build intuition and connect the theory to the machine learning practice. We said irreps are the atomic building blocks of any possible representation and we can decompose more complicated representations (types) into their atomic components. Let us see an example of this and decompose an arbitrary rank-2 Cartesian tensor $\gt{T}^{[2]}$ into irreps (visually illustrated in \Cref{fig:cartesian_to_spherical_decomposition}):

\definecolor{blue0}{HTML}{016170}
\definecolor{blue1}{HTML}{3A7294}
\definecolor{blue2}{HTML}{02AFBE}
\renewcommand*{\arraystretch}{0.6}
\begin{align*}
    \gt{T}^{[2]} =& 
    \footnotesize
    \begin{bmatrix}
        T_{xx} & T_{xy} & T_{xz} \\ 
        T_{yx} & T_{yy} & T_{yz} \\ 
        T_{zx} & T_{zy} & T_{zz} 
    \end{bmatrix}
    \\ &=
    \color{blue0}
    \dfrac{T_{xx} + T_{yy} + T_{zz}}{3}
    \footnotesize
    \begin{bmatrix}
        1 &  &  \\ 
         & 1 &  \\ 
         &  & 1
    \end{bmatrix} \hspace{3cm}& (l=0)
    \\ &\color{blue1}+
    \footnotesize
    \begin{bmatrix}
         & \frac{T_{xy} - T_{yx}}{2} & \frac{T_{xz} - T_{zx}}{2} \\ 
        -\frac{T_{xy} - T_{yx}}{2} &  & \frac{T_{yz} - T_{zy}}{2} \\ 
        -\frac{T_{xz} - T_{zx}}{2} & -\frac{T_{yz} - T_{zy}}{2} &  
    \end{bmatrix} \hspace{3cm}& (l=1)
    \\ &\color{blue2}+
    \footnotesize
    \begin{bmatrix}
        T_{xx} & \frac{T_{xy}+T_{yx}}{2} & \frac{T_{xz}+T_{zx}}{2} \\ 
        \frac{T_{xy}+T_{yx}}{2} & T_{yy} & \frac{T_{yz}+T_{zy}}{2} \\ 
        \frac{T_{xz}+T_{zx}}{2} & \frac{T_{yz}+T_{zy}}{2} & T_{zz}
    \end{bmatrix} 
    - \dfrac{T_{xx} + T_{yy} + T_{zz}}{3}
    \footnotesize
    \begin{bmatrix}
        1 &  &  \\ 
         & 1 &  \\ 
         &  & 1
    \end{bmatrix}
    \hspace{3cm}& (l=2)
    \\ &=
    \color{blue0}
    \overbrace{\dfrac{T_{xx} + T_{yy} + T_{zz}}{3}}^{\lambda_1}
    \footnotesize
    \begin{bmatrix}
        1 &  &  \\ 
         & 1 &  \\ 
         &  & 1
    \end{bmatrix} \hspace{3cm}& \color{blue0} (l=0)
    \\ &
    \color{blue1}
    + \overbrace{\frac{T_{yz}-T_{zy}}{2}}^{\lambda_2}
    \footnotesize
    \begin{bmatrix}
         &  &  \\ 
         &  & 1 \\ 
         & -1 &  
    \end{bmatrix} 
    \normalsize 
    + \overbrace{\frac{T_{zx}-T_{xz}}{2}}^{\lambda_3}
    \footnotesize
    \begin{bmatrix}
         &  & -1 \\ 
         &  &  \\ 
        1 &  &  
    \end{bmatrix}
    \normalsize 
    + \overbrace{\frac{T_{xy}-T_{yx}}{2}}^{\lambda_4}
    \footnotesize
    \begin{bmatrix}
         & 1 &  \\ 
        -1 &  &  \\ 
         &  &  
    \end{bmatrix}
     & \color{blue1} (l=1)
    \\ &
    \color{blue2}
    +\overbrace{\frac{2T_{xx} - T_{yy} - T_{zz}}{3}}^{\lambda_5}
    \footnotesize
    \begin{bmatrix}
        1 &  &  \\ 
         & -1 &  \\ 
         &  &  
    \end{bmatrix} +
    \overbrace{\frac{T_{xy}+T_{yx}}{2}}^{\lambda_6}
    \footnotesize
    \begin{bmatrix}
         & 1 &  \\ 
        1 &  &  \\ 
         &  &  
    \end{bmatrix} +
    \overbrace{\frac{T_{xz}+T_{zx}}{2}}^{\lambda_7}
    \footnotesize
    \begin{bmatrix}
         &  & 1 \\ 
         &  &  \\ 
        1 &  &  
    \end{bmatrix} \\ &
    \color{blue2}
    + \overbrace{\frac{T_{yz}+T_{zy}}{2}}^{\lambda_8}
    \footnotesize
    \begin{bmatrix}
         &  &   \\ 
         &  & 1 \\ 
         & 1 &  
    \end{bmatrix} +
    \overbrace{\frac{2T_{zz} - T_{xx} - T_{yy}}{3}}^{\lambda_9}
    \footnotesize
    \begin{bmatrix}
         &  &   \\ 
         & -1 &  \\ 
         &  & 1 
    \end{bmatrix}& \color{blue2} (l=2)
\end{align*}

In the above equations, missing matrix entries represent $0$'s and the $l=0,1,2$ components are color-coded for clarity. Each component corresponds to a subspace of the $9$-dimensional linear space that contains $\gt{T}^{[2]}$. To understand how these components relate to quantities we already know, assume for a moment that $\gt{T}^{[2]} = \gv{v} \otimes \gv{w}$ is the result of a tensor product of two vectors. In this case $T_{xy} = v_x w_y, T_{yx} = v_y w_x$ and so on. The $(l=0)$ part in the decomposition has one component ($\lambda_1$) and is therefore one-dimensional. $\lambda_1$ corresponds to the trace of $\gt{T}^{[2]}$ and it remains unchanged under global rotations. In the case where $\gt{T}^{[2]}$ is the tensor product of two vectors, the $(l=0)$ component holds the same information as the dot-product $\gv{v} \cdot \gv{w}$ which is, of course, invariant:

\[
\gv{v} \cdot \gv{w} = v_xw_x + v_yw_y + v_zw_z  = 3\lambda_1 
\]

The $(l=1)$ part is the asymmetric part of the matrix and it corresponds to the cross-product space. To see this, assume $\gt{T}^{[2]} = \gv{v} \otimes \gv{w}$, then:

\begin{equation}\label{eq:l1_cross_prod}
\gv{v} \times \gv{w} = 
\footnotesize 
\begin{bmatrix}
    v_yw_z - v_zw_y \\
    v_zw_x - v_xw_z \\
    v_xw_y - v_yw_x
\end{bmatrix}
= 
\footnotesize 
\begin{bmatrix}
    \lambda_2 \\
    \lambda_3 \\
    \lambda_4
\end{bmatrix}
\end{equation}

and we observe that the coefficients in the decomposition of the $(l=1)$ component of $\gv{v} \otimes \gv{w}$ coincide with those of the cross product $\gv{v} \times \gv{w}$! Therefore the $(l=1)$ part is three-dimensional and transforms under rotations just as a vector would, as can be seen from \Cref{eq:l1_cross_prod}.

Finally, the $(l=2)$ component of $\gt{T}^{[2]}$ corresponds to a traceless, symmetric matrix and from the basis decomposition above we can see that it is 5 dimensional, with coefficient $\lambda_5$ to $\lambda_9$. Unfortunately, the $(l=2)$ subspace does not have as easy of a relation to quantities we have already seen -- but from the relation between the irreps to the spherical harmonics, which we will see shortly, we can gain the intuition that $l\geq2$ irreps capture ever higher frequency information and therefore increasingly fine-grained angular information.


To summarize, we have seen that the 9-dimensional rank-2 Cartesian tensor can be decomposed into a 1D, 3D and 5D part which correspond to the $l=0,1,2$ irreps respectively
\begin{equation}\label{eq:decomposition}
    3 \otimes 3 = 1 \oplus 3 \oplus 5.
\end{equation}
In this slightly loose notation, the $\oplus$ symbol denotes the direct sum of a 1D, 3D and 5D linear space. This amounts to a concatenation in machine learning lingo.

More generally, we can decompose \emph{any} Cartesian tensor into irreps of various $l$, albeit with slightly more complicated decompositions than the above. And as we have seen in the above example, the special cases $l=0$ and $l=1$ correspond to familiar scalar-type and vector-type information respectively and transform under global rotations as scalars or vectors would.

\paragraph{Rotating spherical tensors.} For types of higher $l$, the transformation behaviour under rotations is also known and given by the so-called Wigner D-matrices. The \emph{D} here stands for \emph{Darstellung}, the German word for \emph{representation}. The Wigner D-matrix is how one can represent a rotation in the $2l+1$ dimensional space of a degree $l$ spherical tensor.
Explicit formulas for these matrices exist\footnote{The formulas for Wigner D-matrix coefficients are basis-specific and we will get into the choice of basis for the irreps in the next section. In short, the real space spherical harmonics provide the basis of irreps that is typically used in the Machine Learning community.} and are implemented in packages such as \texttt{e3nn} \citep{geiger2022e3nn}. For example, the $(l=2)$ part above transforms under a rotation $\rot$ as:
\begin{equation}
    \label{eq:wigner-d-trafo}
    [\lambda_5 \dots \lambda_9]^\top  \to \mathcal{D}^{(2)}(\rot) [\lambda_5 \dots \lambda_9]^\top ,
\end{equation}
where $\mathcal{D}^{(2)}(\rot)$ is the $5 \times 5$ Wigner matrix representing the rotation $\rot$ acting on a feature of type $l=2$.

\paragraph{Tensor products of spherical tensors} The tensor product and tensor contraction were the two key operations we encountered that allowed us to move up and down the rank-ladder of Cartesian tensors to produce tensors of higher rank or contract them down to lower ranks. Unfortunately, the tensor product $\gt{S}^{(l_1)} \otimes \gt{T}^{(l_2)}$ of two spherical tensors $\gt{S}^{(l_1)}$ and $\gt{T}^{(l_2)}$ is generally not a spherical tensor anymore. However, as we have learned the spherical tensor types are the atomistic building blocks into which all other types can be decomposed, and so we can decompose the product $\gt{S}^{(l_1)} \otimes \gt{T}^{(l_2)}$ back into spherical tensors.

\begin{figure}
    \centering
    \includegraphics[width=0.7\textwidth]{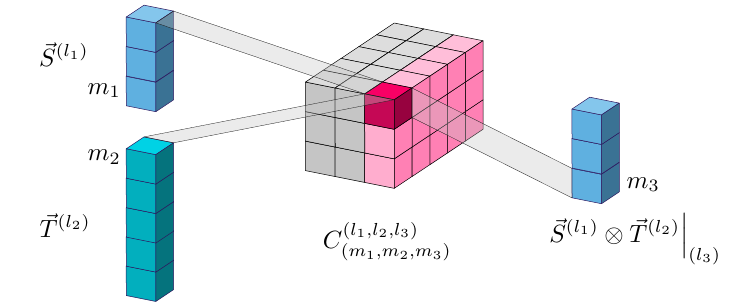}
    \caption{Illustration of the Clebsch-Gordan coefficients drawn as a three-dimensional list of numbers. Note that these are \emph{not} directly a Cartesian or spherical tensor, but merely arranged as a multidimensional list of numbers for visual convenience. In the illustration, $l_1=1, l_2=2, l_3=1$. Therefore, $C^{(l_1, l_2, l_3)}$ has $(2l_1+1)(2l_2+1) (2l_3+1) = 45$ components. The element $C^{(l_1, l_2, l_3)}_{(m_1, m_2, m_3)}$ is highlighted in red for clarity. It encodes how to weight the product of $m_1$ and $m_2$ for $m_3$ such that the output transforms as an $l_3$ tensor. Bear in mind that this term is only one of the 12 terms (highlighted in pink) that make up $m_3$ and that for our selection of $l_1=1, l_2=2$ there would also be an $l_4=2$ and $l_5=3$ output in the tensor product, which are not shown in the visualisation.}
    \label{fig:clebsch_gordan}
\end{figure}

Luckily for us, a general formula for the decomposition of a tensor product of irreps of $SO(3)$ is known. As a rule, the $(l_1 l_2)$-dimensional tensor product of two spherical tensors of ranks $l_1$ and $l_2$ decomposes into:
\begin{equation}
    \label{eq:general_decomposition}
    l_1 \otimes l_2 = \vert l_1 - l_2 \vert \oplus \vert l_1 - l_2 + 1\vert \oplus \dots \oplus (l_1 + l_2 -1) \oplus (l_1 + l_2).
\end{equation}
This means the $l_1l_2$-dimensional product decomposes into exactly one spherical tensor for each rank between the absolute difference $\vert l_1 - l_2\vert$ and the sum $l_1 + l_2$. As a result, the tensor product of two spherical tensors results in $l_1 + l_2 - \vert l_1 - l_2 \vert + 1$ new spherical tensors and each valid combination of $(l_1, l_2, l_3)$, where $l_3$ is the type of the output tensor, is colloquially referred to as a tensor \emph{path}.

As an example, if $l_1=1, l_2=2$, then the output contains an $l_3=1, l_3=2$ and $l_3=3$ part: $1\otimes2 = 1 \oplus 2 \oplus 3$. Each of these components corresponds to one tensor path and \Cref{fig:clebsch_gordan} shows the $l_3=1$ part pictorially.


The coefficients of the decomposition are given by the \emph{Clebsch-Gordan coefficients}, and these are again implemented in packages such as \texttt{e3nn}. The Clebsch-Gordan coefficients are clearly basis-dependent, and we will get to the common choice of basis in the machine learning community shortly. 
In component form, they are often denoted by the symbol $C^{(l_1, l_2, l_3)}_{(m_1, m_2, m_3)}$. Specifically, this symbol denotes the weight of the product of component $m_1$ of the $l_1$-type factor with component $m_2$ of the $l_2$ type factor that goes into making the $m_3$ component of the outgoing $l_3$-type spherical tensor. \Cref{fig:clebsch_gordan} has a visual illustration. $C^{(l_1, l_2, l_3)} \in \mathbb{R}^{(2l_1+1) \times (2l_2+1) \times (2l_3+1)}$ can be seen as a three-dimensional list of numbers. 


\paragraph{Spherical harmonics and picking a basis for spherical tensors.} Now that we have learned about some of the properties of spherical tensor types and how they behave under rotations and tensor product operations, let us turn to the question of choice of basis. This is crucial, since we always have to present these tensors through a list of numbers in the computer. To obtain spherical tensors, we use a family of functions that can be used to map a point on the unit sphere to a spherical tensor of a given degree $l$: the real-valued\footnote{The physics community, most textbooks and internet resources tends to use the complex-valued spherical harmonics. The machine learning community on the other hand typically uses the real valued version for memory and computational reasons. While this makes reading about spherical harmonics confusing at times, the two formulations are related by a simple change of basis. See \cite{geiger2022e3nn} for more details on this.} spherical harmonics $Y^l_m$.

\begin{figure}
    \centering
    \includegraphics[width=\textwidth]{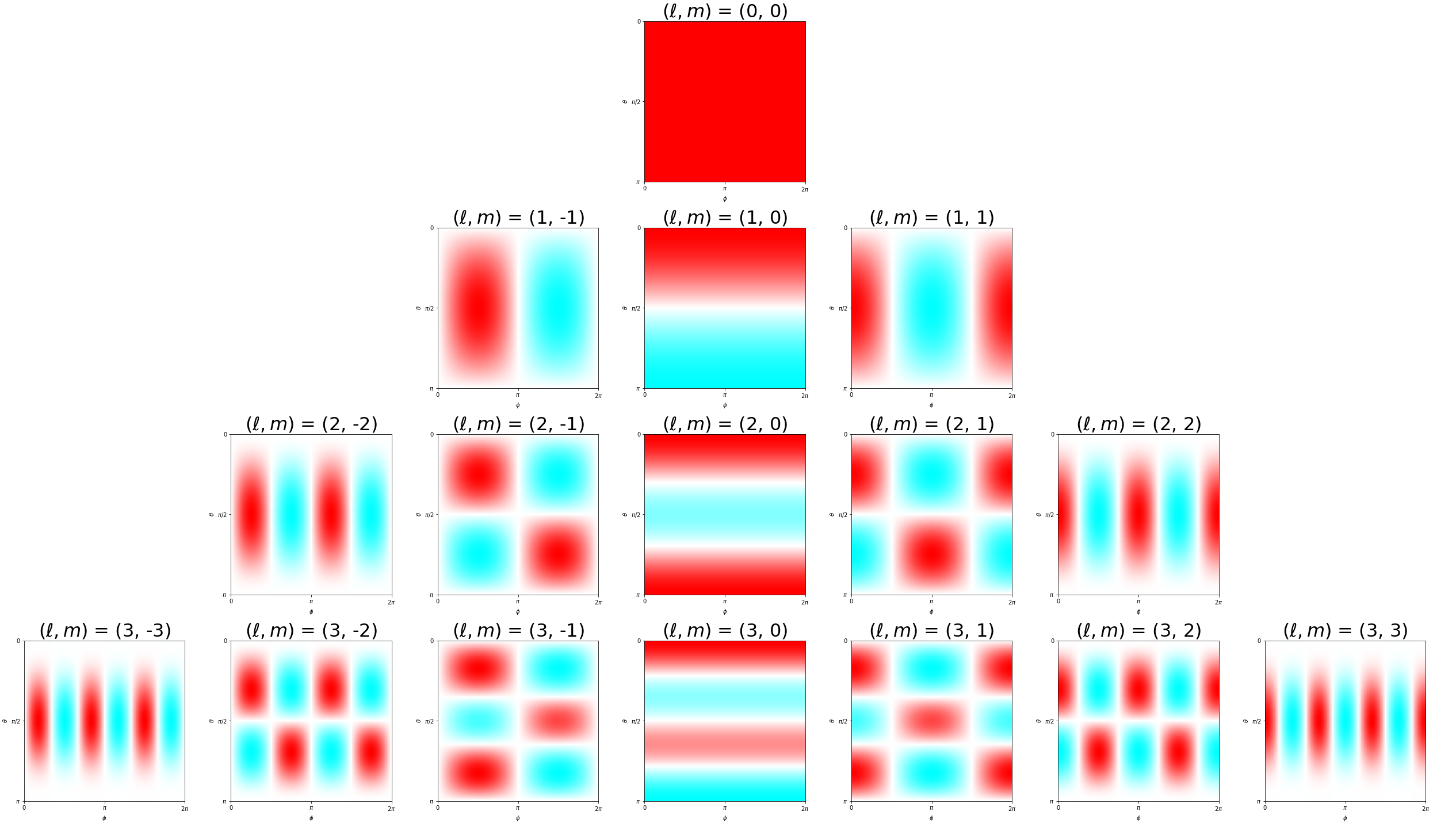}
    \caption{The real spherical harmonics $\gt{Y}^{(l)}_m(\hat{x})$ plotted as a function of $\hat{x} = (\phi, \theta)$ on the unit sphere. The horizontal axis corresponds to the azimuthal angle $\phi$, and the vertical axis to the polar angle $\theta$. The saturation of the color at any point represents the magnitude of the spherical harmonic and the hue represents the sign. Figure taken from \href{https://en.wikipedia.org/wiki/Table_of_spherical_harmonics\%23Visualization_of_real_spherical_harmonics}{Wikipedia}.}
    \label{fig:spherical-harmonics}
\end{figure}

The real spherical harmonic $Y^{l}_m: S^2 \to \mathbb{R}$ 
takes a point on the 2-dimensional unit sphere $S^2$ and maps it to a number. For a given degree $l$ there are $2l+1$ possible values of $m$, usually numbered by $-l, \dots, l-1, l$. If we collect all components for a given $l$ into $\gt{Y}^{(l)} = (Y^l_{-l}, \dots, Y^l_{l-1}, Y^l_{l}): S^2 \to \mathbb{R}^{2l+1}$, then the resulting $2l+1$ dimensional object will transform equivariantly as
\begin{equation}
    \gt{Y}^{(l)}(\rot\hat{v}) = \mathcal{D}^{(l)}(\rot) \gt{Y}^{(l)}(\hat{v}).
\end{equation}
This is precisely the transformation behaviour of spherical tensors of degree $l$ we saw in \Cref{eq:wigner-d-trafo}, making $\gt{Y}^{(l)}(\hat{v})$ a spherical tensor.
In other words, the functions $Y^l_m$ for a given $l$ provide a basis set of functions for the order $l$ irrep of $SO(3)$.

Upon choosing a basis for each irrep $l$, the spherical harmonics are unique up to a sign and normalisation constant\footnote{See \cite{geiger2022e3nn} for a proof. If we assume a usual Cartesian coordinate system, then we can pick a basis for each irrep $l$ by choosing an arbitrary rotation axis and requiring the infinitesimal generator for rotations around that axis be diagonal. In most conventions people choose the $z$-axis, and this results in the equations for the spherical harmonics that we give below. In \texttt{e3nn} the \href{https://docs.e3nn.org/en/latest/guide/change_of_basis.html}{$y$-axis is chosen as a default}, such that the components of $\gt{Y}^{(1)}$ are proportional to $(x, y, z)$.}. Let $\hat{v}=(x,y,z)$ be a point on the unit sphere $S^2 \subset \mathbb{R}^3$. Then, in the most common basis convention, the real spherical harmonics up to $l=2$ are given by
\begin{align}
    \label{eq:spherical_harmonics}
    Y^{(0)}_0(\hat{v}) &= c_0 \\
    Y^{(1)}_{-1}(\hat{v}) &= c_1 y, \hspace{0.2cm}
    Y^{(1)}_{0}(\hat{v}) = c_1 z, 
    \hspace{0.2cm}
    Y^{(1)}_{1}(\hat{v}) = c_1 x, \\
    Y^{(2)}_{-2}(\hat{v}) &= c_2 xy, 
    \hspace{0.2cm}
    Y^{(2)}_{-1}(\hat{v}) = c_2 yz, 
    \hspace{0.2cm}
    Y^{(2)}_{0}(\hat{v}) = \dfrac{c_2}{2\sqrt{3}} (2z^2 - x^2 - y^2), \hspace{0.2cm} \\
    & Y^{(2)}_{1}(\hat{v}) = c_2 xz, 
    \hspace{0.2cm}
    Y^{(2)}_{2}(\hat{v}) = \dfrac{c_2}{2} (x^2-y^2),\notag
\end{align}
where $c_0, c_1$ and $c_2$ are constants that depend upon the choice of normalisation\footnote{See \href{https://en.wikipedia.org/wiki/Table_of_spherical_harmonics\%23\%E2\%84\%93_=_0_2}{this table on Wikipedia} for a more extensive list including a common choice of normalisation constants.}. 

At this point you might become somewhat uneasy, because the spherical harmonics were only defined on the unit sphere $S^2$, so how can we use them to go from an arbitrary vector in $\mathbb{R}^3$ to a spherical tensor? It turns out that the spherical harmonics can be extended beyond the unit sphere\footnote{Indeed, we gave the equations in \Cref{eq:spherical_harmonics} already in the generalised form. In the generalised form, the spherical harmonics $Y^l_m$ can be chosen to be polynomials in the $x, y, z$ coordinates. In fact, the generalised spherical harmonics of degree $l$ are \href{https://en.wikipedia.org/wiki/Spherical_harmonics\%23Spherical_harmonics_in_Cartesian_form}{homogeneous harmonic polynomials of degree $l$}, i.e. polynomials such that $\Delta p(\vec{x})$ = 0 and that each term contains a product of $l$ variables. The spherical harmonics can then be interpreted as the restriction of homogeneous harmonic polynomials of degree $l$ onto the sphere, giving them an algebraic rather than a representation theoretic definition as we have done above.} to the rest of $\mathbb{R}^3$. While this works, it can lead to numerical instabilities when working with vectors of non-unit length and requires careful normalisation. As an alternative, we could split the vector into a radial part $\norm{\vec{v}}$ and a directional part $\hat{v} = \vec{v}/\norm{\vec{v}}$ and treat them separately. This is the path most architectures based on spherical tensors choose.

Besides the nice fact that the spherical harmonics form a basis of the irreps of $SO(3)$ they also have many other nice properties. Because most are less directly relevant for building equivariant neural networks out of them, we only mention some that help gain an intuition about those functions in passing. The spherical harmonics are solutions to the Laplace equation $\Delta f(\hat{x}) = 0$ on the sphere and they form a complete and orthogonal basis of all smooth functions on the sphere. Therefore any smooth function on the sphere can be decomposed in a (possibly infinite) sum of spherical harmonics, weighted by different coefficients. Decomposing smooth functions into their spherical harmonics components is analogous to decomposing functions on $\mathbb{R}$ into their Fourier components, and the degree $l$ of the spherical harmonics determines the frequency of the function. Low $l$ functions change slowly as one walks along the sphere, while higher $l$ functions change ever more quickly, as can also be seen in \Cref{fig:spherical-harmonics}.


\paragraph{Using spherical tensors to build equivariant architectures.}
Now that we have learned about spherical tensors, what if instead of Cartesian tensors, we used tensors based on these irreps, as our basis for message passing and accounting? Then we know that these in principle capture all possible types of transformations under rotational symmetries (if we go to high enough $l$), answering question (1) that we posed at the outset of this subsection. We could further make use of many nice theoretical properties of these irreps, that are known from the representation theory of $SO(3)$. This idea is in fact precisely what many of the models in the equivariant Geometric GNN literature follow, and it leads us to equivariant GNNs with spherical tensors.


\newpage
\subsection{Equivariant GNNs with spherical tensors -- Irreducible representations}
\label{sssec:equivariant_gnns_spherical_tensors}

\begin{tcolorbox}[enhanced,attach boxed title to top left={yshift=-2mm,yshifttext=-1mm,xshift = 10mm},
colback=cyan!3!white,colframe=cyan!75!black,colbacktitle=white, coltitle=black, title=Key idea,fonttitle=\bfseries,
boxed title style={size=small,colframe=cyan!75!black} ]
Spherical EGNNs not only restrict the set of learnable functions to equivariant ones, they also use \emph{spherical} tensor components, which correspond to the irreducible representations of $SO(3)$, as their feature types. This choice comes naturally because of the intimate relationship of spherical tensors with the rotation group $SO(3)$, which gives spherical tensors many convenient properties.
\end{tcolorbox}

In this section we focus on equivariant GNNs that operate with spherical tensors, the irreducible represntations and therefore the \emph{natural} types of the rotation group $SO(3)$. As with Cartesian tensors, the crux to building equivariant networks is to diligently keep track of the types of different features and how they transform. The architectures in this category leverage the tools from representation theory that we have introduced in the previous section. In summary:
\begin{itemize}
    \item The \emph{spherical harmonics} are useful as a basis of irreps of degree $l$ and to project vectors onto their spherical tensor components\footnote{In a functional or Fourier theory picture, when evaluating $Y^l_m(\hat{x}^\prime)$ we are projecting a delta-function $\delta(\vec{x}-\vec{x}^\prime)$, or a sum thereof in the case of a point cloud, onto its spherical degree $l$ component. To better understand the functional perspective of these models, \citep{uhrin2021through} and \citep{blum2022machine} are excellent references and written mathematically more precisely than our more loose and introductory discussion.}.
    \item The \emph{Clebsch-Gordan coefficients} allow us to decompose a tensor products of spherical tensors into its spherical tensor components.
    \item The \emph{Wigner D-matrices} represent rotations $\rot$ for spherical tensors of degree $l$ and enable us to rotate these tensors.
\end{itemize}
As we venture into equations for an example spherical EGNN, the formulas will inevitable become more complex and decorated with indices. When this happens, its important to remind yourself that the simple idea from \Cref{ssec:equivariant_gnns} that underlies all this: We need to perform diligent accounting of tensor types, while adding learnable parameters where we can. The three tools above and all the indices are simply there to help us operate on and keep track of our accounting with spherical tensors.



\paragraph{Example spherical EGNNs in the literature.}
Before we dive into details, let us name a few recent examples of spherical EGNNs. The spherical EGNN family contains methods like Clebsch-Gordan Net \citep{kondor2018clebsch}, TFN \citep{thomas2018tensor}, NeuquIP \citep{batzner2022nequip}, SEGNN \citep{brandstetter2021geometric}, MACE \citep{batatia2022mace}, Equiformer \citep{liao2022equiformer} and many others, including networks using the concepts of steerability and equivariance introduced by \cite{cohen2016group}. Most of these models build on the \href{https://e3nn.org/}{e3nn} library \citep{geiger2022e3nn}, which makes it easy to work with spherical tensors by implementing the real spherical harmonics, Wigner D-matrices and tools to easily compute, decompose and parameterise tensor products for network layers. 

\subsubsection{An example spherical EGNN}
Having heard of some of the spherical EGNNs in the literature, let us now construct a convolutional, spherical EGNN that will be similar to TFN \citep{thomas2018tensor} to showcase the main ideas. 

\begin{figure}
    \centering
    \includegraphics[width=\textwidth]{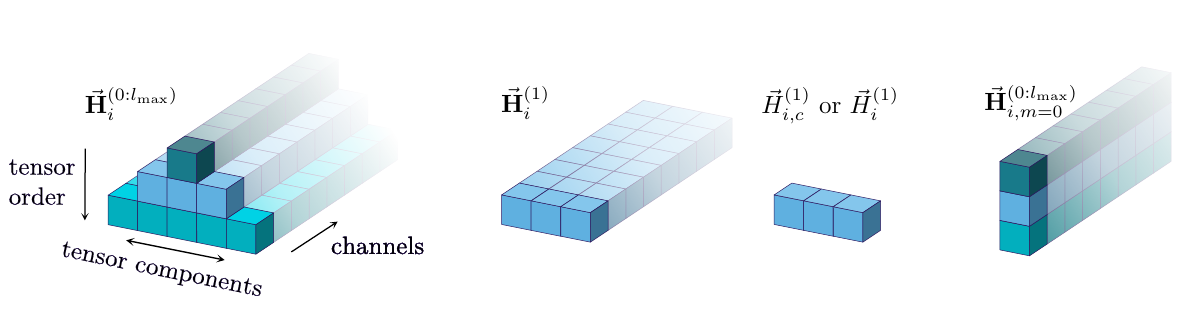}
    \caption{Visualisation of the spherical tensor list at node $i$ and various of its slices. The `lists' are drawn as pyramids to separate out the three axes.}
    \label{fig:spherical_tensor_list}
\end{figure}

\paragraph{Node features and spherical tensor lists.}
Spherical EGNNs use tensors for node and sometimes also for edge features. We  consider only node features, but the concepts extend straightforwardly to edge feature tensors. Let us denote the (hidden) features at node $i$ by a list of geometric tensors of various $l$, from $0$ to $l_\text{max}$\footnote{For current architectures this is often $l_\text{max}=1$ or $2$. Also, if you have never encountered the symbol $:=$ before, it means ``the left-hand-side is defined as ...''.}:

\begin{equation}
    \label{eq:spherical_tensor_list}
    \gt{\c{H}}^{(0:l_\text{max})}_i
    := \bigoplus_{l=0}^{l_\text{max}}  \gt{\c{H}}^{(l)}_{i}  
    = \begin{bmatrix}
     \gt{\c{H}}^{(0)}_{i} \\
     \vdots
     \\
     \gt{\c{H}}^{(l_\text{max})}_{i}
    \end{bmatrix}
\end{equation}
It is important to pause and realise that the condensed notation $\gt{\c{H}}^{(0:l_\text{max})}_i$ has 3 axes: It has a \emph{channel-axis} (indicated by the boldface), because we might have more than a single type $l$ tensor at that node. For simpler notation, we assume the same number of channels for all types $l$. $\gt{\c{H}}^{(0:l_\text{max})}_i$ also has a \emph{tensor-axis} (indicated by the $(0:l_\text{max})$ superscript), which reminds us that we are dealing with a concatenation of spherical tensors of various types. Finally, each tensor type $l$ also a \emph{tensor-component-axis}, which has $2l+1$ elements. \Cref{fig:spherical_tensor_list} illustrates this visually. When dealing with slices along these axes, we use the letters $c$, $l$ and $m$ for channel, tensor and tensor-component axes respectively. Also remember from \Cref{sec:cartesian_to_spherical} that $\gt{\c{H}}^{(0)}$ are simply scalars, and $\gt{\c{H}}^{(1)}$ are vectors\footnote{Possibly with axes interchanged and a different normalisation, depending on the definiton of the spherical harmonics that is used.}.

\paragraph{Splitting the radial and angular parts.} Next, let us briefly touch upon how to obtain spherical tensors from vectors, as most atomistic problems start from a point-cloud of various atom types in 3D space. How do we get spherical tensors from such a system? Spherical EGNNs split the vectors such as the displacement vector $\vec{x}_{ij}$ between node $i$ and $j$ into a radial $x_{ij} = \norm{\vec{x}_{ij}}$ and a directional part $\hat{x}_{ij}$. The directional part is projected to spherical tensors with the spherical harmonics $\gt{Y}^{(l)}(\hat{x}_{ij})$. The radial part is a scalar, but because of its range $x_{ij} \in [0, \infty)$ it can quickly lead to diverging or vanishing scales when multiplied several times with itself, which happens as we stack multiple equivariant layers. To remedy this, we can apply various non-linear functions to $x_{ij}$ to map the range $[0, \infty) \to [-1, 1]^{n_b}$, just as we did for distance-based invariant GNNs in \Cref{sec:invariant-gnns}. Here the number of basis\footnote{This is not a basis in the mathematical sense, as the functions do not span the entire space. Usually, the `basis' here is a subset of mathematical basis the space of (reasonable, e.g. square-integrable) functions over $[0, \infty)$, because there are infinitely many basis functions.} functions $n_b$ as well as the choice of basis functions $\psi_b$ are hyperparameters. A common choice for $\psi_b$ are the radial basis functions centered around various length-scales of interest\footnote{Another choice are the Bessel functions for example.}. We would then carry one channel, or one set of channels, per basis function.

\paragraph{Equivariance and parameterising the tensor product.} 
To allow learning to take place, we need to assign learnable parameters in our architecture that can be updated during training. We know how to do parameterise operations between scalars, as they behave just like normal numbers in standard neural networks. But, how can we parameterize the tensor product? 

In \Cref{sec:cartesian_to_spherical} we learned that the tensor product of two spherical tensors $\gt{S}^{(l_1)} \otimes \gt{T}^{(l_2)}$ can be decomposed into $2 \ \text{min}(l_1, l_2) + 1$ spherical tensors of various degrees. The Clebsch-Gordan coefficients $C^{l_1, l_2, l_3}_{m_1, m_2, m_3}$ dictate how the components of the tensors need to be combined to retain equivariance, so we cannot add learnable weights to simple products of tensor components without breaking equivariance. Instead we can assign a learnable weight to each output spherical tensor in the decomposition, indexed by $l_3$, as these tensors are independently equivariant:
\begin{equation}
    \gt{S}^{(l_1)} \l{\otimes} \gt{T}^{(l_2)} = \bigoplus_{l_3 = \vert l_1 - l_2 \vert}^{l_1+l_2} \l{w}_{l_1, l_2, l_3} \ \gt{S}^{(l_1)} \otimes\gt{T}^{(l_2)}\big\vert_{(l_3)}.
    \label{eq:simple_parametrised_tp}
\end{equation}
In \Cref{eq:simple_parametrised_tp} the $\l{w}_{l_1, l_2, l_3}$ are learnable parameters, as indicated by the underline. There is one learnable parameter for each path $(l_1, l_2 \to l_3)$ in the Clebsch-Gordan decomposition and $\gt{S}^{(l_1)} \otimes\gt{T}^{(l_2)}\big\vert_{(l_3)}$ means extracting only the $l_3$ component of the tensor product. We can generalise this to the case of two lists of spherical tensors. Assuming a single channel each for the start, we get
\begin{equation}
    \gt{S}^{(0:l_\text{max})} \l{\otimes} \gt{T}^{(0:l_\text{max})} 
    =
    \bigoplus_{l_3=0}^{l_\text{max}}\left( \sum_{\text{Paths} (l_1, l_2 \to l_3)} \l{w}_{l_1, l_2, l_3} \ \gt{S}^{(l_1)} \otimes\gt{T}^{(l_2)}\big\vert_{(l_3)}  \right).
    \label{eq:list_parameterised_tp}
\end{equation}
Here $\text{Paths} (l_1, l_2 \to l_3)$ refers to all valid paths from an $l_1$ and $l_2$ to and $l_3$ part. We will sometimes just abbreviate this by $(l_1, l_2, l_3)$. 

Let us consider a specific example to get a feel for this equation. Consider the case where we want to take the parameterised tensor product of a feature $\gt{H}^{(0:1)}$ with $\gt{Y}^{(0:1)}{(\hat{x}_{ij})}$, which may represent a spherical tensor list coming from an initial feature embedding or from a convolutional filter as we will see in a moment.
Suppressing the dependence of $\gt{Y}^{(l)}(\hat{x}_{ij})$ on $\hat{x}_{ij}$ for a moment for readability
\begin{align}
\label{eq:param_tp_example}
\begin{bmatrix}
     \gt{Y}^{(0)}  \\
     \gt{Y}^{(1)} 
\end{bmatrix}
&\l{\otimes}
\begin{bmatrix}
     \gt{H}^{(0)}  \\
     \gt{H}^{(1)}
\end{bmatrix}
\\ &= 
\begin{bmatrix}
     \l{w}_{0,0,0} (\gt{Y}^{(0)} \otimes \gt{H}^{(0)})\big\vert_{(0)} +
     \l{w}_{1,1,0} (\gt{Y}^{(1)} \otimes \gt{H}^{(1)})\big\vert_{(0)}\\
     \l{w}_{0,1,1} (\gt{Y}^{(0)} \otimes \gt{H}^{(1)})\big\vert_{(1)} + 
     \l{w}_{1,0,1} (\gt{Y}^{(1)} \otimes \gt{H}^{(0)})\big\vert_{(1)} + 
     \l{w}_{1,1,1} (\gt{Y}^{(1)} \otimes \gt{H}^{(1)})\big\vert_{(1)}
\end{bmatrix}.\notag
\end{align}
If this equation looks somewhat intimidating, remember that a tensor product with an $l=0$ term is just a verbose way of writing scalar multiplication, and that the tensor product of two $l=1$ terms essentially amounts to the dot product or the cross-product if the output $l=0$ or $l=1$ respectively.

We can also make the weights in \Cref{eq:param_tp_example} dependent on a continuous parameter, such as the distance $x_{ij}$ between nodes $i$ and $j$ from which the tensors might originate. If we have chosen a radial basis functions, this would make simply the weights channel dependent. In such cases, we write $\l{\otimes}_{(x_{ij})}$ and $\l{w}_{l_1, l_2, l_3}(x_{ij})$.

\begin{figure}
    \centering
    \includegraphics[width=0.5\textwidth]{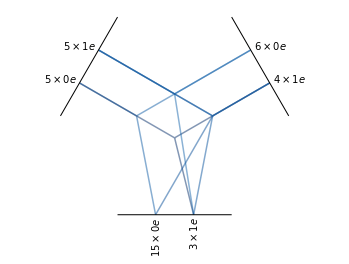}
    \caption{Fully connected tensor product two a spherical tensor lists. One with 5 $(l=0)$ and $(l=1)$ tensors and one with 6 $(l=0)$ and 4 $(l=1)$ tensors. At the output, we ask for 15 $(l=0)$ and 3 $(l=1)$ channels. How many paths are there? The answer is discussed in the main text. We ignore the parity indices \texttt{o} (odd) and \texttt{e} (even), as we do not consider reflections for simplicity. Figure taken from the \href{https://docs.e3nn.org/en/stable/api/o3/o3_tp.html}{\texttt{e3nn}} documentation by \citep{geiger2022e3nn}.}
    \label{fig:e3nn_fully_connected_tp}
\end{figure}

The expression above was for a single channel only. When $\gt{\c{S}}^{(0:l_\text{max})}$ and $\gt{\c{T}}^{(0:l_\text{max})}$ are lists of multiple spherical tensors in different channels, we can perform the tensor product between each valid degree-channel pair and collect the results in different channels through different weighted combinations.
\begin{equation}
    {\gt{\c{U}}^{(0:l_\text{max})}} = {\gt{\c{S}}^{(0:l_\text{max})}} \l{\otimes} \gt{\c{T}}^{(0:l_\text{max})}.
\end{equation}
Note that we did not collect all possible degrees in the output, but restricted $\gt{U}$ to tensors of maximally degree $l_\text{max}$. All higher order spherical tensors that would be created by the tensor product are simply dropped.
In this multi-channel, multi-degree case, the same idea as before holds, but many more paths become available. In \texttt{e3nn}, such operations are adequately called \emph{fully-connected tensor products} and \citep{geiger2022e3nn} have a nice way of illustrating what is going on, which is illustrated in \Cref{fig:e3nn_fully_connected_tp}. The illustration shows the fully connected tensor product two a spherical tensor lists: One with 5 $(l=0)$ and $(l=1)$ tensors and one with 6 $(l=0)$ and 4 $(l=1)$ tensors. At the output, we ask for 15 $(l=0)$ and 3 $(l=1)$ channels. How many paths are there? There are $5\cdot6$ paths from $(0,0\to0)$ and $5\cdot4$ from $(1,1\to0)$. We ask for 15 $(l=0)$ features in the output, so there are 750 paths ending in $(l=0)$. For $l=1$, the analogous computation over all valid paths $(0,1\to1)$, $(1,0\to1)$ and $(1,1\to1)$ gives 210 paths ending in $(l=1)$, so in total we can assign 960 learnable parameters to this fully-connected tensor product.

\paragraph{Building graph-convolutional filter from fully-connected tensor products}
We can now use the fully-connected tensor product, together with the projections of displacement vectors to spherical tensors from the previous two sections to define a filter function which we can use in the graph convolution to construct messages. Here is an example of a filter function, that is essentially\footnote{The subtle and not very important difference is that in the fully connected tensor product we allow different weights for all paths, while the original TFN ties together all weights with the same filter and input $l$ (i.e. $l_1$ and $l_2$).} the filter function used in TFN, 
\begin{equation}
    \l{\textsc{Filter}}\left(\vec{x}_{ij}, \gt{\c{H}}_j^{(0:l_\text{max})}\right) = \gt{Y}^{(0:l_\text{max})}(\hat{x}_{ij}) \l{\otimes}_{(x_{ij})} \gt{\c{H}}_j^{(0:l_\text{max})}.
\end{equation}
We can then use this with a permutation invariant aggregation function of our choice to define the message step. For sum-aggregation,
\begin{equation}
    \label{eq:learnable_graph_conv_spherical}
    \gt{\c{M}}^{(0:l_\text{max})}_{i} = \sum_{j \in \mathcal{N}(i)} \l{\textsc{Filter}}\left(\vec{x}_{ij}, \gt{\c{H}}_j^{(0:l_\text{max})}\right)
\end{equation}
is a possible, equivariant message and aggregation definition.

\paragraph{Channel-mixing.}
The fully-connected tensor product gave us a way to parameterise multiplications of spherical tensor lists. Another simple way to add learnable parameters for a single spherical tensor list is to perform type-wise channel mixing. The type-wise channel mixing operation linearly combines the channels of the same type within a spherical tensor list:
\begin{align}
    \l{\textsc{Mix}}\left(\gt{\c{M}}_i^{(0:l_\text{max})}\right) 
    = 
    \bigoplus_{l=0}^{l_\text{max}} \l{w}^{(l)}_{c^\prime c} \gt{M}^{(l)}_{i,c} 
\end{align}
Again, for simplicity we set the same number $c$ of channels for all types $l$, but in practice we could, of course, have a different number of channels for each type.

\paragraph{Non-linearities and gating.}
Finally, we can apply component-wise non-linearities to our features or messages. For scalars, we can apply component-wise non-linearities just as normal. For higher degrees of $l$ in a spherical tensor list $\gt{\c{M}}_i^{(0:l_\text{max})}$, applying component-wise non-linearities looses equivariance. Instead, a common trick is to perform non-linear \emph{gating}. This means that we apply a nonlinearity to a, possibly learnable, combination of scalars\footnote{This can be a simple MLP of the scalars $\gt{\c{M}}_i^{(0)}$ for instance.} in $\gt{\c{M}}_i^{(0:l_\text{max})}$ and then use this as a weight to scale the tensors of a given degree $l$. In equations, 
\begin{equation}
    \l{\textsc{Gated-nl}}\left( \gt{\c{M}}_i^{(0:l_\text{max})}\right)
    = 
    \eta_0(\gt{\c{M}}_i^{(0)}) 
    \oplus
    \left(
    \bigoplus_{l=1}^{l_\text{max}} \eta_l\left(\l{\textsc{Gate}}\left(\gt{\c{M}}_i^{(0)}\right)\right) \ \gt{\c{M}}_i^{(l)}
    \right)
\end{equation}
Here, $\eta_l$ are non-linear functions and the above equation is understood to apply channel-wise and $\l{\textsc{Gate}}: \mathbb{R} \to \mathbb{R}$ is an arbitrary function with parameters, such as an MLP, that is applied channel-wise.

\paragraph{Putting the pieces together.}
Finally let us put the pieces together and write out the equations for a full layer of our spherical EGNN.

\begin{equation}
    \l{\textsc{Upd}}\left(\gt{\c{H}}^{(0:l_\text{max})}_{i}, \gt{\c{M}}^{(0:l_\text{max})}_{i}\right) = \gt{\c{H}}^{(0:l_\text{max})}_{i}
    + \l{\textsc{Gated-nl}}\left(\l{\textsc{Mix}}\left(\gt{\c{M}}^{(0:l_\text{max})}_{i}\right)\right)
\end{equation}

The update function performs a residual update of the spherical tensor features at each node $i$ by first performing the graph convolution with the learnable filter (\Cref{eq:learnable_graph_conv_spherical}) to construct and aggregate messages. Next, we mix the channels within a message and apply gated nonlinearities.

The updated node feature then becomes 
\begin{equation}
    \gt{\c{H}}^{(0:l_\text{max})}_{i} \leftarrow \l{\textsc{Upd}}\left(\gt{\c{H}}^{(0:l_\text{max})}_{i}, \gt{\c{M}}^{(0:l_\text{max})}_{i}\right).
\end{equation}
This completes a single layer of our example spherical EGNN. As with the scalar-vector EGNN in \Cref{subsec:equivariant-gnns-cartesian}, we can then apply pooling and read out only the scalar channel at the output, in case we are interested in an invariant prediction. As an added bonus, we can now also read out higher equivariant types, in case we wanted to predict equivariant quantities such as an atomic systems dipole moment  (an $(l=1)$ quantity) or quadrupole moment (an $(l=2)$ quantity) for instance.

\paragraph{Relations to Tensor Field Networks.}
Tensor Field Network \citep{thomas2018tensor} is one of the earliest approaches proposing this type of $\group{G}$-equivariant GNN with spherical tensors, and is very similar to the exemplary spherical EGNN we just built. It has the benefits of processing higher-order tensors directly and in an equivariant fashion, resulting in SO(3)-equivariant model predictions. Despite its well devised theoretical framework, TFN suffers from a few shortcomings. For instance, retaining all the non-zero tensor products $\l{\otimes}$ up to degree $l_\text{max}$ becomes computationally unfeasible as we scale $l_\text{max}$ since it requires $O(l_\text{max}^3 C)$ multiplications. Therefore most e3nn networks are limited to the computation of $l_\text{max}=1$ or $2$ and $l_f \le 2$ or $3$ for the filter functions $l$. Notably, when restricting the tensor product to only scalars (up to $l = 0$), we obtain updates of the form similar to \Cref{eq:schnet}. Similarly, when using only scalars and vectors (up to $l = 1$), we obtain updates of the form similar to \Cref{eq:painn-s}, \Cref{eq:painn-v} and \Cref{eq:painn-u}.


\subsubsection{Optimisations and improvements.} Several issues highlighted in TFNs have been tackled by subsequent approaches. Below we give a brief selection of a few papers that use interesting ideas for optimisations and improvements. We only give the key ideas, and refer to the cited papers for details.

\citet{fuchs2020se} proposed SE(3)-Transformers, which adapt the TFN framework to use an equivariant variant of the self-attention operation during aggregation.
SEGNN \citep{brandstetter2022geometric} proposes equivariant non-linear convolutions by introducing steerable MLPs that transform equivariant representations. 
SEGNN follows a standard MPNN framework and uses steerable MLPs for the message as well as update steps, essentially providing a recipe for for building equivariant MPNNs with Spherical tensor features.
The Equiformer model \citep{liao2022equiformer} successfully combines these ideas by interleaving equivariant self-attention aggregation with equivariant non-linear updates into Transformer-style blocks (note that Equiformer is still a local model aggregating from neighbours within a cutoff radius).
It is also worth noting that the self-attention weights for both spherical as well as Cartesian equivariant GNNs are invariant quantities which are computed by contracting or scalarising geometric information.
The attention weights are used to re-weight neighbourhood features during equivariant message passing.

\begin{tcolorbox}[enhanced,attach boxed title to top left={yshift=-2mm,yshifttext=-1mm,xshift = 10mm},
colback=cyan!3!white,colframe=cyan!75!black,colbacktitle=white, coltitle=black, title=Key idea,fonttitle=\bfseries,
boxed title style={size=small,colframe=cyan!75!black} ]
\textbf{Why do we need higher rank tensors if all geometric quantities exist only in 3D?}
For learning to simulate molecular dynamics with a TFN-style model (NequIP), \citet{batzner2022nequip} showed empirical evidence that increasing the tensor rank lead to a steeper learning curve and improved data efficiency for (albeit models were evaluated on the same systems they were trained on).
It is intuitive to expect that higher rank tensors enable models to build more \emph{expressive}\appref{app:sec:expressivity} representation that can better fit the training data.
\citet{joshi2022expressive} present one possible explanation for why we need higher order tensors for maximally powerful equivariant models.
\\
\includegraphics[width=0.8\linewidth]{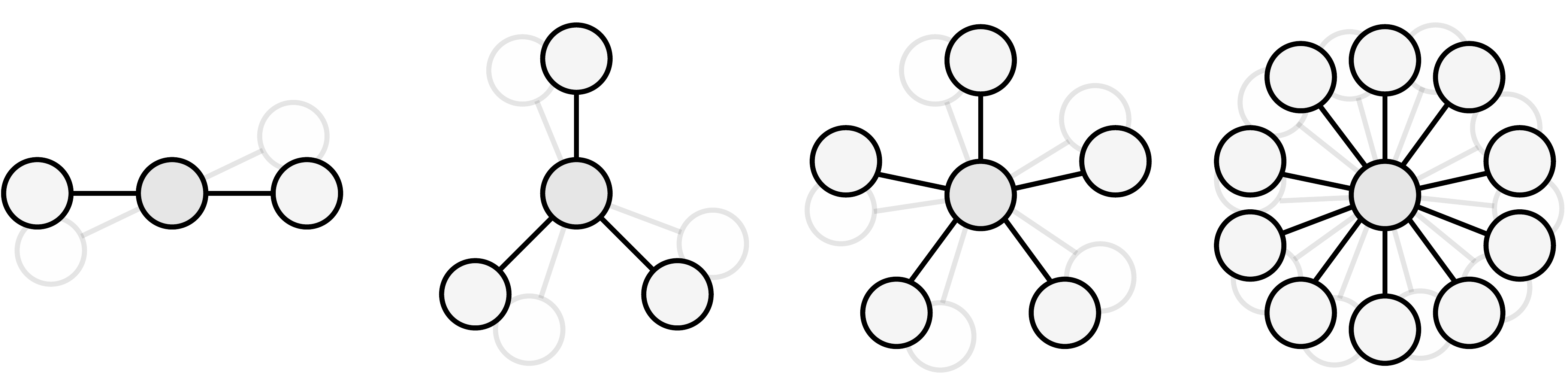}\\
Consider an experiment where a single layer equivariant GNN (i.e. one round of message passing) is tasked to distinguish two \emph{distinct} rotated versions of the $L$-fold symmetric structures above using the aggregated equivariant features. An $L$-fold symmetric structure does not change when rotated by an angle $\frac{2\pi}{L}$ around a point (in 2D) or axis (3D).\\
\citet{joshi2022expressive} showed that layers using order $L$ spherical tensors are unable to identify the orientation of structures with rotation symmetry higher than $L$-fold, \textit{i.e.} two distinct rotated versions of the input having the same equivariant features.
Try it yourself: summing together a symmetric set of 3D vectors pointing to the origin will always cancel out and lead to a zero vector, no matter how the structure is rotated!
\\
This observation can be understood based on the rotational symmetry of the spherical harmonics which serve as the underlying orthonormal basis for equivariant tensor features. 
Similar to the Fourier expansion for 1D signals, 
the spherical harmonic expansion is employed for converting Cartesian vectors to spherical signals in equivariant GNNs. 
The tensor order of the spherical harmonic bases determines the rate of oscillation of the approximated function on the sphere. 
In the Fourier expansion, it is not feasible to accurately approximate a high-frequency function solely using low-frequency sinusoidal waves. Similarly, when truncating the spherical harmonic expansion to an order lower than the fold of the rotational symmetry, the rotationally symmetric vectors act as a higher frequency function. Consequently, the lower frequency bases cannot preserve the orientation of these vectors.\\
Thus, higher rank tensors enable equivariant GNNs to construct spherical features of local sub-graphs at finer angular resolution and granularity of detail.
\end{tcolorbox}

\begin{figure}[h!]
    \centering
    \includegraphics[width=0.55\linewidth]{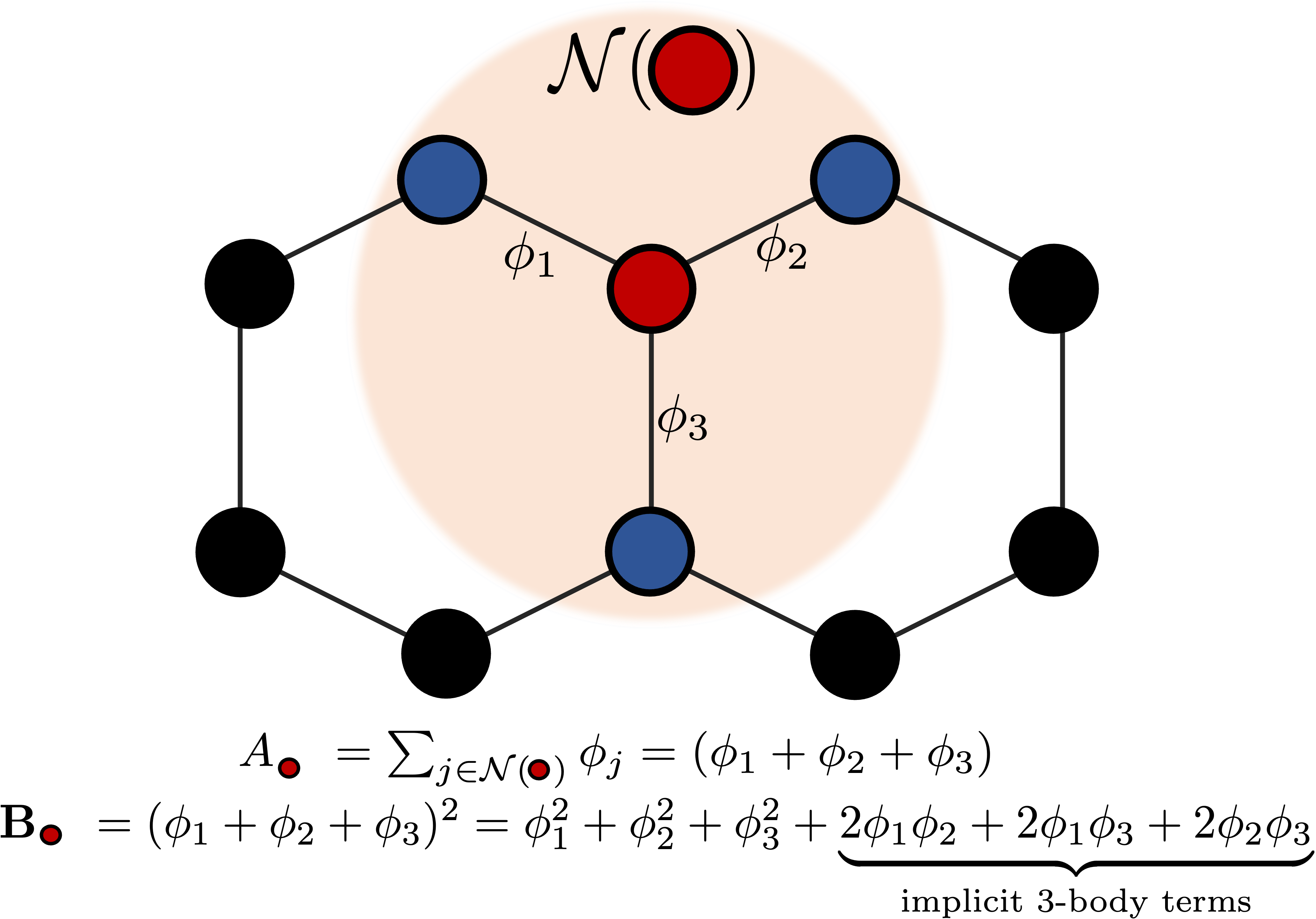}
    \caption{\textbf{The density trick for computing many-body terms} \citep{drautz2019atomic}. 
    After aggregating the neighbourhood features, we compute repeated tensor products of these summed neighbourhood features with themselves. 
    This approach saves the effort of having to symmetrise or generate all $k$-tuples in more standard many-body expansions such as \citet{klicpera2020directional, gasteiger2021gemnet}. 
    Calculating the product $(a + b + \dots)^k$ implicitly includes terms such as $a^l b^{k-l}$, instead of calculating each of them individually.
    Figure credit: Harry Shaw.}
    \label{fig:density-trick}
\end{figure}

MACE \citep{batatia2022mace} attempts to incorporate many-body interaction terms\footnote{Many-body effects refer to the collective behaviour of a large number of interacting constituents. They are needed for an accurate description of both the structure and dynamics of large chemical systems.} by relying on a clever factorization of higher-order terms into products of two-body representation, which builds on the popular Atomic Cluster Expansion (ACE) formalism \citep{drautz2019atomic}. The key idea here is to exchange summation and multiplication\footnote{This is referred to as \emph{density-trick} in the physics-ML literature and comes from \citep{drautz2019atomic}.} to reduce the number of multiplication operations: As a simple example without tensors, think of the expression $(a+b)^2$. This contains the terms $a^2, ab, ba, b^2$ but required us to perform only one multiplication\footnote{The cost of this is that the coefficients of these terms are coupled.}. MACE essentially introduces an efficient algorithm to compute a parameterised tensor-product of expressions of the form $\l{\bigotimes}_{b=1}^B(\sum_{i=1}^{n_i} \gt{A}^{(l)}_i)$, illustrated in \Cref{fig:density-trick}. The same idea is also used in Allegro \citep{musaelian2022learning}, but without message passing. This allows easier parallelisation for simulating very large systems across multiple GPUs as the computation at each node is fully local.

eSCN \citep{passaro2023reducing} tackles the computational scalability issue of higher rank tensor products by reducing the SO(3) equivariant convolutions to mathematically equivalent convolutions in $SO(2)$, making the tensor product easier to compute. This is achieved by aligning the node embeddings' primary axis with the edge vector\footnote{This trick sometimes called as \emph{point-and-shoot} in computational electromagnetics.}, which reduces the rotational symmetry to rotations around that axis, making the problem effectively 2 dimensional. While coming at the cost of two extra Wigner D-matrix rotations to align and unalign the spherical tensors to the given axis, this trick effectively sparsifies the Clebsch-Gordan coefficients and thereby leads to computational speedups for $l>1$.
EquiformerV2 \citep{liao2024equiformerv} uses the eSCN trick to scale the Equiformer model to hundreds of millions of parameters and improved performance on the large-scale catalysis tasks.

\clearpage


\section{Unconstrained Geometric GNNs}
\label{subsec:non-spm-gnns}


\begin{tcolorbox}[enhanced,attach boxed title to top left={yshift=-2mm,yshifttext=-1mm,xshift = 10mm},
colback=cyan!3!white,colframe=cyan!75!black,colbacktitle=white, coltitle=black, title=Key idea,fonttitle=\bfseries,
boxed title style={size=small,colframe=cyan!75!black} ]
Unlike other methods, architecturally unconstrained GNNs do not `bake' symmetries into their architecture, leading to greater flexibility in model design and more diverse optimization paths. Instead, they let the model learn approximate symmetries, encourage approximate symmetries through loss terms or data augmentation, or enforce symmetries through alternate strategies such as (global or local) canonization. 
\end{tcolorbox}

\textbf{Overview. }
We have seen that previous Geometric GNN families, by design, confine the set of learnable functions to equivariant ones, aligning with the goal of accurately modeling equivariance. However, this constraint raises concern about potential impediments to the neural network optimization process. The idea is that a more unconstrained model, i.e. not bound to equivariance, may traverse more diverse optimization paths (ultimately converging to equivariant functions). As motivated by \citet{duval2023faenet}, this increased flexibility could empower the model to capture the intricacies of the data more effectively. In contrast, the strict adherence to equivariant constraints may limit the optimization paths available, or change the optimization landscape in a way that hinders our algorithms, for example, by altering the conditioning or prevalence of local minima\footnote{The two arguments can converge if the impairment corresponds to a spectrum distribution of the eigenvalues of the Hessian around the fixed points, significantly limiting escape possibilities (e.g., negative eigenvalues) around these fixed points.}. 
This raises the question of whether the benefits of enforcing Euclidean equivariance as an inductive bias truly offset a potential reduction in optimization diversity within constrained learning spaces.

A useful parallel may be Deep Learning regularization, where equivariant models, akin to a regularization paradigm, enforce specific constraints on the functional learning space. While these constraints are intended to promote desirable properties, such as provable or guaranteed equivariance, unconstrained GNNs are concerned by the potential downside — the risk that such constraints may overly regularize the model, hindering its capacity to fully express the intricacies of the data. 
A few approaches \citep{wang2022regularized, hu2021forcenet, du2022se, zitnick2022spherical, kaba2023equivariance, duval2023faenet, wang2023generating, pozdnyakov2023smooth} make this argument, proposing a data-driven view of symmetries as opposed to the usual model-based view. 

\textbf{Data augmentation and soft constraints}. To explore the potential of this idea, let us consider image classification. While most modern Convolutional Neural Networks and Visual Transformers \citep{dosovitskiy2020image} are not scale- or rotation-equivariant \citep{weiler20183d}, they can still learn approximate equivariance through rotation and scale diversity in the training data. This adaptability extends to invariant prediction tasks on 3D images, where \citet{gerken2022equivariance} showed that data augmentation methods match invariant networks in accuracy with significantly lower computational cost at inference time. While the results are less successful for equivariant tasks, their work suggest that leveraging unconstrained GNNs holds promise as an effective approach.

ForceNet \citep{hu2021forcenet} was one of the first unconstrained GNN architectures to explore data augmentation as a soft symmetry constraint. This adaptation of \citet{sanchez2020learning} to 3D atomic systems proposes to implicitly learn symmetries via data augmentation procedures such as adding diverse rotations of the same geometric graph to the training data. 
Despite showing promising results, ForceNet's accuracy vs. scalability gains were not significant enough to constitute a true Pareto optimal improvement. 

Interestingly, architectures from the parallel field of 3D point cloud processing for natural scenes and shapes \citep{guo2020deep} also rely heavily on data augmentation.
Models are generally trained with random rotations, translations, and crops of the point cloud, along with corruptions to its categorical values \citep{qi2017pointnet++}.
Additional loss terms can also be added as regularisation or soft constraints that encourage the model to preserve symmetries without strictly enforcing it.


Another Geometric GNN with soft constraints is the SCN model \citep{zitnick2022spherical}, an
equivariant architecture that initially utilizes spherical harmonics to represent channel embeddings with explicit orientation information. However, it later relaxes the equivariant constraint to enable more expressive non-linear transformations. Specifically, it projects spherical channels onto a grid and conducts pointwise convolutions followed by a non-linearity, facilitating intricate mixing of various degrees of spherical harmonics at the expense of strict equivariance. When released, it reached state-of-the-art performance on the Open Catalyst dataset OC20 \citep{chanussot2021open}.

\begin{figure}[h!]
    \centering
    \includegraphics[width=\linewidth]{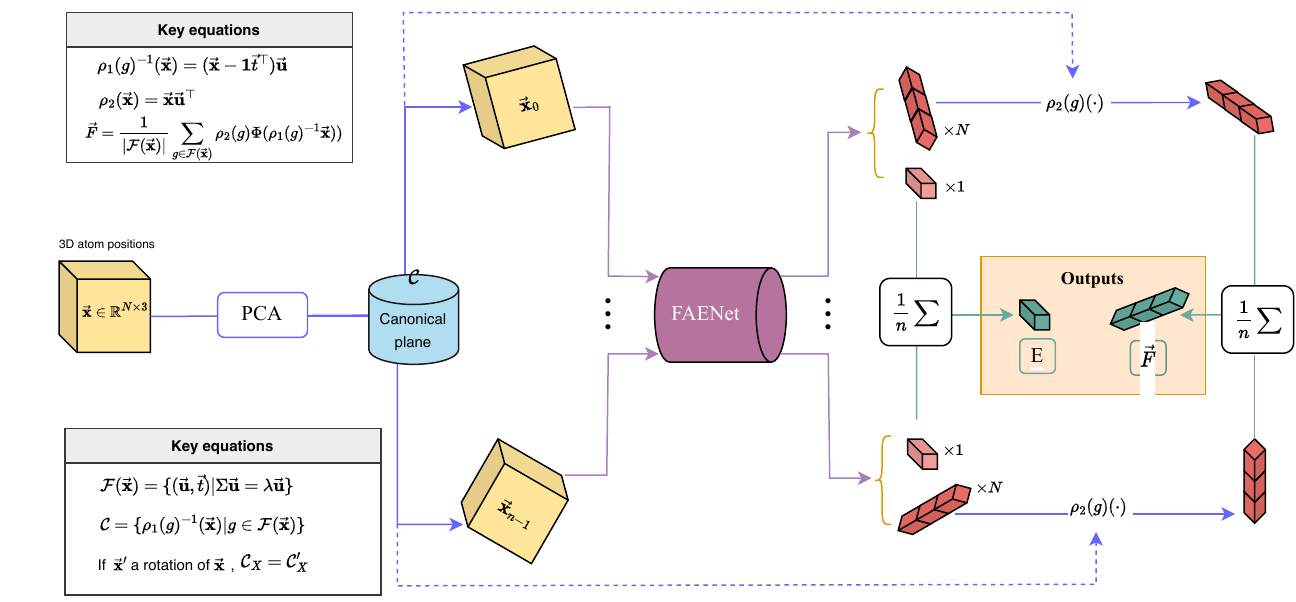}
    \caption{(Stochastic) Frame Averaging pipeline utilized by FAENet to enforce explicit (or approximate) equivariance. Input data is mapped to a global reference frame, offering a unique representation of all Euclidean transformations. It is then passed to the Geometric GNN model called FAENet before being aggregated to yield invariant or equivariant predictions.}
    \label{fig:frame-averaging}
\end{figure}

\textbf{Globally and locally canonized GNNs. }
Iterating on the data augmentation and soft constraint perspective, 
FAENet \citep{duval2023faenet} proposes a model architecture completely free of all design constraints, which directly processes atom relative positions $\vec{\vx}_{ij}$ using non-linear functions (MLPs):
\begin{align}
    \vs_i^{(t+1)} &= \l{f_1}\bigg( \vs_i^{(t)}, \sum_{j \in \nei_i} \vs_j^{(t)} \odot \l{f_2} \big(\ \gv{x}_{ij}, \ \vs_i^{(t)}, \ \vs_j^{(t)}\big) \bigg),
\end{align}
where $\l{f_2}(\cdot)$ is the convolution filter encoding the 3D geometric information (MLPs with swish activation) and $\l{f_1}(\cdot)$ the message passing update function. This inner working is ``possible'' because FAENet outsources equivariance to the data representation, allowing the model to process geometric information with complete freedom. Building upon Frame Averaging \citep{puny2022frame}, FAENet projects data points into a canonical space via Principal Components Analysis (PCA), offering a unique representation of all Euclidean transformations. This canonization, illustrated in \Cref{fig:frame-averaging}, enables to rigorously (or empirically) preserve symmetries while retaining maximal expressiveness. 

Let us briefly highlight the mechanisms of FAENet. We first need to compute the centroid of an atomic graph with $n$ nodes $\g{t} = \frac{1}{n}\cgv{x}^{\top}\mathbf{1} \in \reals^3$ and the centred covariance matrix $\Sigma = (\cgv{x} - \mathbf{1}\g{t})^\top(\cgv{x} - \mathbf{1}\g{t})$. Solving for $(\g{u}, \lambda): \Sigma \g{u} = \lambda \g{u}$ and assuming distinct eigenvalues $\lambda_1 > \lambda_2 > \lambda_3$ we can then create the \textit{frame} 
$\mathcal{F}(\cgv{x}) = \left\{ (\cgv{u}, \g{t})\ |\ \cgv{u} = [\pm \g{u}_1, \pm \g{u}_2, \pm \g{u}_3] \right\}$,
For any given function (including GNNs) $\l{f}$, $\langle\l{f}\rangle_{\mathcal{F}}(\cgv{x})$ is $\group{G}$-equivariant or invariant depending on $\rho_2$:

\begin{align}
    \langle\l{f}\rangle_{\mathcal{F}}(\cgv{x}) &\defeq \frac{1}{|\mathcal{F}(\cgv{x})|} \sum_{g\in\mathcal{F}(\cgv{x})} \rho_2(g) \l{f} (\rho_1^{-1}(g)\cgv{x})\\
    \rho_1(g)\cgv{x} &\defeq \cgv{x}\cgv{u}^\top + \mathbf{1}\g{t}^\top\\
    \rho_2(g)\cgv{y} &\defeq \begin{cases}
        \cgv{y}\cgv{u}^\top \quad \text{for equivariant predictions} \ \cgv{y} \in \R^{n\times 3} \\\
        \cgv{y} \quad\quad\hspace{1.1mm} \text{for invariant predictions}\\
    \end{cases}
\end{align}

The intuition here is that one can choose appropriate representatives of a group, act on the data ($\rho_1$), infer with a function ($\l{f}$), act back to the original space ($\rho_2$) and finally average those predictions to produce invariant or equivariant outputs. One downside of this approach is that it requires $|\mathcal{F}(\cgv{x})|$ inferences (8 for the group E$(3)$, 4 for SE$(3)$) through $\l{f}$, which can be an expensive GNN. 

To mitigate this effect, FAENet uses Stochastic Frame-Averaging (SFA) where members of a frame are \textit{sampled} instead of being exhaustively averaged over. In different terms, we apply the function $\l{f}$ on a single frame element, selected randomly at inference and at each training epoch, instead of all members. This approach does not \textit{guarantee} invariance/equivariance but allows the model to \textit{learn} it. In this perspective, it is akin to a geometry-informed data augmentation procedure over a restricted set of $|\mathcal{F}(\cgv{x})|$ frames.
It is interesting to note how the frame averaging paradigm relates to existing families of invariant and equivariant GNNs, defined in \Cref{sec:invariant-gnns} and \Cref{ssec:equivariant_gnns}. FAENet is similar to invariant GNNs as it treats geometric information as scalar quantities and uses a standard message passing scheme (i.e. unconstrained) to update node representations non-linearly. But unlike them, it can leverage relative atom positions directly instead of a pre-defined scalarisation of geometric information. Besides, by avoiding the application of symmetry-preserving equivariant operations on internal representations, FAENet `breaks' the expressivity limits of invariant and equivariant GNNs and trivially distinguishes between all known counterexamples proposed by \citet{pozdnyakov2020incompleteness} and \citet{joshi2022expressive}. Since it aggregates relative atom positions using any non-linear function, it can be considered as a \textit{many-body} approach that iteratively incorporates information coming from higher-order neighborhoods. 

A related line of theoretical work from \citet{dym2024equivariant} propose \textit{weighted} frames that provably preserve continuity, while \cite{lim2024expressive} tackle the sign ambiguity issue of PCA by utilizing a sign-equivariant network, allowing the use of a single frame but with equivariance guarantees.

\textbf{Other approaches to unconstrained GNNs. }
Instead of designing the canonization function by hand like in Frame Averaging, \citet{kaba2023equivariance} proposed to learn it using a shallow equivariant neural network. \citet{mondal2024equivariant} extend the above work by aligning the learned canonization function with the training data distribution.
Additionally, some approaches focus on building local frames rather than global canonization \citep{du2022se, pozdnyakov2023smooth}, projecting tensor information at given
orders onto these local frames when performing message passing. \citet{pozdnyakov2023smooth} proposed an unconstrained Geometric Transformer architecture based on an alternative to the frame averaging protocol, termed Equivariant Coordinate System Ensemble, which defines local coordinate systems at each atom and averages over the predictions of a non-equivariant network for each coordinate system.




\textbf{Summary. }
Geometric GNNs not explicitly enforcing symmetries into the model architecture are an emerging and under-explored line of work. Relaxing symmetry constraints in the model design allows for the implementation of more expressive architectures allowing for an interesting accuracy vs. scalability trade-off at inference time. 
However, the partial or approximate enforcement of symmetries may come at the cost of precision and physical violations in the model's prediction.
For instance, a recent benchmark by \citet{fu2023forces} found ForceNet-based molecular simulators to be unstable over long timesteps and for large systems.
Thus, understanding when to impose strict equivariance in the model as opposed to letting the model learn flexibly is an open question. 
Similarly, looking at new ways to implicitly learn exact symmetries from data \citep{dehmamy2021automatic} could lead to significant improvements in the field. The discussion is continued in \Cref{sec:discussion}.

\begin{tcolorbox}[enhanced,attach boxed title to top left={yshift=-2mm,yshifttext=-1mm,xshift = 10mm},
colback=cyan!3!white,colframe=cyan!75!black,colbacktitle=white, coltitle=black, title=Opinion,fonttitle=\bfseries,
boxed title style={size=small,colframe=cyan!75!black} ]
Unconstrained GNNs do not state or show that building symmetries into neural networks is not worthwhile.
They simply tackle the interplay between the space of functions that a model can learn and the ease of optimisation of ML algorithms. 
Whether we should rigorously enforce symmetries or not is an open question that probably depends on the application and scale of data available. 
Here, SE(3) is a small group in terms of the degrees of freedom (a rigid transformation is uniquely determined by 6 parameters), so not rigorously enforcing equivariance may be tolerable and, perhaps, desirable from the optimisation perspective. However, in other cases, incorporating such inductive bias into the model will reduce sample complexity and improve generalization. 
Finally, note that unconstrained GNNs still enforce node permutation symmetry via permutation equivariant message passing. 
\end{tcolorbox}



\clearpage

\section{Applications}
\label{sec:appli-data-coding}

\begin{figure}[h!]
    \centering
    \begin{subfigure}[b]{0.32\linewidth}
        \centering
        \includegraphics[width=\linewidth]{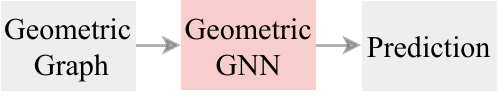}
        \caption{Property prediction}
        \label{fig:app-pred}
    \end{subfigure}
    \hfill
    \begin{subfigure}[b]{0.28\linewidth}
        \centering
        \includegraphics[width=\linewidth]{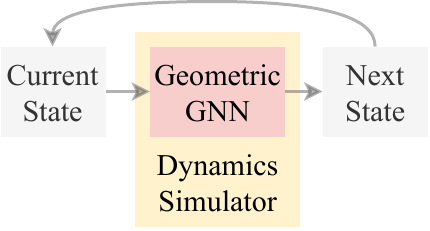}
        \caption{Dynamics simulation}
        \label{fig:app-dyn}
    \end{subfigure}
    \hfill
    \begin{subfigure}[b]{0.32\linewidth}
        \centering
        \includegraphics[width=\linewidth]{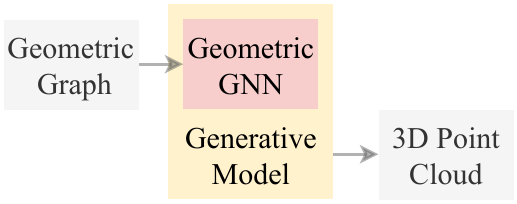}
        \caption{Generative modeling \& \\structure prediction}
        \label{fig:app-gen}
    \end{subfigure}
    \caption{\textbf{Geometric GNN applications. } Representations learnt by Geometric GNNs are used as part of task-specific pipelines in property prediction, molecular dynamics simulation, generative modeling, and structure prediction.
    }
    \label{fig:application}
\end{figure}

\textbf{Overview. }
Geometric GNNs have shown promising results across a range of application areas, spanning structural biology, biochemistry, and materials science \citep{zhang2023artificial, wang2023scientific}. These applications can be broadly categorised broadly in \Cref{fig:application} as (1) property prediction, (2) molecular dynamics simulations, (3) generative modeling, and (4) structure prediction.
This section concisely describes the utility of Geometric GNNs for each task, followed by a detailed discussion of relevant datasets.

\subsection{Tasks}
\label{subsec:applications}


\subsubsection{Property Prediction}

The most common application of Geometric GNNs is to predict functional and physical properties of geometric graphs, ranging from quantum mechanical properties of molecules and materials \citep{gasteiger2021gemnet} to the outcomes of experimental assays in drug discovery \citep{stokes2020deep}.

Conventional approaches to determining properties of 3D atomic systems are known to be resource-intensive, both for simulation and experimental approaches. For instance, quantum mechanical properties are computed using simulation techniques such as Density Functional Theory (DFT)\appref{app:sec:quantum-chemistry}\citep{kohn1996density} which often require supercomputers. 
Similarly, obtaining experimental data usually requires specialized equipment tailored to a particular application, with different requirements for biological systems, chemical systems and complex materials. 
Geometric GNNs have emerged as a fast alternative by learning to predict properties of new systems from large annotated datasets generated by historical simulations or experiments.
This expedites the screening of large libraries of molecules and materials in order to discover new candidates with specific desired properties.
Notable use cases of Geometric GNNs for property predictions include:

\begin{enumerate}
    \item Drug discovery -- where Geometric GNNs help identify promising drug candidates by modeling relevant properties of potential drugs, including proteins and small molecules interacting with the human body \citep{huang2022artificial, jamasb2023evaluating}. This type of accelerated screening can help practitioners understand a drug's potential efficacy and safety profiles, and prioritize designed/novel molecules for further experimental testing. 
    \item Material discovery -- where, by accurately predicting properties such as energy, stability, bandgap and conductivity, Geometric GNNs can help researchers optimize material compositions and structures for specific applications. Similar to drug discovery, this can help researchers discover new materials for a diversity of applications \citep{chanussot2021open, lee2023matsciml}.
    A recent example of the success of geometric GNNs is GNoME \citep{merchant2023scaling}, which merges a NequIP potential \citep{batzner2022nequip} with an active learning pipeline to create a dataset of 2.2 millions stable material structures\footnote{
    Note that challenges still exist for systems such as GNoME to effectively identify materials with specific functional properties or search the space of all possible materials; see \citet{leeman2024challenges} for more details.
    }.
    \item Environmental impact assessment -- where they help understand the behavior of substances in the environment, evaluating their toxicity, persistence, and bioaccumulation potential. This enables to make informed decisions regarding their use and disposal \citep{feinstein2021uncertainty, epa2020user}.
    \item Process optimization -- where, by accurately predicting reactivity, selectivity, and solubility, Geometric GNNs can help researchers optimize reaction conditions, design efficient separation processes, and minimize waste generation leading to more sustainable and cost-effective manufacturing processes \citep{kearnes2021open, mercado2023data}. 
\end{enumerate}

Formally, the property prediction task is formulated as a regression or classification problem, most often targeted to the full graph (e.g. energy, band gaps, thermal conductivity, stability) although it can also apply at a node level (e.g. atomic charges). A Geometric GNN model is trained to minimize the loss function $\sum_i \mathcal{L}( \ \texttt{GNN} \ ( \ \graph{G}_i) \ , \ y_i \ )$ over a dataset of $n$ 3D geometric graphs $\graph{G}_i$, paired with their corresponding property values $y_i$. The objective is to optimize the model's ability to predict a specific property by minimizing the difference between the predicted value and the ground truth property value. Commonly used loss functions include mean absolute error (MAE) for regression tasks or cross-entropy for classification tasks. Geometric GNNs are very effective for predicting materials and molecules' properties due to their ability to leverage atomic and geometric attributes while respecting data symmetries.


\subsubsection{Interatomic Potentials for Molecular Dynamics Simulation}

Molecular Dynamics (MD) simulations, also known as atomistic simulations, predict how every atom in a 3D system will move over time based on a general model of the physics governing interatomic interactions \citep{karplus2002molecular}. MD simulations are used to model a diversity of materials, including periodic crystal structures, molecules and large-scale protein structures, providing useful insights into various properties and behaviours. 
MD simulations also represent atomistic systems as a set of $N$ atoms with position vectors $\cgv{x}$ and atomic types $\m{S}$ and aim to solve Newton's equation of motion for all atoms in the system. As such, the system behaviour is governed by the potential energy $U$, which depends on the interaction of the various atoms in the system. Concretely, $U$ is the summation of of one-body $U(\gv{x}_i)$, two-body $U(\gv{x}_i,\gv{x}_j)$, three-body $U(\gv{x}_i,\gv{x}_j,\gv{x}_k)$, up to $N$-body interaction terms:

\begin{equation}
U = \sum_{i=1}^N U(\gv{x}_i) + \sum_{\substack{i,j=1;\\i\neq j}}^{N} U(\gv{x}_i,\gv{x}_j) + \sum_{\substack{i,j,k=1;\\i\neq j \neq k}}^{N} U(\gv{x}_i,\gv{x}_j,\gv{x}_k) + \cdots
\label{Eq:energy}
\end{equation}

MD simulations also have the capability to include desired thermodynamic conditions, such as temperature and pressure, by including relevant thermostat and barostat settings found in common MD simulation packages \citep{thompson2022lammps, van2005gromacs, ase-paper}. Once the potential energy $U$ of the system is known, one can obtain the forces on each atom using differentiation leading to forces on each atom:  $\gv{F}_i= -\partial U/\partial \gv{x}_i$. The acceleration of each atom is then calculated by dividing the forces by the atomic mass $m_i$. Next, the updated atomic positions are computed by numerically integrating the equations of motion enabling one to study the dynamics of atomic systems.

Geometric GNNs can be infused into MD simulations by using GNN-based interatomic potentials, meaning that $U=\Phi^1(\cgv{x}, \m{S})$ with the forces being determined by differentiation of the GNN-based potential \citep{batzner2022nequip, batatia2022mace, batatia2022design}. Alternatively, one can perform direct force prediction for each atom with another GNN (head): $\cgv{F}=\Phi^2(\cgv{x},\m{S})$ \citep{gasteiger2021gemnet}. Note that, in both cases, we minimize a loss function that measures the discrepancy (e.g. MAE) between the predicted and the ground truth values. A thorough analysis of the advantages and shortcomings of MD potentials is available in \citet{bihani2023egraffbench} and \citet{fu2023forces}. These studies suggest that MAE-based regression training methods for GNN-based MD simulations are not sufficient to guarantee MD simulation stability likely to due distribution shift. One promising alternative may be training on the dynamics of the trajectory data itself as explored by \citet{bhattoo2023learning}. Nonetheless, \citet{schaarschmidt2022learned} suggests that Geometric GNNs can avoid local minima that classical quantum mechanical simulators often get trapped in, showcasing that further research is needed to better understand the capabilities and limitations of using Geometric GNNs in MD simulations.

\subsubsection{Generative Modeling and Design}


Geometric GNNs are emerging as a powerful data encoder tool to facilitate the synthesis of new complex geometric structures including molecules, crystals, and 3D objects. Their utilization in the generation pipeline is quite recent and does not seem to have reached its full potential yet: most existing generative approaches still use traditional GNN models \citep{de2018molgan, ragoza2020learning}. 
To be more specific about their utilisation within the generative pipeline, let's first distinguish two main categories of generative methods where they are (or could be) used:

\textit{Iterative methods} sequentially generate geometric graphs, selecting actions at each step. G-Schnet and G-SphereNet \citep{gebauer2019symmetry, luo2022autoregressive}, for example, are auto-regressive models that generate 3D molecules by performing atom-by-atom completion using invariant Geometric GNNs \citep{schutt2017schnet, liu2021spherical} on the current structure. 
Alternatively, the generation of 3D atomic systems can be decomposed into compositional objects, constructed step by step using Reinforcement Learning (RL). In this case, Geometric GNNs are used to steer the action choices of the RL algorithm, acting as a reward function that optimises the generation of materials/molecules with targeted desirable properties. \cite{ai4science2023crystal} employs a GFlowNet \citep{bengio2021flow} to sequentially sample 3D crystals through the selection of the composition, space group and lattice parameters, with any property prediction model as an objective function. This domain-inspired approach enables the flexible incorporation of physical and geometrical constraints and allows to search efficiently through the entire material space.

\textit{Full-graph methods} generate all attributes at once (i.e. scalar features, geometric features and adjacency matrix), building on top of Variational Auto Encoders (VAE) \citep{ren2022invertible, pakornchote2023diffusion}, Generative Adversarial Networks (GANs) \citep{nouira2018crystalgan, long2021constrained}, Normalizing Flows \citep{satorras2021en, ahmad2022free} and Diffusion models \citep{zheng2023towards, xu2022geodiff, jiao2023crystal}. In general, these methods are trained to reconstruct the training data from a latent distribution by maximizing its likelihood, or by minimizing the discrepancy between generated samples and the real data distribution. Once trained, they can generate new instances by sampling from the learned latent space, transforming latent representations into valid geometric graphs. How are Geometric GNNs used in this framework? They are natural data encoders. Since they preserve data symmetries, they are useful to preserve the likelihood of generated samples when the 3D atomic system is rotated, reflected or translated. For VAE, they are used at train time to create latent representations of the input geometric graph before using an MLP or a GCN to (re-)produce a 3D graph \citep{xie2021crystal}. Regarding diffusion models, which is a very active area of research, \cite{hoogeboom2022equivariant} employs an equivariant network jointly operating on continuous (atom coordinates) and categorical features (atom types) in the denoising phase. Finally, GAN based methods could use a Geometric GNN discriminator to predict if the input geometric graph comes from the generator or from the training dataset.


Overall, applying Geometric GNNs to generative modeling has significant implications for accelerating progress in fields like protein-conditioned molecule generation \citep{corso2022diffdock, schneuing2022structure}, de novo protein design \citep{ingraham2019generative, dauparas2022robust, yim2023se, watson2023novo}, and materials discovery \citep{kolluru2022open}.
An example of the potential of generative models is the recent diffusion model, MatterGen \citep{zeni2023mattergen}, for inorganic crysal discovery. 
MatterGen can be finetuned for generating materials with specific properties such as formation energy, magnetic density, band gaps or bulk modulus properties. 
Despite the promise of generative models, key challenges still remain. For instance, MatterGen fails to adequately account for periodic crystal symmetries, which hinders exploration of diverse candidates.
Additionally, most generative models are currently only evaluated computationally. 
While it is understandably challenging to perform experimental validation of designed molecules and materials, a greater emphasis must be placed on more on meaningful in-silico evaluation metrics for generative models \citep{buttenschoen2023posebusters, harris2023posecheck}.


\subsubsection{Structure Prediction}

\textbf{Protein Structure Prediction. } For biomolecules, the structure prediction task entails predicting or generating a plausible set of 3D coordinates of a biomolecule given its 1D sequence representation \citep{alquraishi2021machine} (for proteins: the sequence of amino acid residues, for nucleic acids: the sequence of nucleotides).
Geometric GNNs play a central role in modern structure prediction systems like AlphaFold \citep{jumper2021highly} and RosettaFold \citep{baek2021accurate, baek2022accurate}.
At a high level, structure prediction systems consist of two modules applied sequentially: 
\begin{enumerate}
    \item The sequence module which constructs and updates latent representations of each residue, generally via a standard Transformer \citep{lin2023evolutionary} or a specialised variant for processing multiple sequence alignments \citep{rao2021msa, jumper2021highly} to capture evolutionary relationships among homologous sequences.
    \item The structure module which initialises a fully connected graph with nodes representing the 3D positions of residues.
    Node representations are then updated via a Geometric GNN with invariant \citep{jumper2021highly} or equivariant message passing \citep{baek2021accurate, lee2023equifold} among all the residues to iteratively refine the predicted 3D structure. 
\end{enumerate}
Models are trained via minimizing a loss function that quantifies the difference between the predicted structure and the ground truth 3D structure.
The training data usually consists of experimentally determined 3D structures from the Protein Data Bank, so models essentially aim to find the most energetically favorable configuration among multiple possible conformational states \citep{lane2023protein}.

Note that, although closely related to generative modeling, structure prediction involves predicting only the 3D positions of a given input sequence.
On the other hand, generative modeling involves learning the distribution of a dataset of molecules and materials, and sampling new systems from the underlying distribution.
Interestingly, protein structure prediction models can be repurposed as generative models for protein design \citep{watson2023novo}.
This line of research has historical roots in physics-based approaches which attempted to build energy functions grounded in a biophysical understanding of protein folding \citep{alford2017rosetta}.


\begin{tcolorbox}[enhanced,attach boxed title to top left={yshift=-2mm,yshifttext=-1mm,xshift = 10mm},
colback=cyan!3!white,colframe=cyan!75!black,colbacktitle=white, coltitle=black, title=Opinion,fonttitle=\bfseries,
boxed title style={size=small,colframe=cyan!75!black} ]
It is interesting to note the contrast in how the input geometric graph is defined across the four tasks. Property prediction and dynamics simulation tend to use \textbf{local radial cutoff} graphs. This is presumably due to locality being a strong inductive bias, e.g. quantum mechanical properties or forces of a system are unlikely to be influenced by long-range interactions. On the other hand, generative modeling and structure prediction require models to develop globally coherent representations and generally necessitate operating on \textbf{fully connected} graphs.
An interesting exception to this observation is the random long-range graph construction scheme in Chroma \citep{ingraham2022illuminating}, a generative model for protein design which was validated in the wet lab.
\end{tcolorbox}


\textbf{Molecular Conformer Prediction. } In the case of molecular conformer prediction one aims to predict the 3D atomic positions from a molecular formula, usually given a molecular string representation \citep{weininger1988smiles,krenn2020self, cheng2023group}. Some of the most successful methods rely on predicting torsion angles from a molecular graph representation where equivariant GNNs serve as encoders for learning effective representations \citep{xu2022geodiff}.
Recently, \citet{wang2023generating} showed how a diffusion model with an unconstrained GNN denoiser which directly predict 3D coordinates can outperform strictly symmetric approaches for molecular conformation generation.

\textbf{Crystal Structure Prediction. } Similar to molecular structure prediction, crystal structure prediction involves the prediction of 3D crystal structures from two-dimensional crystal compositions. Early work by \citet{chen2022universal} and \citet{choudhary2021atomistic} applied Geometric GNNs specifically designed for predicting 3D crystal structures. Using a different approach, \citet{jiao2023crystal} applied an equivariant diffusion to predict 3D crystal structures from atomic compositions, taking into account the periodic structure of crystals as well as relevant symmetries. A detailed review of various ML methods, including Geometric GNNs, for crystal structure prediction, is given in \citet{riebesell2023matbench}. 

\textbf{Structure Prediction from Experimental Data. } 
An emerging application of geometric GNNs is in hybrid experimental and computational pipelines for molecular structure determination.
Notable examples include CryoEM protein structure determination \citep{jamali2023automated} and NMR chemical shift prediction \citep{guan2021real, yang2021predicting}.
Along similar lines, \citet{cheng2023reflectionequivariant} explored the use of equivariant diffusion models to predict 3D molecular structures based on incomplete information from real-world characterization instruments.

\subsection{Datasets}
\label{subsec:datasets}


In the realm of Geometric GNNs, the backbone of success lies in the availability and quality of the data. With numerous datasets proposed to date, it can be challenging to navigate through the landscape of options. 
Although it is not the primary focus on the paper, we provide a selected list of existing datasets for Geometric GNNs (see \Cref{tab:datasets-prop}). A more exhaustive list is accessible on a dedicated \href{https://github.com/AlexDuvalinho/geometric-gnns}{GitHub repository}, which we hope the community will keep up to date.

Our objective in this section goes beyond mere enumeration of available datasets. Since data construction is essential to improve model performance and to allow for true progress in subsequent years, we also discuss promising data directions for the field. By addressing these considerations, we aspire to foster the growth of the Geometric GNN community and empower researchers to build more powerful and robust models. Overall, we encourage practitioners to utilize (and construct) benchmark repositories containing several task-specific datasets, similarly to the Open Catalyst Project \citep{chanussot2021open, tran2023open}, Matbench \citep{dunn2020benchmarking} and the Open MatSci ML Toolkit \citep{lee2023matsciml, miret2023the} for materials research as well as to the Therapeutic Data Commons \citep{huang2022artificial} for drug discovery; all of which provide valuable ways to understand GNN model performance. 
Here are a few reasons why; 

\begin{enumerate}
    \item \textbf{Splits and Leaderboards}. These benchmarks are endowed with thorough and transparent evaluation protocols, including carefully defined train/val/test splits, a visible leaderboard with fixed evaluation metrics and some open-source guides describing how to use each dataset (with some baseline methods implemented). This enables easy utilisation as well as fair evaluation and comparison across methods, which is essential. Besides, the validation datasets contain both In-Domain (ID) and Out-of-Domain (OOD) split, allowing the assessment of model generalisation. 
    \item \textbf{Domain experts}. These datasets are constructed as the fruit of a collaboration between domain experts and the machine learning community, bridging the knowledge gap between scientific domains (e.g. physics, chemistry, materials science, biology) and machine learning communities. 
    This ensures that ML tasks are consistent with underlying practical applications and that the available datasets account for the subtleties of the application domain. 
    As the datasets continue to mature and the underlying problems become more and more complex, the range of domain experts should concurrently expand to include representation from government and corporations in addition to academic researchers. 
    
    \item \textbf{Continual dataset updates}. These datasets have shown regular updates and expansion through the collection or generation of new data, such as when enough new samples are obtained using DFT or experimental methods. The continual dataset expansions allow ML models to solve more complex and relevant challenges motivated by the underlying application. Ultimately the goal is to create more robust ML models, including both highly specialized models for a given application as well as general ML models that can tackle a wide range of modeling problems.
    \item \textbf{Diversity of tasks for generalized learning}. In well-maintained benchmarks, it is often possible to pre-train a model on a specific task and fine-tune it on another. As datasets grow, a greater diversity of tasks enables the community to build towards generalist foundation models for scientific applications in 3D atomic systems. Given the success of foundation models in the natural language and vision domains, these developments have the potential to unlock tremendous future research opportunities. 
\end{enumerate}

\textbf{Software and libraries}. On a slightly different note, one crucial factor for the applicability and development of Geometric GNNs to any dataset is the existence of handy code bases. We provide a list of useful repositories on \href{https://github.com/AlexDuvalinho/geometric-gnns/blob/main/coding-libraries.md}{github}, and encourage the community to maintain it up-to-date.

\begin{tcolorbox}[enhanced,attach boxed title to top left={yshift=-2mm,yshifttext=-1mm,xshift = 10mm},
colback=cyan!3!white,colframe=cyan!75!black,colbacktitle=white, coltitle=black, title=Opinion,fonttitle=\bfseries,
boxed title style={size=small,colframe=cyan!75!black} ]
\textbf{Navigating datasets} of 3D atomic systems is currently difficult due to major differences at every level: ground truth simulations, subsample selection, splits, benchmarks vs single dataset, etc. We advocate for additional work providing structure in this domain.
\end{tcolorbox}


\begin{landscape}

\begin{table}[h!]
    \centering
    \resizebox{1.6\textwidth}{!}{
        \begin{tabular}{llllllllll}
            \toprule
            Task                             & Dataset      & Benchmark & \# Samples & \# Tasks & Domain                    & Split           & Metric      & Date           & Source        \\
            \midrule
            \multirow{11}{*}{PP}         
                & Open MatSci ML Toolkit     & \href{https://github.com/IntelLabs/matsciml}{Open MatSci ML Toolkit}     & 1.5M       & Varied       & Crystal Structures                 & Stratified/Random  & MAE         & 2023* & [\href{https://arxiv.org/abs/2309.05934}{Paper}][\href{https://github.com/IntelLabs/matsciml}{github}]  \\
                  & QM7-b        & \href{https://moleculenet.org/}{MoleculeNet}  & 7k      & 14       & Molecule                 & Random          & MAE            & 2014 & [GDB-13][\href{http://quantum-machine.org/datasets/}{info}][\href{https://deepchem.readthedocs.io/_/downloads/en/2.4.0/pdf/}{info}] \\
                & QM9          & \href{https://moleculenet.org/}{MoleculeNet}  & 130-134k    & 12-19       & Molecule                 & Random          & MAE            & 2012  & [GDB-17][\href{http://quantum-machine.org/datasets/}{info}][\href{https://pytorch-geometric.readthedocs.io/en/latest/generated/torch_geometric.datasets.QM9.html}{PyG}] \\
                & PDBbind      & \href{https://moleculenet.org/}{MoleculeNet}  & 5k-13k  & 1 & Protein-ligand & Time           & RMSE           & 2004*& [\href{https://www.rcsb.org/search}{PDB}][\href{https://deepchem.readthedocs.io/_/downloads/en/2.4.0/pdf/}{info}] \\
                & Alchemy      & \href{https://alchemy.tencent.com/\#leaderboard}{Alchemy Contest}    & 120k    & 12       & Molecule                   & Stratified/Size\textsuperscript{\Cross} & MAE & 2019 & [\href{https://gdb.unibe.ch/downloads/}{GDB MedChem}]  [\href{https://alchemy.tencent.com/}{link}][\href{https://huggingface.co/graphs-datasets}{HF}] \\
                & Matbench     & \href{https://matbench.materialsproject.org/}{Mathbench}     & 1k-100k       & 13       & Crystal Structures                 & StratifiedKFold\textsuperscript{\Cross}  & MAE, ROC-AUC         & 2019 & [\href{https://next-gen.materialsproject.org/}{Materials Project}][\href{https://ml.materialsproject.org/}{data}]  \\   
                & Atom3D      & \href{https://www.atom3d.ai/}{Atom3D}     & Varied         & 8        & Mol., RNA, Prot.      & Varied               & Various              & 2021 & [Varied][\href{https://github.com/drorlab/atom3d}{github}]             \\ 
                & Jarvis & \href{https://pages.nist.gov/jarvis/databases/}{Nist-Jarvis}     & 1k-800k         & Varied        & Molecules      & Random\textsuperscript{\Cross} & MAE & 2020* & [\href{https://pages.nist.gov/jarvis/}{Documentation}][\href{https://www.nature.com/articles/s41524-020-00440-1}{Paper}] \\
                & Therapeutic Data Commons     & \href{https://tdcommons.ai/}{TDC}     & Varied       & Varied       & Molecules \& Proteins                 & Stratified\textsuperscript{\Cross}  & Varied         & 2022 & [\href{https://tdcommons.ai/}{Documentation}][\href{https://github.com/mims-harvard/TDC}{github}]  \\
                & TorchProtein     & \href{https://torchprotein.ai/benchmark}{TorchProtein}     & Varied       & Varied       & Proteins                 & Stratified\textsuperscript{\Cross}  & Varied         & 2022 & [\href{https://arxiv.org/abs/2206.02096}{Paper}][\href{https://github.com/DeepGraphLearning/PEER_Benchmark}{github}] \\
                & TorchDrug     & \href{https://torchdrug.ai/}{TorchDrug}     & Varied       & Varied       & Molecules                 & Stratified\textsuperscript{\Cross}  & Varied         & 2022 & [\href{https://arxiv.org/abs/2202.08320}{Paper}][\href{https://github.com/DeepGraphLearning/torchdrug/}{github}]
                \\

            \midrule
            \multirow{3}{*}{PP and MD}               & OC20         & \href{https://opencatalystproject.org/}{OpenCatalyst Project}   & 560k-133M       & 3        & Catalyst                  & Extrapolation\textsuperscript{\Cross}    & MAE, EwT         & 2020  & [\href{https://pubs.acs.org/doi/10.1021/acscatal.0c04525}{Paper}][\href{https://github.com/Open-Catalyst-Project/ocp}{github}]           \\
                    & OC22         & \href{https://opencatalystproject.org/}{OpenCatalyst Project}   & 50k-10M         & 3        & Catalyst                 & Extrapolation\textsuperscript{\Cross}    & MAE, EwT          & 2022 & [\href{https://pubs.acs.org/doi/10.1021/acscatal.2c05426}{Paper}][\href{https://github.com/Open-Catalyst-Project/ocp}{github}]           \\
                & ODAC23       & \href{https://opencatalystproject.org/}{OpenCatalyst Project}   & $\leq$40M         & 3        & Catalyst \& MOF                & Extrapolation\textsuperscript{\Cross}    & MAE, EwT          & 2023* & [\href{https://pubs.acs.org/doi/10.1021/acscatal.2c05426}{Paper}][\href{https://github.com/Open-Catalyst-Project/ocp}{github}]           \\


            \midrule                                    
            \multirow{2}{*}{MD}              
                    & MD17         & --    & 50k-1M         & 10        & Molecule                 & Extrapolation               & MAE                  & 2017* & [\href{https://www.science.org/doi/10.1126/sciadv.1603015}{Paper}]
                    [\href{https://huggingface.co/graphs-datasets}{HF}] 
                    [\href{https://pytorch-geometric.readthedocs.io/en/latest/generated/torch_geometric.datasets.MD17.html\#torch_geometric.datasets.MD17}{\detokenize{PyG}}] \\
                    & ISO17         & --    & 645K          & 1       & Molecule                 & Extrapolation          & MAE                & 2016 & 
                    [\href{http://www.quantum-machine.org/datasets/\#md-datasets}{info}]
                    [\href{https://www.nature.com/articles/s41467-018-06169-2}{Paper}]             \\
            \midrule 
            \multirow{3}{*}{Gen.Mod}                 & GEOM         & GEOM   & 37M         & 1        & Molecule             & Random               & MAE, RMSD                    & 2021 & [\href{https://www.nature.com/articles/s41597-022-01288-4}{Paper}][\href{https://github.com/learningmatter-mit/geom}{github}]            \\
            & Open MatSci ML Toolkit     & \href{https://github.com/IntelLabs/matsciml}{Open MatSci ML Toolkit}     & 25k       & 1       & Crystal Structures                 & Random  & MAE         & 2023* & [\href{https://arxiv.org/abs/2309.05934}{Paper}][\href{https://github.com/IntelLabs/matsciml}{github}]  \\
            & TorchDrug     & \href{https://torchdrug.ai/}{TorchDrug}     & Varied       & Varied       & Molecules                 & Stratified\textsuperscript{\Cross}  & MAE         & 2022* & [\href{https://arxiv.org/abs/2202.08320}{Paper}][\href{https://github.com/DeepGraphLearning/torchdrug/}{github}] \\
            \midrule
            \multirow{5}{*}{Struct.Pred} & TorchProtein     & \href{https://torchprotein.ai/benchmark}{TorchProtein}     & Varied       & 3       & Proteins                 & Stratified\textsuperscript{\Cross}  & Varied         & 2022* & [\href{https://arxiv.org/abs/2206.02096}{Paper}][\href{https://github.com/DeepGraphLearning/PEER_Benchmark}{github}] \\    
                & SPICE        & --   & 1.1 M        &  6       & Molecules            & Random   & MAE        & 2023 & [\href{https://www.nature.com/articles/s41597-022-01882-6}{Paper}][\href{https://github.com/openmm/spice-dataset}{github}]             \\
                & MatBench Discovery       & \href{https://matbench-discovery.materialsproject.org/models}{MatBench Discovery}   & 150k-250k        &  2       & Crystal Structures            & Varied\textsuperscript{\Cross}   & F1, Accuracy, MAE        & 2023 & [\href{https://matbench-discovery.materialsproject.org/preprint}{Paper}] [\href{https://github.com/openmm/spice-dataset}{github}]             \\
                & ProteinNet   & --   & 35k-105k          & 7        & Proteins                  & Stratified               & Varied                    & 2019 &  [\href{https://bmcbioinformatics.biomedcentral.com/articles/10.1186/s12859-019-2932-0}{Paper}][\href{https://github.com/aqlaboratory/proteinnet}{github}]            \\
                & Molecule3D   & --   & 3.9M  & 4        & Molecules           & Random & MAE, RMSE, validity  & 2021 & [\href{https://arxiv.org/pdf/2110.01717.pdf}{Paper}][\href{https://arxiv.org/pdf/2110.01717.pdf}{link}] \\
            \bottomrule
        \end{tabular}
    }
    \caption{Summary of benchmark datasets for Geometric GNNs. We categorize each dataset with respect to the application task detailed in \Cref{subsec:applications}. We display various properties (from left to right): the benchmark link, the number of samples per dataset, the number of properties which can be predicted, the input data domain, the dataset split method, the metric used to measure performance, the original date of release and the source. * signifies that the dataset has been updated recently. \textsuperscript{\Cross} means that there is an active leaderboard to submit test predictions and compare one's results.}
    \label{tab:datasets-prop} 
\end{table}

\end{landscape}

\clearpage

\section{Conclusion and Future Research Directions}
\label{sec:discussion}


This opinionated survey aims to provide both newcomers and experienced researchers with a pedagogical understanding of Geometric Graph Neural Networks for 3D atomic systems. 
Throughout its course, we delve into the foundations, motivations, and distinctive features that set Geometric GNNs apart from traditional GNNs. 
We hope that our taxonomy of Geometric GNN architectures -- invariant, equivariant with Cartesian basis, equivariant with spherical basis, and unconstrained GNNs -- establishes clear links between different models, granting the reader with a deeper comprehension of the available methods. In addition, we have summarised a wide range of applications and highlighted the high potential for impact of Geometric GNNs. 
We hope to inspire further developments in Geometric GNN modeling towards socially beneficial applications such as the discovery of novel medicine \citep{stokes2020deep}, energy-efficient materials \citep{miret2023the}, and green chemistry \citep{anastas2010green}. 

Our ultimate aspiration is to contribute to the organization and advancement of this emerging field while igniting the curiosity of newcomers to embark on their own journey into the captivating world of Geometric GNNs. 
By combining scholarly rigour with a comprehensive overview, we aspire to make this survey the natural go-to paper for the Geometric GNN community, facilitating knowledge dissemination and fostering future advancements.

As we conclude our survey, the following section reflects on the promising research directions that lie ahead and that we believe are worthy of interest. 

\subsection{To what extent should physics and symmetry be `baked in' to Geometric GNNs?}

\textbf{Enforcing symmetries}: The choice between invariant GNNs, equivariant GNNs, or unconstrained GNNs is an important consideration. 
Exploring the trade-off between (1) an unconstrained local message passing approach that (approximately) preserves global equivariance (e.g. FAENet \citep{duval2023faenet}), and 
(2) a strictly equivariant method that passes local equivariant messages between atoms (e.g. MACE \citep{batatia2022mace}), is an interesting open question. 
Discussing the implications of these design choices and rigorous empirical benchmarking would contribute to a deeper understanding of the trade-offs involved. 
%
For instance, rigorously enforcing symmetries can provide greater data efficiency and generalization abilities for model architectures, which is particularly interesting when getting more high-quality data is costly, or when stronger generalisation guarantees are needed. On the other hand, relaxing these constraints may be desirable if enough data is available, enabling greater expressivity and efficiency. \Cref{subsec:non-spm-gnns} contains additional arguments.

\textbf{Energy conservation}: 
On a related note, for molecular dynamics applications, the debate between predicting forces using the gradient of energy versus predicting forces directly from node representations is crucial. 
While the former improves stability when using Geometric GNNs to run dynamics simulations \citep{fu2023forces}, the latter offers memory, runtime, and sometimes performance gains \citep{gasteiger2021gemnet}. Considering the scalability vs. simulation stability trade-off can help practitioners decide on the appropriate approach for their datasets and tasks. 
Additionally, developing better metrics to quantify the importance of energy conservation would be valuable with initial proposals emerging in the literature \citep{bihani2023egraffbench}. Given the ambitions to scale molecular dynamics simulations to trillions of atoms at longer and longer timescales, research on both improving stability and compute efficiency at scale are needed.

\begin{tcolorbox}[enhanced,attach boxed title to top left={yshift=-2mm,yshifttext=-1mm,xshift = 10mm},
colback=cyan!3!white,colframe=cyan!75!black,colbacktitle=white, coltitle=black, title=Opinion,fonttitle=\bfseries,
boxed title style={size=small,colframe=cyan!75!black} ]
We postulate that \textbf{strict equivariance} may be critical for tasks related to precise geometric prediction, such as structure prediction or molecular simulation, where small errors may compound and lead to unstable or unphysical geometries if we use approximately symmetric models \citep{fu2023forces, bihani2023egraffbench}.
On the other hand, tasks akin to property prediction may benefit from approximately symmetric models which are optimised for a particular dataset, without the hard constraints that come with enforcing physical symmetries \citep{duval2023faenet}.
Similarly, these unconstrained models constitute a promising direction for applications where a unique canonical ordering and orientation of data is possible, such as for antibody proteins \citep{martinkus2023abdiffuser}.
\end{tcolorbox}

\textbf{Deeper theoretical characterisation}: 
A first step towards theoretical understanding of Geometric GNNs was based on their ability to solve the geometric graph isomorphism problem \citep{joshi2022expressive}\appref{app:sec:expressivity}, i.e. mapping unique geometric graphs to unique representations.
A deeper understanding of the expressive power, optimisation behaviour, and generalisation capacity of Geometric GNNs will complement a growing landscape of empirical work, while abstracting away domain-specific implementation details.
For instance, developing a provably universal, equivariant GNN on sparse graphs with finite tensor and body order remains an open question \citep{batatia2022mace}.
The emergence of new equivariant architectures based on Clifford algebras also presents opportunities for theoretical advancement and unifications \citep{brehmer2023geometric, ruhe2023clifford}. These efforts have the potential to further understand the generalization ability of equivariant models, which can help practitioners choose what models may be appropriate for their desired use case.

\begin{tcolorbox}[enhanced,attach boxed title to top left={yshift=-2mm,yshifttext=-1mm,xshift = 10mm},
colback=cyan!3!white,colframe=cyan!75!black,colbacktitle=white, coltitle=black, title=Opinion,fonttitle=\bfseries,
boxed title style={size=small,colframe=cyan!75!black} ]
While equivariant GNNs operating on \textbf{higher-order tensors} have favorable theoretical expressivity, their practical utility remains unclear.
Notably, prominent models such as RosettaFold \citep{baek2021accurate} and DiffDock \citep{corso2022diffdock}, which use the \texttt{e3nn} framework, \emph{do not} use higher-order tensors (they are restricted to tensor order $= 1$). 
In practice, we believe that the theoretical advantages of higher-order tensors are circumvented by the (currently) high GPU memory requirements and slow speed of tensor product operations.
Thus, theoretical studies must be supplemented by empirical benchmarks which fairly compare architectures under compute and time budgets \citep{jamasb2023evaluating} as well as domain-specific evaluation beyond empirical accuracy \citep{harris2023posecheck, buttenschoen2023posebusters}.
\end{tcolorbox}


\subsection{How to construct geometric graphs?}

\textbf{Graph creation and coarse-graining}. 
There are various approaches to constructing a geometric graph as input to a Geometric GNN, including radial cutoffs, $k$-nearest neighbours, and long range connections.
While optimal graph construction heuristics are often highly domain-specific, exploring how this construction could be modified to alleviate common structural bottlenecks for GNNs, such as the over-squashing problem \citep{di2023does, giraldo2023tradeoff}, may hold great promise. Further studies are also needed to better understand the importance of the local neighborhood in an atomic system compared to long-range interactions that could be effectively modeled with targeted graph constuction.

Additionally, the choice of entities included in graph construction needs careful consideration.
An obvious example is that all current Geometric GNN applications assume a coarse-grained implicit solvent system and do not explicitly include entities such as water molecules and ions which play an important role in molecular structure and function \citep{bellissent2016water}.

When working with coarse-grained representations of atomic systems, it becomes critical to analyse the need for atomic-level precision and completeness of the representation (whether there is a one-to-one mapping back to the all-atom scale) \citep{badaczewska2020computational}. 
Imagine analyzing a molecular system where each atom's position is recorded with ultra-high accuracy. This level of detail might be unnecessary for certain research objectives, and it may even introduce undue computational complexity or overfitting to artefacts from the structure determination process \citep{dauparas2022robust}. 
In such cases, it raises inquiries about the deliberate introduction of controlled noise to coordinate data. The key question becomes: should this noise be intentionally motivated by physical principles, and if so, what is the optimal amount of noise to inject? 

\begin{tcolorbox}[enhanced,attach boxed title to top left={yshift=-2mm,yshifttext=-1mm,xshift = 10mm},
colback=cyan!3!white,colframe=cyan!75!black,colbacktitle=white, coltitle=black, title=Opinion,fonttitle=\bfseries,
boxed title style={size=small,colframe=cyan!75!black} ]
We suspect that optimizing the way \textbf{information flows} from the perspective of graph machine learning may be promising for boosting performance beyond the physics-based perspective of radial cutoffs and interaction thresholds.
\end{tcolorbox}


\textbf{Temporal dynamics and conformational flexibility}:
Ideal computational representations of molecules and materials should account for both geometric structures as well as temporal dynamics.
Looking beyond learning from static structures, dynamics and conformational flexibility is key to the functionality of several classes of proteins important for drug discovery, such as antibodies and membrane proteins \citep{carugo2023structural, lane2023protein}, as well as the dynamic behavior of materials in diverse applications \citep{bihani2023egraffbench, fu2023forces}.
Learning from molecular conformational ensembles may be the next frontier for advancing geometric representation learning of small molecules \citep{axelrod2020molecular, zhu2024learning}, crystals \citep{lee2023matsciml, riebesell2023matbench}, proteins \citep{noe2019boltzmann, janson2023direct}, and RNA \citep{joshi2023multi}.


\subsection{How to scale up Geometric GNNs?}

\textbf{Foundation models}: 
Drawing inspiration from the success of self-supervised learning for large pre-trained foundation models \citep{bommasani2021opportunities}, the question of why we do not yet have Geometric GNNs which can generalise across the range of atomic systems is worth contemplating. 
Exploring new self-supervised learning tasks to understand the dynamics of general-purpose atomic interactions (so-called \emph{universal potentials}) could significantly enhance downstream performance across a wide range of application domains.
On the other hand, while it is pedagogical to group application domains together under the umbrella of `atomic systems', it may be the case that interactions governing small molecules are semantically different than those in proteins or materials.
While not taking a definitive stance, this question presents an interesting research direction for the field. 
What architectures are expressive and scalable, and how can we efficiently train them?

Recently, \citet{krishna2023generalized} and \citet{deepmind2023alphafold} generalised protein structure prediction models to full biological assemblies including proteins, small molecules, nucleic acids, and other ligands.
These works show promising results, even outperforming specialised models for certain tasks, and represent an exciting first step towards foundation models for structural biology.
Geometric GNNs specialised for biomolecular complexes and supramolecular systems \citep{steed2022supramolecular, gallego2022recent} may require deeper consideration of higher-order symmetries present in self-assembling biological systems.
For instance, naturally occurring DNA, perhaps the best-known self-assembling structure, exists in a double helical form.

\textbf{Large-scale datasets and engineering infrastructure}:
Training foundation models necessitates large datasets and associated software as well as hardware infrastructure. 
The availability of large-scale collections of predicted protein structures derived from both AlphaFold2 \citep{jumper2021highly} and ESMFold \citep{lin2023evolutionary} is an extremely promising data source towards this end.
The AlphaFold Database \citep{Varadi2021} and ESM Atlas (MGnify 2023 release \citep{Richardson2022}) contain over 200M and 772M predicted structures, respectively.
Bespoke software tools for predicted protein structures such as FoldSeek \citep{vanKempen2023} for efficient clustering and FoldComp \citep{Kim2023} for compression have started being used as part of pipelines to scale up Geometric GNNs for protein structure annotation \citep{jamasb2023evaluating}.

Similarly to the Open Catalyst Project \citep{zitnick2020introduction, tran2023open} and Open MatSci ML Toolkit \citep{miret2023the, lee2023matsciml} for materials discovery, the release of large public datasets and benchmarks targeted to specific use cases and specially curated for advancing deep learning architecture development are much needed.
Such efforts may necessitate rethinking the data generation process to be tailored towards training large deep learning models, as motivated by recent initiatives for small molecules \citep{mathiasen2023repurposing, beaini2023towards}.
We hope to see more community-driven and public initiatives in this direction, as motivated in \Cref{subsec:datasets}. 

Finally, it is worth mentioning the \emph{hardware lottery} \citep{hooker2021hardware} -- the marriage of architectures and hardware that determines which research ideas rise to prominence in the community.
At present, fully-connected message passing via dense matrix multiplications (i.e. Transformers \citep{joshi2020transformers}) is generally significantly faster on GPUs than sparse message passing via scatter-gather operations (i.e. GNNs).
At the same time, sparse graph processing tends to consume less GPU memory, enabling GNNs to scale to extremely large graphs up to millions of nodes, which is beyond the capabilities of standard Transformers.
Future hardware successors of GPUs specialised for graph structured data and GNN-style computation could have a significant impact on modelling 3D atomic systems and supramolecular complexes at scale.
A case-in-point which was previously highlighted are equivariant GNNs with higher rank tensor \citep{geiger2022e3nn}: these models are theoretically expressive which makes them ideal for large-scale pre-training. However, they are also practically challenging to scale due to high memory usage and slow tensor product operations.
Better software and hardware support for equivariant GNNs is needed to unlock their full potential.

\clearpage

{
    \bibliographystyle{abbrvnat}
    \bibliography{main}
}


\clearpage
\appendix
\section{Lexicon}
\label{app:lexicon}

In this section, we provide  background material on a variety of essential concepts to Geometric GNNs for 3D atomic systems. 

\subsection{Geometric vocabulary}
\label{app:sec:geom-voc}

\begin{enumerate}
    \item \textit{Scalars}: are quantities that do not have any direction or orientation and consequently remain unchanged under rotations or reflections. For example, the temperature or the energy of a system.
    \item \textit{(Geometric) vectors}: here we use the word vector to refer to an object that transforms with `simple' rotation matrices $\rot$, i.e. in the standard representation of $SO(3)$. In contrast to the general machine learning literature, where vectors are  simply one-dimensional lists of values, vectors in our usage of the word are objects that have both magnitude and direction. In the context of 3D atomic systems, a geometric vector can represent geometric attributes such as position, velocity, or forces acting on atoms. They are often expressed as \textit{Cartesian vectors}, that is, geometric vectors that are described within a Cartesian coordinate system (i.e. they have components along the coordinate axes $(x, y, z)$ in 3D space). They have the same dimension as the space they exist in.
    \item \textit{Geometric Tensors}: are mathematical objects that generalize geometric vectors to higher-dimensional spaces. In contrast to general machine learning, in which tensor just means a multi-dimensional list of numbers, we use the word tensor to describe objects that transform in a consistent way under under group actions (rotations in $SO(3)$ for us). Scalars and vectors are two special subclasses of tensors that are so common that we gave them their extra name. Objects that still transform consistently under rotations, but not invariantly (i.e. with the identity) or through simple rotation matrices are tensors. An example of a tensor would be the moment of inertia tensor in physics.
    Tensors are essential in many areas of physics, for example in differential geometry and general relativity, where they describe quantities in curved spacetimes. In the context of 3D atomic systems, geometric tensors can be used to represent higher-order geometric attributes or relationships. 
    For instance, the stress and strain\footnote{ The strain tensor in a material describes how a material's shape deforms when subjected to mechanical loads. In a 3D Cartesian coordinate system, the strain tensor has components corresponding to the changes in lengths and angles in different directions.} tensors in a material are second-order geometric tensors, capturing how the material deforms under external forces. 
    \begin{itemize}
        \item \textit{Cartesian Tensors}: are geometric tensors whose components aligned with the coordinate directions of a Cartesian coordinate system. These tensors are used to represent quantities that are directly related to the physical axes of the coordinate system.
        \item  \textit{Spherical Tensors}: are used to describe quantities that are invariant under rotations and have components aligned with spherical harmonics. They are particularly useful in problems involving spherical symmetry and rotational invariance.
    \end{itemize}
    \item The \textit{tensor order} (of a feature): refers to the way a Cartesian tensor transforms under rotation. It indicates the level of complexity or the number of components needed to fully describe the feature. In the context of Geometric GNNs, features are often represented as tensors, where each dimension corresponds to a specific aspect or property of the feature. The tensor order determines the number of indices needed to access and manipulate the feature's values. 
    \begin{itemize}
        \item Scalar features are zero-order tensors because they do not have any additional dimensions or indices.
        \item Geometric vectors are first-order tensors because they have one index to represent their components along different axes.
        \item Geometric tensors have tensor order equal or greater than two. The strain tensor, for instance, has a tensor order of two because it requires two indices to access its entries. Higher-order tensors (e.g. spherical tensors) can have a tensor order greater than 2 as they involve more complex structures with multiple indices. These tensors can represent more intricate features, such as higher-order interactions or relationships between elements in a system.
    \end{itemize}

    \item The \textit{type} of a tensor: is often used interchangeably with the \textit{tensor order}, but they can have slightly different meaning depending on the context. In particular, the type of a feature tells us how it changes under symmetry transformation, i.e.  how the tensor components change when the coordinate system or basis vectors are transformed. 
    %
    \item The \textit{body order} of a message, feature or model layer: refers to the number of atoms involved in the related computations. It is often used to describe the level of complexity (i.e. number of atoms considered) in a specific interaction model.  For example, a two-body message involves pairwise interactions between two atoms (e.g. distances) while a three-body message involves interactions among three atoms (e.g. bond angles). We often use \textit{many-body} interactions to describe interactions with more than 3-4 atoms \citep{batatia2022mace}.
    \item The \textit{frame of reference}, in the context of 3D atomic systems, refers to the specific spatial arrangement and atom ordering used to represent the system. Essentially, it is the viewpoint from which the researcher reads the atomic system.
\end{enumerate}

\goingfurther{
    \begin{itemize}[leftmargin=*,label=\ding{213}]
        \setlength\itemsep{1em}
        \item \href{https://www.grc.nasa.gov/www/k-12/Numbers/Math/documents/Tensors_TM2002211716.pdf}{An Introduction to Tensors for Students
of Physics and Engineering} (Kolecki, 2002) \resourcetag{article} \resourcetag{technical}
        \item \href{https://www.math3ma.com/blog/the-tensor-product-demystified}{The Tensor Product, Demystified} (math3ma, 2018) \resourcetag{blog} \resourcetag{visual}
        \item \href{https://jeremykun.com/2014/01/17/how-to-conquer-tensorphobia/}{How to Conquer Tensorphobia
} (Kun, 2014) \resourcetag{blog} \resourcetag{technical}
    \end{itemize}
}

\subsection{Group theory}
\label{app:sec:groups}

Group theory is a branch of mathematics that studies the symmetries and transformations of objects. In the context of 3D Geometric GNN models, group theory is particularly relevant because it helps us capture and exploit the symmetries present in atomic systems.

Indeed, atoms in a molecule or in a material exhibit specific spatial arrangements and undergo transformations such as rotations, reflections, and translations. These transformations preserve the overall structure and properties of the system. Group theory allows us to formally describe and analyze these transformations, enabling us to uncover hidden relationships and patterns.

Formally, a \textbf{group} $(\group{G}, \star)$ is a set of elements $\group{G}$ together with a binary operation $\star : \group{G} \times \group{G} \rightarrow \group{G}$ satisfying the following three conditions:
\begin{enumerate}
    \item \textit{Associativity}: $\forall \gel{g}_1, \gel{g}_2, \gel{g}_3 \in \group{G},$ we have $(\gel{g}_1 \star \gel{g}_2) \star \gel{g}_3 = \gel{g}_1 \star (\gel{g}_2 \star \gel{g}_3).$
    \item \textit{Identity}: there exists an identity element $\gel{e} \in \group{G}$ such that $\forall \gel{g} \in \group{G}$, we have $\gel{e} \star \gel{g} = \gel{g} \star \gel{e} = \gel{g}$
    \item \textit{Inverse}: each element has an inverse - that is, $\forall \gel{g} \in \group{G}$, $\exists \gel{h} \in \group{G}$ such that $\gel{g} \star \gel{h} = \gel{h} \star \gel{g} = \gel{e}$.
\end{enumerate}

We denote by $\vert \group{G} \vert$ the size of a group $\group{G}$, and call this the \textbf{order} of $\group{G}$. If $(\group{G}, \star)$ is a group and $\group{H} \in \group{G}$  is a subset such that $(\group{H}, \star)$ satisfies the above group axioms, then we call $\group{H}$ a \textbf{subgroup} of $\group{G}$, which we write as $\group{H} \leq \group{G}$.

\begin{figure}
    \centering
    \includegraphics[width=0.7\linewidth]{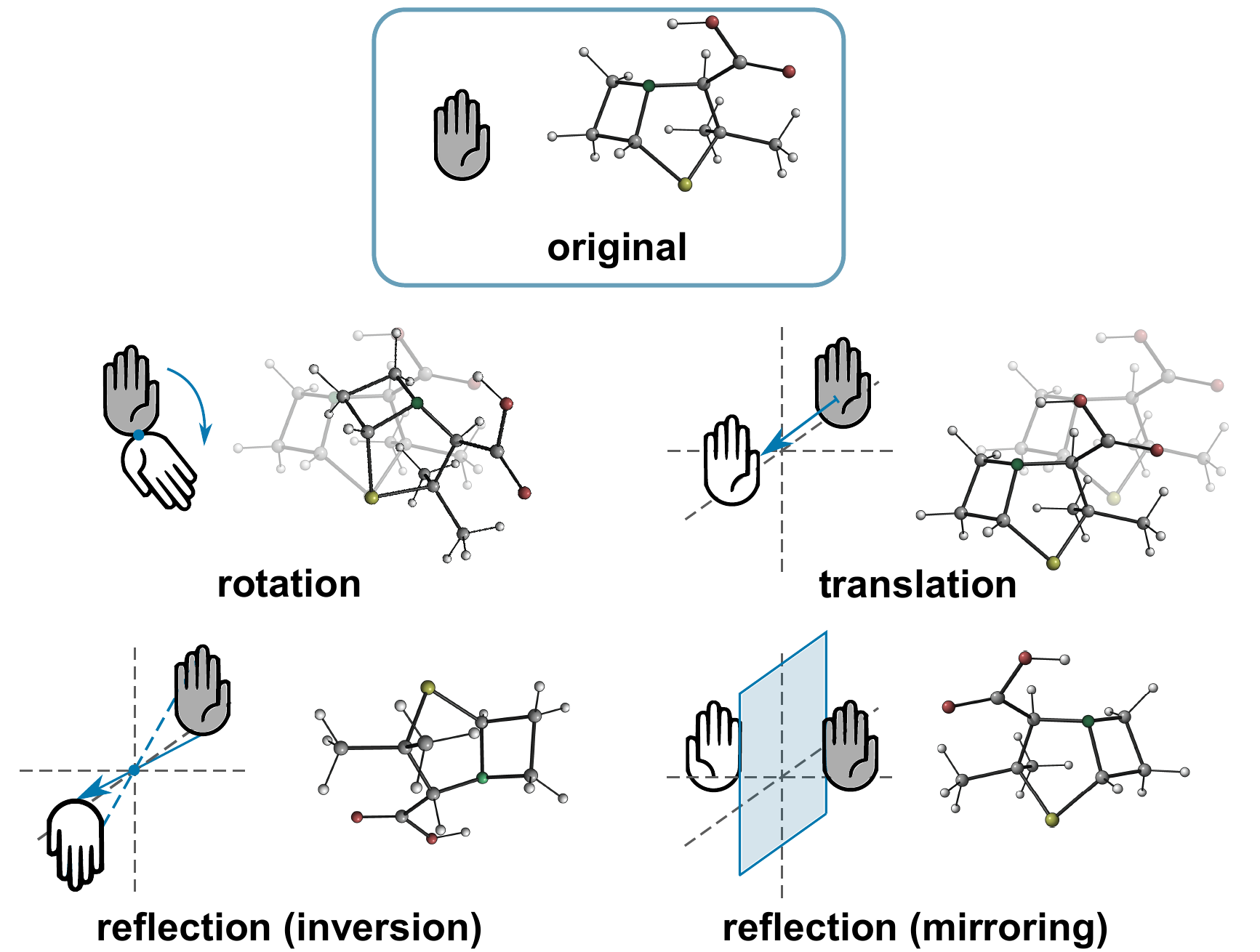}
    \caption{Illustration of the different euclidean symmetries for 3D atomic systems. Source: \citep{atz2021geometric}.}
    \label{fig:symmetries}
\end{figure}

Now that we have seen what a group is, let's see the \textbf{different symmetry groups of interest} for 3D atomic systems, illustrated in \Cref{fig:symmetries}. 
\begin{enumerate}
    \item $\text{E}(d)$: the Euclidean group includes translations, rotations, and reflections. It represents all possible transformations in d-dimensional Euclidean space.
    \item $\text{O}(d)$: the orthogonal group represents rotations and reflections. $\text{O}(d) \leq \text{E}(d)$.
    \item $\text{SO}(d)$: the Special Orthogonal group consists of rotations without reflections, in the d-dimensional space. $\text{SO}(d) \leq \text{O}(d)$.
    \item $\text{SE}(d)$: the Special Euclidean group combines translations and rotations. $\text{SE}(d) \leq \text{E}(d)$.
    \item $\text{T}(d)$: the Translation group represents the symmetry transformations of pure spatial translations in d-dimensions, without any rotation or reflection. $\text{T}(d) \leq \text{SE}(d)$.
    \item $\text{S}_n$: the symmetric group is the group of all permutations of the set of $n$ atoms.
\end{enumerate}

For example, for a rotation angle $\theta$, a rotation matrix around the z-axis around would be written:
\[
R = \begin{bmatrix}
\cos(\theta) & -\sin(\theta) & 0 \\
\sin(\theta) & \cos(\theta) & 0 \\
0 & 0 & 1 \\
\end{bmatrix} \in \text{SO}(3)
\]

\begin{tcolorbox}[enhanced,attach boxed title to top left={yshift=-2mm,yshifttext=-1mm,xshift = 10mm},
colback=cyan!3!white,colframe=cyan!75!black,colbacktitle=white, coltitle=black, title=What it means for Geometric GNNs?,fonttitle=\bfseries,
boxed title style={size=small,colframe=cyan!75!black} ]
Typically, we would like Geometric GNNs to exhibit $\text{E}(3)$ or $\text{SE}(3)$-equivariance. While rotations and translations do not pose a concern, reflections do. In isolation, a molecule and its mirror image share the same internal features and properties, regardless of its chirality. Since ML datasets often showcase molecules in isolation, E$(3)$-equivariance is desirable. However, molecular functionality is most often conferred by intermolecular interactions with surrounding components, meaning that a molecule's properties may differ from those of its mirror image. In such cases, we no longer require equivariance to reflections, making $\text{SE}(3)$-equivariance desirable.
\end{tcolorbox}

\subsection{Data structures}
\label{app:sec:data-structures}

All data structures defined below: molecules, proteins and material, are encapsulated under the term ``atomic systems``. Alternatively, we refer to them as ``molecules and materials``. 

\begin{figure}[h!]
    \centering
    \begin{subfigure}[b]{0.25\textwidth}
        \includegraphics[width=\textwidth]{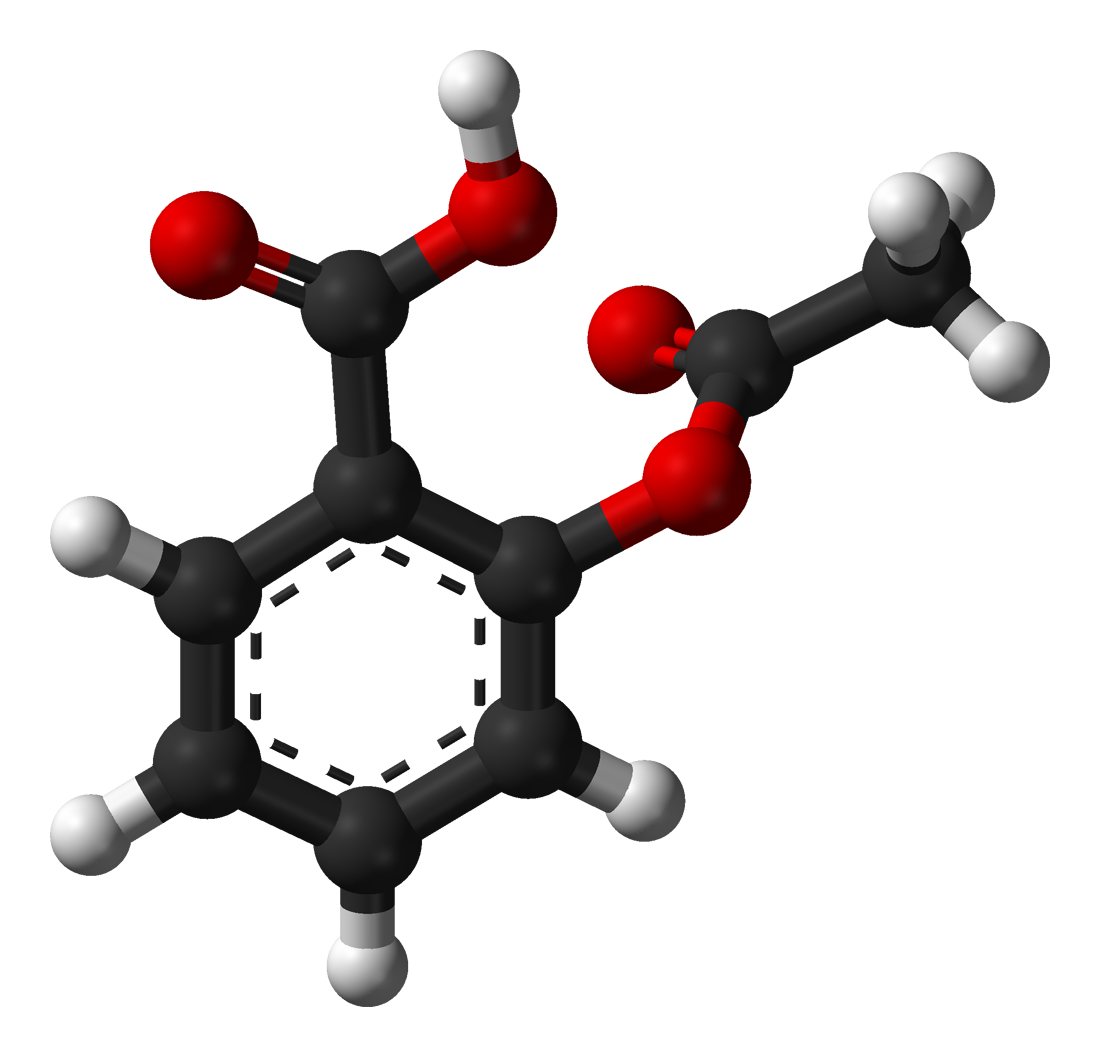}
        \caption{Small molecule}
    \end{subfigure}
    \quad
    \begin{subfigure}[b]{0.27\textwidth}
        \includegraphics[width=\textwidth]{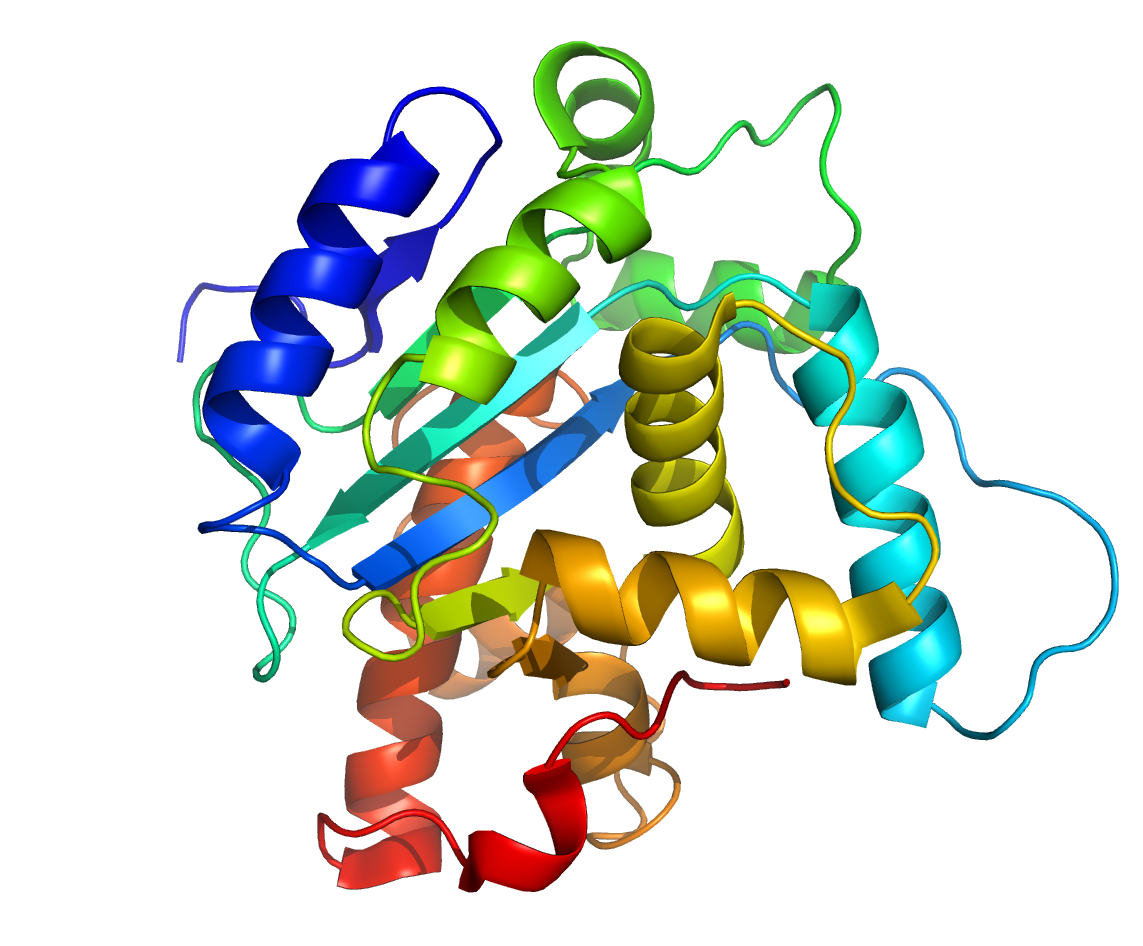}
        \caption{Protein}
    \end{subfigure}
    \quad
    \begin{subfigure}[b]{0.23\textwidth}
        \includegraphics[width=\textwidth]{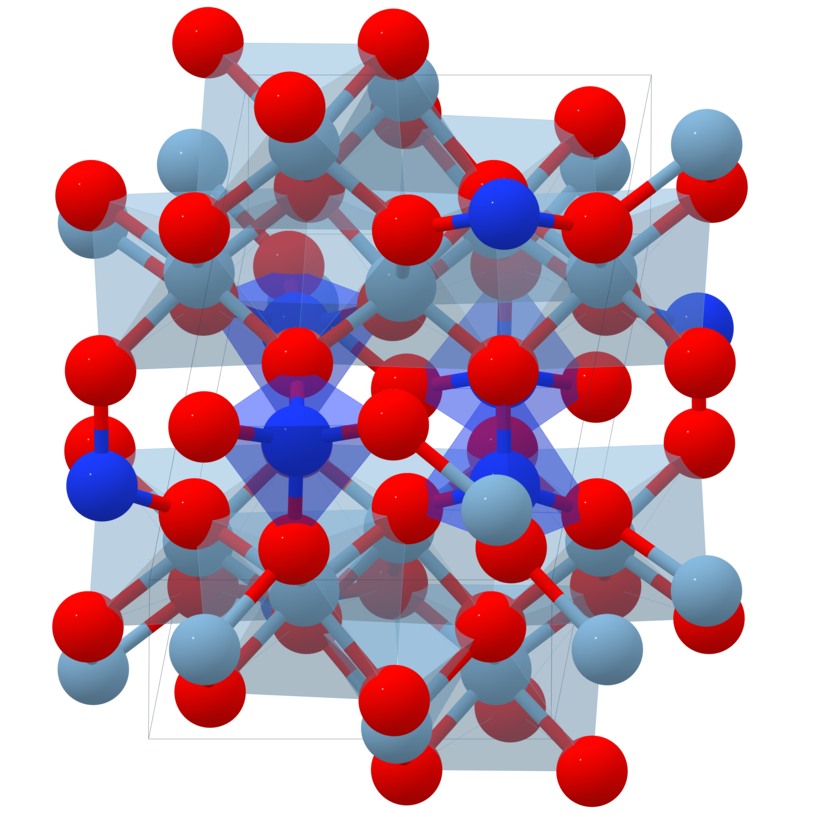}
        \caption{Material}
    \end{subfigure}
    \caption{Examples of different data structures.}
\end{figure}

\subsubsection{(Small) Molecules}
\label{app:subsec:mol}

A molecule is a group of atoms bonded together in a non-periodic manner, forming the basic unit of a substance. Molecules can exist in different forms like gases, liquids, or solids. Atoms, which are the building blocks of molecules, combine in specific ways to create molecules with unique structures and properties. Scientists analyze molecular structures to understand how they interact, participate in chemical reactions, and contribute to the properties of substances. This knowledge leads to advancements in various domains including medicine,, technology, environmental science, etc.

\begin{figure}[h!]
    \centering
    \begin{subfigure}[T]{0.43\textwidth}
        \includegraphics[width=\textwidth]{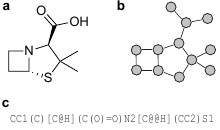}
    \end{subfigure}
    \begin{subfigure}[T]{0.51\textwidth}
        \includegraphics[width=\textwidth]{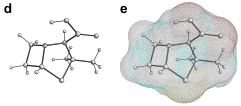}
    \end{subfigure}
    \caption{Exemplary molecular representations for a targeted molecule (i.e., the penam substructure of penicillin). 
    \textbf{a.} 2D Kekulé structure. \textbf{b.} 2D molecular graph. \textbf{c.} SMILES string \cite{weininger1988smiles}, in which atom type, bond type and connectivity are specified by alphanumerical characters. \textbf{d.} 3D graph with atom positions. \textbf{e.} Molecular surface represented as a mesh colored according to the respective atom types. Source: \citep{atz2021geometric}.}
    \label{app:fig:all-mol-rep}
\end{figure}

In the field of machine learning and computational chemistry, molecules are typically described using different representations (see \Cref{app:fig:all-mol-rep}). Traditional approaches have focused on 1D descriptions such as molecular fingerprints or SMILES strings, and 2D topology graphs, where atoms and bonds are represented by nodes and edges, respectively. To construct meaningful representations that capture the molecule's topology, scientists have opted for Graph Neural Networks on 2D graphs due to their ability to efficiently account for atomic interactions through message passing mechanisms.

Going beyond 2D graphs, there has been an increasing recognition of the importance of considering the 3D geometric conformations\footnote{conformations represent the 3D structure of an atomic system: the specific arrangement of atoms, their bond lengths, bond angles, and torsion angles. Different conformations often impact molecular properties.} of molecules for property prediction tasks. This highlights the necessity of leveraging 3D structures to enhance the understanding and prediction of molecular properties. 

\goingfurther{
    \begin{itemize}[leftmargin=*,label=\ding{213}]
        \setlength\itemsep{1em}
        \item \href{https://onlinelibrary.wiley.com/doi/full/10.1002/qua.26870}{Molecular representations for machine learning applications in chemistry} (Raghunathan, 2021) \resourcetag{article} \resourcetag{visual} \resourcetag{technical}
    \end{itemize}
}

\subsubsection{Proteins}
\label{app:subsec:prot}

A protein is a large molecule that plays a crucial role in various biological processes in living organisms. Protein sequences are made up of smaller units called amino acids, which are connected together in a specific order to form a long chain called a polypeptide. 

Proteins have a unique 3D structure that is essential for their function. This structure can be divided into different levels. The primary structure refers to the linear sequence of amino acids in the polypeptide chain. The secondary structure describes the local folding patterns that arise from interactions between neighboring amino acids. Examples of secondary structures include helices and sheets. The tertiary structure represents the overall 3D arrangement of the protein, including its folds and twists. Finally, in some cases, multiple protein chains can come together to form a complex known as the quaternary structure.

Scientists can use ML to modelize and analyse the 3D structure of proteins. Doing so, they can gain insights on how it interacts with other molecules and performs specific tasks in the body. This understanding is crucial for various fields including medicine, buichemistry and drug discovery. 


\goingfurther{
    \begin{itemize}[leftmargin=*,label=\ding{213}]
        \setlength\itemsep{1em}
        \item \href{https://www.sciencedirect.com/science/article/pii/S2001037022004044}{Protein–protein interaction prediction with deep learning: A comprehensive review} (Soleymani, 2022) \resourcetag{article}
        \item Paper list \href{https://github.com/lirongwu/awesome-protein-representation-learning}{github.com/lirongwu/awesome-protein-representation-learning}
    \end{itemize}
}

\subsubsection{Solid-State Materials}
\label{app:subsec:crystals}


Solid-state materials are characterised by their composition and structure leading to unique properties making them amendable to a diverse set of applications. Solid-state materials lose their fixed structure when they transition to liquid or gas phases, which have less restrictions on what spaces atoms can occupy. The solid, or condensed matter, state lends itself to a a fixed structure and definite volume. Solids exhibiting a very regular, periodic structure are referred to as crystals while solids with no such positional large scale order are called amorphous.


\textit{Crystals}, also known as crystalline materials, are solid substances characterized by a periodic arrangement of atoms and molecules in all spatial directions. A crystal structure includes atom types and positions as well as the translational axes along which the structure repeats. Since crystal structures extend infinitely, scientists define the smallest part needed to fully define the crystal as the \textit{unit cell}. This unit cell describes the crystal structure and composition, as well as how the crystal is repeated in space.  
This repeated ordered structure gives crystals unique shapes and properties, e.g. transparency or high melting points, making it crucial for the development of new materials with customised properties. As a result, ML methods often attempt to harness this periodicity to be as accurate as possible in their predictions.



\goingfurther{
    \begin{itemize}[leftmargin=*,label=\ding{213}]
        \setlength\itemsep{1em}
        \item \href{https://iopscience.iop.org/article/10.1088/2632-2153/aca23d}{Self-supervised learning of materials concepts from crystal structures via deep neural networks} (Suzuki, 2022) \resourcetag{article} \resourcetag{technical}
        \item \href{https://proceedings.neurips.cc/paper_files/paper/2022/file/1abed6ee581b9ceb4e2ddf37822c7fcb-Paper-Conference.pdf}{Equivariant Networks for Crystal Structures} (Kaba, 2023) \resourcetag{article}
    \end{itemize}
}

\subsection{Periodic boundary conditions}
\label{app:sec:pbc}

Periodic boundary conditions (pbc) are commonly used to represent the infinite repeating nature of crystals and other periodic systems. Instead of considering just a single 3D unit cell where all atoms would lie, one imagine an infinite array of identical unit cells extending in all directions.

To incorporate this periodicity into simulations, one imposes pbc on the simulation cell. This means that if an atom or molecule exits the cell on one side, it re-enters on the opposite side as if it were moving through a periodic lattice. By doing this, scientists can simulate the behavior of an infinite crystal using a finite computational domain.

In practice, we implement pbc by defining a simulation cell $\cgv{c}$ and a set of cell offsets $\cgv{o}_{ij} = [a, b, c]$ where $a,b,c \in \{0,1,-1\}$ are associated with each edge $(i,j)$ of the graph to encode edges across neighboring unit cells. These cell offsets thus help determine distances between pairs of atoms while considering the periodic repetition of the crystal lattice. Distances under pbc are written $d_{ij} = || (\mathbf{x}_i - \mathbf{x}_j) + \gv{o}_{ij} \cdot \cgv{c}||$. \Cref{fig:pbc} provides a visual illustration. 

In different terms, when two atoms are adjacent across a cell boundary, we calculate their distance by accounting for the periodicity of the lattice. For instance, if the cells are on top of each other (wrt z-axis), the cell offset vector would be $[0, 0, -1]$ or $[0, 0, 1]$. By considering these pbc, we can more accurately model the behavior of crystals and other periodic systems.

Note that, in the context of PBCs, the Euclidean group actions must be extended to account for rotations not only within the primary unit cell but also across its periodic replicas. This means that the Geometric GNN must also exhibit equivariance to group actions that extend across neighboring unit cells, allowing for a robust representation of euclidean symmetries in the presence of periodicity.

    \goingfurther{
        \begin{itemize}[leftmargin=*,label=\ding{213}]
            \setlength\itemsep{1em}
            \item \href{https://python.plainenglish.io/molecular-dynamics-periodic-boundary-conditions-21f957bbb294}{Molecular Dynamics: Periodic Boundary Conditions} (Mcelfresh, 2020) \resourcetag{blog} \resourcetag{visual}
            \item \href{https://computecanada.github.io/molmodsim-md-theory-lesson-novice/04-Periodic_Boundary/index.html}{Periodic Boundary Conditions
} (Digital Research Alliance of Canada, 2023) \resourcetag{blog}\resourcetag{visual}
        \end{itemize}
    }




\subsection{Quantum chemistry background}
\label{app:sec:quantum-chemistry}

Quantum chemistry serves as the foundation for understanding the behavior of atoms and molecules at the atomic scale. It provides insights into the interactions of atoms and electron leading to a better understanding of electronic structures, molecular interactions, and the fundamental mechanisms governing chemical processes. This section outlines some key concepts that form the basis for machine learning projects applied to physics and chemistry.

In quantum chemistry, electrons govern how atoms form interatomic (i.e. between atoms) bonds and thereby interact with each other. Their behaviour is modeled using \textit{wave functions}, meaning, by probability distributions expressing the likelihood of finding an electron at a specific location around an atom. Having knowledge about electrons' wave functions is essential because they hold the key to understanding how atoms in a given structure behave, therefore unlocking insights into chemical reactions, material properties and behaviors that are critical to understanding current materials systems and potential new designs.

The \textit{Schrödinger equation} is the fundamental equation in quantum mechanics that governs the wavefunctions' behavior. The equation incorporates a Hamiltonian operator, which represents the total energy of the system. Solving the Schrödinger equation for a given system yields the wavefunctions and corresponding energy levels for its particles. However, it is mathematically intractable beyond the Hydrogen atom.

The \textit{Density Functional Theory} (DFT) is a computational approach which determines the electronic structure of molecules and solids by approximating electron density, i.e. the Schrödinger equation using diverse sets of functionals for different atomic systems. While computationally tractable for many systems, running DFT remains computationally expensive and becomes impractical for intricate systems, we propose the use ML to efficiently approximate DFT calculations. ML models hold the potential to significantly reduce computational costs while maintaining accuracy, making them invaluable tools for researchers. 

For instance, ML models that efficiently and accurately approximate DFT-based energy calculations could be used to model the relationship between the potential energy of a molecule and its nuclear coordinates, called the \textit{potential energy surface} (PES). The PES provides valuable insights into molecular stability, chemical reactions, and the geometry of molecules. Analyzing it helps us understand how molecules interact, react, and transition between different energy states.

\begin{tcolorbox}[enhanced,attach boxed title to top left={yshift=-2mm,yshifttext=-1mm,xshift = 10mm},
colback=cyan!3!white,colframe=cyan!75!black,colbacktitle=white, coltitle=black, title=Opinion,fonttitle=\bfseries,
boxed title style={size=small,colframe=cyan!75!black} ]
While a deep understanding of quantum chemistry is not ``required'' to use and develop Geometric GNNs, being familiar with basic quantum chemistry principles may enhance your comprehension of the field as well as your ability to design more meaningful models. Besides, running DFT (or any other quantum chemistry numerical method) remains essential to construct bigger and more versatile databases to train machine learning models on.
\end{tcolorbox}



    \goingfurther{
        \begin{itemize}[leftmargin=*,label=\ding{213}]
            \setlength\itemsep{1em}
            \item \href{https://youtu.be/vHny-vpg57c?feature=shared}{How do we model atoms?} \resourcetag{video}
        \end{itemize}
    }

\subsection{Energy conservation}
\label{app:sec:energy-conservation}

The potential energy of a system represents the energy stored within it. It arises from the interplay of attractive and repulsive forces between atoms, which are determined by factors such as atom types and atom positions. The potential energy represents the energy that can be released when these components undergo positional changes, determining the stability and behaviour of the system. When the force field is conservative, the forces experienced by the atoms can be derived from the potential energy by taking the derivative with respect to their positions,
\begin{equation}
    E = - \int F(\gv{x})d\gv{x}, \quad \text{ and } \quad F = - \nabla E(\gv{x})
\end{equation}
By definition, these forces will always tend to minimize the potential energy, driving the system towards a relaxed state, analogous to gradient descent optimization.

In ML models for chemical systems, a single neural network is used to predict the total energy of the system by summing the contributions from individual atoms. To maintain energy conservation, the model calculates the forces on each atom by computing the derivative of the final predicted energy with respect to the atom's position. This allows for accurate predictions of both energy and forces, enabling efficient exploration and analysis of chemical systems.

In practice, this energy conservation requirement is also incorporated into the model's loss function, which includes a new term calculating the difference between predicted forces (obtained through backpropagation of the predicted energy) and ground truth forces (energy conserving). By considering the forces, the model adheres to the fundamental principle of energy conservation and constrains the space of functions explored by the neural network.

However, it is worth noting that this energy conserving requirement may hamper model performance. In certain cases, breaking free from this constraint may be beneficial. For example, in the Open Catalyst Project \citep{chanussot2021open}, non-energy conserving models\footnote{These models predict forces from final atom representations using a separate neural network and include a loss term about force predictions.} outperformed energy-conserving ones \citep{gasteiger2021gemnet, shuaibi2021rotation, duval2022phast}. Whether to strictly enforce the energy conservation principle remains an active area of research. Furthermore, energy conservation itself does not necessarily yield accurate approximation of forces given that modeling function does not guarantee effectively modeling its gradient. As such further research is needed to explore the effects of energy and force conservation in atomic systems.

\subsection{GNN architectural details}
\label{app:sec:gnn-architectural-details}

\subsubsection{Message Passing}
\label{app:subsec:message-passing}

Message passing is often used to describe the functioning of the family of Graph Neural Networks (GNN) models. Why? Because updating the representation of each node $i$ can be seen as passing a message from neighbouring nodes ($j \in \mathcal{N}_i$) to the node of interest. In its simplest form, the message $\cs{m}_{ij}$ is computed via a learnable message function $\l{f_1}$ of the neighbour's representation $\vs_j$ and $\vs_i$. The updated representation at layer (t+1) is obtained by applying a learnable update function $\l{f_2}$ on the aggregated messages $\oplus_{j \in \nei_i} \cs{m}_{ij}$ coming from neighbouring nodes and the existing representation $\cs{s}_i$. $\oplus$ is not learnable and often denotes the sum or mean operator.
\begin{align*}
    \cs{m}_{ij} &= \l{f_1}(\cs{s}_i^{(t)}, \cs{s}_j^{(t)}) \\
    \cs{s}_i^{(t+1)} &= \l{f_2}(\cs{s}_i^{(t)}, \oplus_{j \in \nei_i} \cs{m}_{ij})
\end{align*}

Message Passing has proven very successful so far but it also comes with its set of own limitations. Among them we find \textit{over-smoothing}, where node features become too similar after multiple message passing layers, loosing discriminative power due to excessive aggregation, and \textit{over-squashing}, which denotes an excessive information compression through bottleneck edges, likely leading to information loss \citep{giraldo2023tradeoff}.

\goingfurther{
\begin{itemize}[leftmargin=*,label=\ding{213}]
    \setlength\itemsep{1em}
    \item \href{https://distill.pub/2021/gnn-intro/\#passing-messages-between-parts-of-the-graph}{A Gentle Introduction to Graph Neural Networks} (Distil, 2021) \resourcetag{blog} \resourcetag{visual}
    \item \href{https://www.cs.mcgill.ca/~wlh/grl_book/}{Graph Representation Learning Book} (Hamilton, 2020) \resourcetag{book} 
\end{itemize}
}

\subsubsection{Activation function}
\label{app:subsec:activation-function}

As for any deep learning model, the choice of non-linear activation function plays a central role in modeling complex non-linearities of atomic interactions. The Rectified Linear Unit (\textbf{ReLU}) activation~\citep{glorot2011deep} is widely used in many deep learning models~\citep{sanchez2020learning,bapst2020unveiling}.
However, given the inherent nature of forces, ReLU may not be ideal to model atomic forces because its output is modeled as  piece-wise linear hyperplanes with sharp boundaries. As a result, a wide array of alternatives have been explored: Tanh, Leaky-ReLU \citep{maas2013rectifier}, SoftPlus~\citep{dugas2001incorporating}, Shifted SoftPlus~\citep{schutt2018schnet}, and Swish~\citep{ramachandran2017swish}.
 
 While the choice of activation function ultimately depends on the specific problem domain and dataset characteristics, Geometric GNNs most commonly use the \textbf{Swish} activation ${\rm swish}(x) = x\cdot {\rm sigmoid}(x)$ as it offers several advantages. It provides a smoother output landscape and has non-zero activation for negative inputs. This smoothness is essential when dealing with atomic systems, ensuring that small perturbations in the input space result in gradual changes in the output space, which promotes stability and avoids abrupt changes in predictions. Additionally, Swish has a smooth gradient, making backpropagation more stable and efficient, mitigating issues like vanishing or exploding gradients in deep networks. Finally, its non-linear behavior allows Geometric GNNs to model complex molecular interactions and spatial relationships effectively.

\begin{tcolorbox}[enhanced,attach boxed title to top left={yshift=-2mm,yshifttext=-1mm,xshift = 10mm},
colback=cyan!3!white,colframe=cyan!75!black,colbacktitle=white, coltitle=black, title=Opinion,fonttitle=\bfseries,
boxed title style={size=small,colframe=cyan!75!black} ]
The swish activation works well but the community should not stop looking for better alternatives, endowed with similar desirable properties. 
\end{tcolorbox}

\subsubsection{Basis functions}
\label{app:subsec:basis-functions}

In Geometric GNNs, the choice of basis function $\basis: \reals^d \to \reals^a$ is essential to transform geometric information, i.e. atom relative positions $\gv{x}_{ij}$, into discriminative representations. In different words, basis functions encode the spatial relationship between atoms, enabling Geometric GNNs to effectively model the structural properties of the system.
The dimension $a$ of the encoded geometric information is a hyperparameter depending on the choice of the basis function. It is usually chosen to be significantly larger than the default dimensionality to capture fine-grained distinctions in atom positions. Below, we describe the most widespread basis functions:

\begin{enumerate}
    
\item {\bf Identity:} $\basis_{\rm id}(\gv{x}_{ij}) = \cs{x}_{ij}$. The standard basis function uses the ``scalarised'' geometric information for each edge $(i,j)$, basically keeping the same values but loosing its geometric aspect ($\gv{x} \in \reals^d$ to $\cs{x} \in \reals^d$).  

\item {\bf Radial Basis Function} (RBF): $\basis_{\rm rbf}(\gv{x}_{ij}) = [\basis_1,\ldots,\basis_a]$, where $\basis_k$ is the output of the $k$-th basis function $\basis_k( \gv{x}_{ij}) = {\rm e}^{ \left( \Vert\gv{x}_{ij}\Vert - \mu_k \right)^2 / (2 \cdot \sigma^2)}$ representing the distance between atoms $(i,j)$, encoded using Gaussian functions that decay with distance. By considering pairwise distances, the RBF basis function can encode the varying degrees of influence that atoms exert on each other based on their spatial proximity. The Gaussian means are evenly distributed on $[0,1]$, i.e., $\mu_k = k / (a-1)$ and the standard deviation is $\sigma = 1/(a-1)$. Values of $\gv{x}_{ij}$ are often normalized to lie in $[0,1]$. $\basis_{\rm rbf}$ results in a $a$-dimensional vector.

\item {\bf Sine:} $\basis_{\rm sin}(\hat{x}_{ij}) = [\basis_1,\ldots,\basis_b]$, where $\basis_k$ is the output of the $k$-th basis function $\basis_k(\hat{x}_{ij}) = \sin(1.1^k \hat{x}_{ij})$. This design is based on function approximation using the Fourier series.  $\basis_{\rm sin}$ is applied to each dimension of the unit vector $\hat{x}_{ij}$, resulting in an $a = b \times d$ vector. 


\item {\bf Spherical Harmonics:} $\basis_{\rm sph}(\hat{x}_{ij}) = [\cgv{Y}^{(0)}(\theta, \phi), \ldots, \cgv{Y}^{(L)}(\theta, \phi)]$, where the polar and azimuthal angles $\theta$ and $\phi$ are directly computed from the unit directional vector $\hat{x}_{ij} \in \R^3, \Vert \gv{x}_{ij} \Vert = 1$. The vector-valued function $\cgv{Y}^{(l)}: S^2 \rightarrow \reals^{(2l+1)}$ is the list of Laplace's spherical harmonics
$\gv{Y}^{(l)}_m : S^2 \rightarrow \reals$ with order $l \geq 0$ and degree $m \in \{-l, \ldots, l\}$~\citep{macrobert1947spherical}.  The spherical harmonics are special function which encode angular information from the surface of the sphere, forming an orthonormal basis for Fourier transformations of functions on the sphere, like sine waves on $\reals$. They are equivariant in SO$(3)$, transforming in a predictable manner when the input is rotated by $\rot \in \text{SO}(3)$:
\begin{equation*}
    Y^{(l)}_m (\rot \cdot \hat{x}_{ij}) = \sum_{m'} \gv{D}^{(l)}_{mm'}(\rot) \gv{Y}^{(l)}_{m'}(\hat{x}_{ij})
\end{equation*}
where $\gv{D}^{(l)}_{mm'}$ are the entries of the Wigner-D matrix $\cgv{D}^{(l)} \in \reals^{(2l+1)\times (2l+1)}$. For Geometric GNNs, spherical harmonics are often used as convolutional filter (or simple input feature) as they enable to capture the spatial orientation and angular relationships between atoms.
The list of spherical harmonics is displayed  
\href{https://en.wikipedia.org/wiki/Table_of_spherical_harmonics#Real_spherical_harmonics}{here} and an implementation is available \href{https://docs.e3nn.org/en/latest/api/o3/o3_sh.html#e3nn.o3.spherical_harmonics}{here}.

\item {\bf MLP:} $\basis_{\rm MLP}(\gv{x}_{ij}) = \sigma (\l{\bm{W}}\cdot \gv{x}_{ij} + \l{\bm{b}})$, where $\sigma(\cdot)$ is a non-linear activation function, and $\l{\bm{W}}$ and $\l{\bm{b}}$ are learnable parameters. This method is flexible because it can be applied to any geometric quantity (atom relative position, concatenation of scalar distance and angles, \citep{duval2023faenet}); and powerful because two-layers MLP with enough hidden layers are universal approximators. However, the MLP basis function cannot preserve equivariance due to the presence of non-linearities. 

\end{enumerate}

\goingfurther{
    \begin{itemize}[leftmargin=*,label=\ding{213}]
        \setlength\itemsep{1em}
        \item \href{https://medium.com/dataseries/radial-basis-functions-rbf-kernels-rbf-networks-explained-simply-35b246c4b76c}{Radial Basis Functions, RBF Kernels, \& RBF Networks Explained Simply} (Ye, 2020) \resourcetag{blog} \resourcetag{visual}
        \item \href{https://papertalk.org/papertalks/35863}{Achieving Rotational Invariance with Bessel-Convolutional Neural Networks} (Delchevalerie, 2021) \resourcetag{article} \resourcetag{video}
        \item \href{https://brilliant.org/wiki/spherical-harmonics/}{Spherical Harmonic} \resourcetag{blog}
        
    \end{itemize}
}

\subsubsection{Examples of architecture}
\label{app:subsec:examples-archi}

In this subsection, we display the architecture of \textbf{GemNet} \citep{gasteiger2021gemnet} to demonstrate that best-performing invariant methods often come at the cost of a complex functioning. We also 
describe below the well known SchNet \citep{schutt2017schnet} architecture, having as main objective to depict the widely used \textit{continuous convolution}. 

\textbf{SchNet} introduced the use of continuous filters to handle unevenly spaced data, e.g. atoms at arbitrary positions. Given 3D atom input positions $\cgv{x} \in \reals^{n \times 3}$, the continuous-filter convolutional layer $t$ requires a filter-generating function
\[
\l{\basis}^t: \mathbb{R}^3 \rightarrow \mathbb{R}^a,
\]
that maps positions to the corresponding discrete convolution filter values. This learnable filter generating function is modeled with a neural network where the (invariant scalar) output $\cs{s}_i^{t+1}$ of the continuous convolutional layer at position $\gv{x}_i$ is given by
\begin{equation}
\cs{s}_i^{t+1} = \sum_{j \in \nei_i} \cs{s}^t_j \odot \l{\basis}^t(\gv{x}_{ij}) = \sum_{j \in \nei_i} \cs{s}^t_j \odot \l{f}^t(\basis(d_{ij}))
\end{equation}
where "$\odot$" represents the element-wise multiplication, $\l{f}$ represents a two-layer MLP with softplus activation function and $\basis$ a radial basis function\appref{app:subsec:basis-functions}. These feature-wise convolutions are applied for computational efficiency. The interactions between feature maps are handled by separate object-wise or, specifically, atom-wise layers in SchNet.

\begin{figure}[h!]
    \centering
    \includegraphics[width=\linewidth]{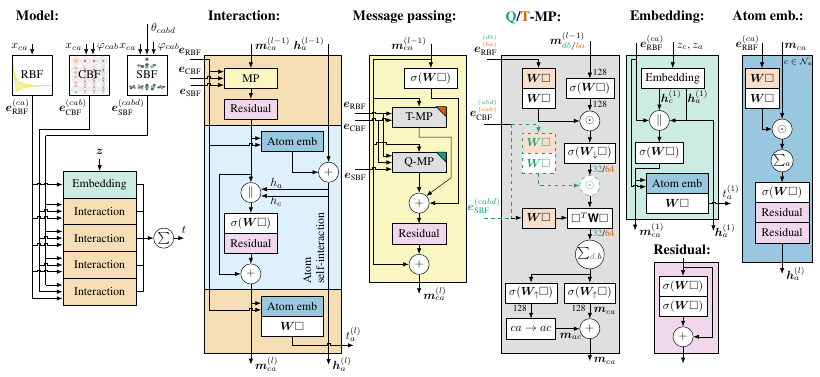}
    \caption{The full GemNet architecture. $\square$ denotes the layer's input, $\|$ concatenation, $\sigma$ the SiLU non-linearity, and orange a layer with weights shared across interaction blocks. Differences between two-hop message passing (Q-MP) and one-hop message passing (T-MP) are denoted by dashed lines. Numbers next to connecting lines denote embedding sizes. Taken from the original paper \citep{gasteiger2021gemnet}.}
    \label{app:fig:gemnet-archi}
\end{figure}

\goingfurther{
    \begin{itemize}[leftmargin=*,label=\ding{213}]
        \setlength\itemsep{1em}
        \item \href{https://towardsdatascience.com/graph-ml-in-2023-the-state-of-affairs-1ba920cb9232}{Graph ML in 2023: The State of Affairs}, (Galkin, 2022) \resourcetag{blog} \resourcetag{visual}
    \end{itemize}
}

\subsection{Expressive power of Geometric GNN}
\label{app:sec:expressivity}

\begin{figure}[t!]
    \centering
    \includegraphics[width=0.8\linewidth]{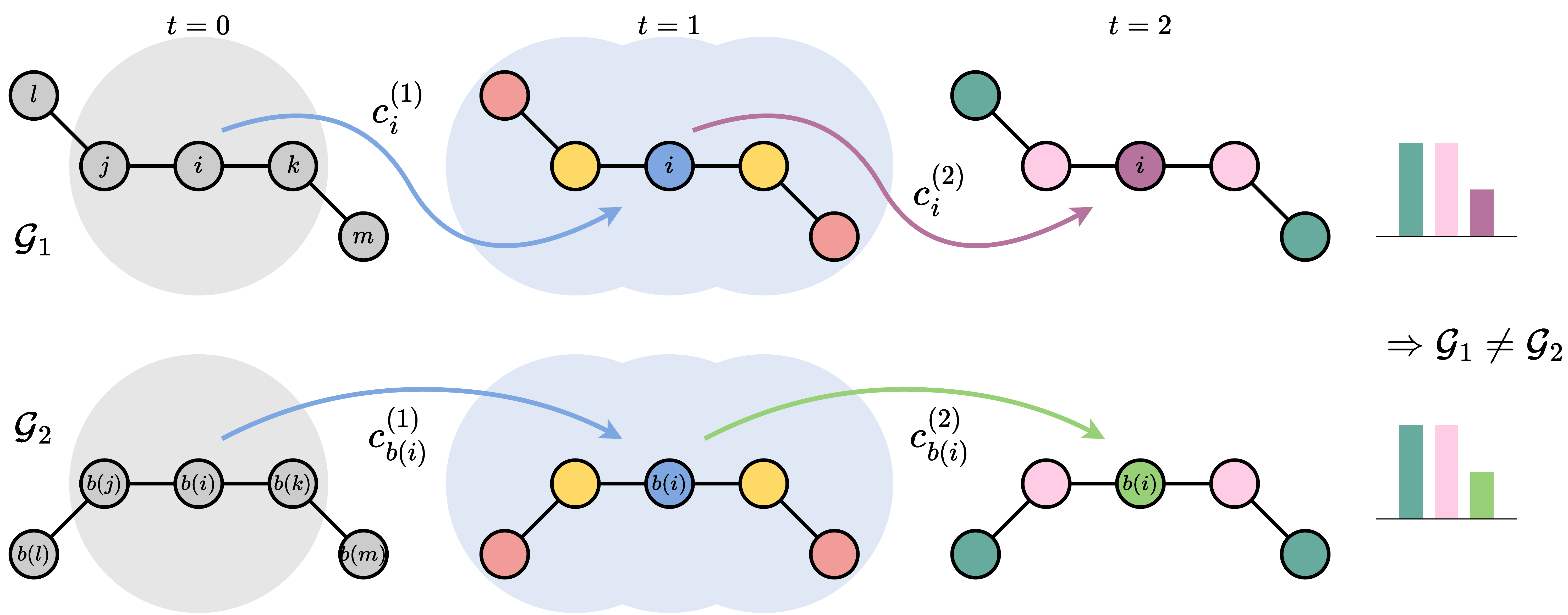}
    \caption{The Geometric Weisfeiler-Leman Test, an upper bound on the expressive power of equivariant GNNs \citep{joshi2022expressive}.
        GWL distinguishes non-isomorphic geometric graphs $\mathcal{G}_1$ and $\mathcal{G}_2$ by injectively assigning colours to distinct neighbourhood patterns, up to global symmetries (here $O(d)$).
        Each iteration expands the neighbourhood from which geometric information can be gathered (shaded for node $i$). Example inspired by \cite{schutt2021equivariantmp}.
    }
    \label{fig:gwl}
\end{figure}

The graph isomorphism problem and the Weisfeiler-Leman (WL) test for distinguishing non-isomorphic graphs have become a powerful tool for analysing the expressive power of traditional GNNs \citep{jegelka2022theory}.
It was shown by \cite{xu2018how, morris2019weisfeiler} that message passing GNNs are at most as powerful as WL at distinguishing non-isomorphic graphs and suffer from the same failure modes as WL.
The WL framework has since become a major driver of progress in designing more expressive GNNs \citep{dwivedi2020benchmarking, bodnar2021cwl}.

However, WL does not directly apply to geometric graphs  as they exhibit a stronger notion of geometric isomorphism that must accounts for spatial symmetries.
Two geometric graphs can only be \emph{geometrically isomorphic} if the underlying graphs are isomorphic \emph{and} the geometric attributes are equivalent, up to global group actions like rotations and reflections.

\begin{figure}
    \centering
    \includegraphics[width=\textwidth]{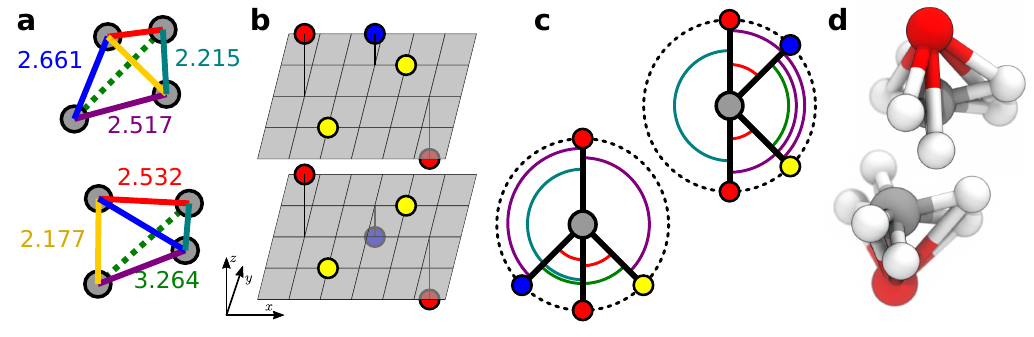}
    \caption{Counterexamples from \citep{pozdnyakov2022incompleteness} where pairs are distinct but cannot be discriminated by the unordered list of distances or distances and angles between atoms. They were created to demonstrate that some Geometric GNNs cannot distinguish between these pairs of structures using k-body scalarisation. (a) The two tetrahedra share the same list of pairwise distances, as per color coding. (b) The two structure share the same list of pairwise distances, and in addition the list of distances of each atom relative to its neighbors. (c) The two environments share the same list of distances and angles relative to the central (gray) atom. (d) The two environments share the same list of distances, angles and tetrahedra around the central (gray) atom.}
    \label{app:fig:counterexamples}
\end{figure}

\citet{joshi2022expressive} recently proposed the Geometric WL (GWL) framework for characterising the expressivity of Geometric GNNs by their ability to solve geometric graph isomorphism, i.e. to provide distinct representations for any two different geometric graphs, up to group actions. 
In addition to their theoretical contributions, they proposed several synthetic experiments to test new Geometric GNNs' expressivity in a practical manner.
For instance, one task asks models to distinguish between several counterexamples from \citet{pozdnyakov2022incompleteness}. These edge cases consists of pairs of configurations that are indistinguishable when comparing their set of k-body scalars, illustrated in \Cref{app:fig:counterexamples}. In other terms, any model for atom-centred properties that uses 2, 3 or 4-body order features will incorrectly give identical results for the different configurations. Hence, Geometric GNNs that do not process enough information to uniquely distinguish two atomic systems, such as SchNet or DimeNet, automatically fails this experiment.

\begin{tcolorbox}[enhanced,attach boxed title to top left={yshift=-2mm,yshifttext=-1mm,xshift = 10mm},
colback=cyan!3!white,colframe=cyan!75!black,colbacktitle=white, coltitle=black, title=Opinion,fonttitle=\bfseries,
boxed title style={size=small,colframe=cyan!75!black} ]
Under the GWL framework, Geometric GNNs require injective aggregation, update and readout functions to be maximally expressive or universal.
While some current architectures have claimed to be universal or \emph{complete} under specific conditions \citep{gasteiger2022gemnet, wang2022comenet}, we believe that a provably universal, equivariant GNN on sparse graphs with finite tensor and body order remains an open question.
\end{tcolorbox}

\goingfurther{
    \begin{itemize}[leftmargin=*,label=\ding{213}]
        \setlength\itemsep{1em}
        \item \href{https://arxiv.org/pdf/2301.09308.pdf}{On the Expressive Power of Geometric Graph Neural Networks.} \resourcetag{article}
        \item \href{https://github.com/chaitjo/geometric-gnn-dojo}{Geometric GNN Dojo}: Reference implementations and synthetic experiments to evaluate Geometric GNN expressivity in practice. \resourcetag{visual}
    \end{itemize}
}

\subsection{On inductive biases}
\label{app:sec:inductivebiases}

A major research area in deep learning is to find ways to express preferences over the kinds of functions we would like to solve our problems. For instance with Maximum A Posteriori (MAP) estimation one could use a prior distribution over a model's parameters to express a preference, a prior belief, over the possible values of the function's parameters. Those preferences are called \textit{inductive biases} because they are baked-in the algorithms in ways that voluntarily restrict the accessible function space during learning, or favour specific regions. Those restrictions are most often designed to improve data efficiency (the number of data points required to get to some level of performance) and/or generalization (the ability to keep performing well on new data). Typical inductive biases include: parameter regularization (as in the MAP example)~\citep{6796505, JMLR:v15:srivastava14a,Nusrat2018ACO}, enforcing invariance or equivariance to certain input transformations~\citep{lee2019set, rowley1998rotation,bronstein2017geometric} (as is the case for Geometric GNNs) or into the very architecture of the neural network~\citep{7780459, yu2015multiscale, battaglia2018relational,sherstinsky2020fundamentals,wang2023acmp}. Discovering, leveraging and evaluating inductive biases is therefore of paramount importance when developing learning approaches to solve real-world problems, especially when data (or compute) is scarce.

\goingfurther{
    \begin{itemize}[leftmargin=*,label=\ding{213}]
        \setlength\itemsep{1em}
        \item \href{https://arxiv.org/abs/2011.15091v4}{
Inductive Biases for Deep Learning of Higher-Level Cognition} (Goyal \& Bengio, 2020) \resourcetag{article} \resourcetag{technical}
        \setlength\itemsep{1em}
        \item \href{https://www.youtube.com/watch?v=xDdD0tKLxvM}{
Useful Inductive Biases for Deep Learning in Molecular Science} (Welling, 2022) \resourcetag{video}

    \end{itemize}
}

\section{An opinionated history of methods}
\label{sec:history-of-methods}

Similarly to other field, the first prediction models for molecular energy and forces relied on hand-crafted representations \citep{behler2016perspective, halgren1996merck, chmiela2018towards} built on physical properties. This was the case until recently, where research moved to end-to-end machine learning models based on Graph Neural Networks \citep{gori2005new}.

Multiple work came expanding the application of ML techniques to a broad set of materials modeling tasks ranging from solid-state \citep{zitnick2020introduction, miret2023the, lee2023matsciml} to molecular \citep{hoja2021qm7, ramakrishnan2014quantum} structures. Most existing GNN architectures have tried to incorporate physics-informed 3D symmetries, either directly in the model architecture, making model predictions explicitly invariant or equivariant to the desired transformations, or via the data. There are Geometric GNNs. 

A first line of Geometric GNNs were constructed to be \textbf{\text{E}(3)-invariant} by extracting scalar representations from atomic relative positions~\citep{unke2019physnet, klicpera2020directional, liu2021spherical, shuaibi2021rotation, ying2021transformers, adams2021learning}. The evolution was steady as we went from SchNet \citep{schutt2017schnet} using pair-wise distances, to DimeNet \citep{klicpera2020directional} adding bond angles, to SphereNet \citep{liu2022spherical} and GemNet \citep{gasteiger2021gemnet, gasteiger2022gemnet} additionally incorporating torsion angles (i.e. quadruplet of atoms). The scalarisation of geometric information enables to apply traditional message passing schemes (with any non-linearity) and the use of additional information enables models to distinguish a larger set of atomic systems. Nevertheless, this comes at a greater computational cost since GemNet must look at 3-hop neighbourhoods to compute torsion information for each update step, at both training and inference time. Besides, these models extract a set of pre-defined scalars representing geometric information and cannot represent equivariant properties directly, due to their invariant nature. 

In parallel, \textbf{Equivariant GNNs with Spherical Coordinates} \citep{thomas2018tensor, anderson2019cormorant, fuchs2020se, brandstetter2021geometric, batatia2022mace, frank2022so3krates}, also called Spherical EGNN, focused on enforcing equivariance using irreducible representations of the SO(3) group.  Such models build upon the concepts of steerability and equivariance introduced by Cohen and Welling \citep{cohen2016group}. They use spherical tensors as node embeddings and ensure equivariance to SO(3) by placing constraints on the operations that can be performed \citep{kondor2018clebsch}. Specifically, they compute linear operations with a generalized tensor product between the atom embeddings and edges' directions. Tensor Field Networks \citep{thomas2018tensor}, NequIP \citep{batzner2022nequip}, SEGNN \citep{brandstetter2021geometric}, MACE \citep{batatia2022mace}, Allegro \citep{musaelian2022learning}, Equiformer \citep{liao2022equiformer}, and various others lie in this category. They are commonly referred to as \textit{e3nn networks}. While these methods are expressive and generalize well, they can be hard to implement, constraint a lot the functional space and optimization landscape, and very computationally expensive at training and inference (due to the Clebsch-Gordan tensor product with spherical representations). 

For this reason, \textbf{Equivariant GNNs with Cartesian Coordinates} \citep{schutt2021equivariant, satorras2021n, tholke2022torchmd}, also called Cartesian EGNNs, started to model equivariant interactions in the Cartesian space, updating both scalar and vector representations. These GNNs achieve good performance while being relatively fast by avoiding the expensive equivariant operations of Spherical EGNNs. The authors manually design two separate sets of functions to deal with each type of representation and often use complex operations to mix their information, 
restricting the set of possible operations (e.g. vector element-wise dot product) to preserve equivariance.. As downsides, in addition to rendering the global architecture hard to understand, the decomposition of Cartesian tensors into spherical tensors offers many nice properties that Cartesian EGNNs lack. For instance, they lack structured, hierarchical and compact representation of geometric information, which improves model efficiency and memory utilization. They also don't capture information across different angular directions, which makes the model more sensitive to angular variations and dependencies present in atomic interactions. 

Overall, we have witnessed in recent years incredible progress in terms of model architectures across invariant and equivariant GNNs. However, all above approaches restrict the model learning space, reducing the number of possible operations inside the model architecture. Although this is theoretically desirable, in practice, it may hamper the learning capacity of the model. For this reason, we have seen the appearance of \textbf{unconstrained Geometric GNNs}. Unconstrained GNNs do not enforce symmetries via the model architecture. Instead, they attempt to implicitly learn them through simple data augmentation \citep{hu2021forcenet}; they relax them \citep{zitnick2022spherical} or they enforce them by mapping input data to a unique canonical space of all euclidean representations \citep{duval2023faenet}. 

In this work, we attempt to \textbf{bridge an existing gap in the literature} by
providing a holistic and opinionated overview of the field of Geometric GNNs, encompassing all important aspects. While we are the first work of such kind, we acknowledge the presence of some great recent works also attempting to bridge this gap \citep{wang2022graphsurvey, han2022geometrically, liu2023symmetry, atz2021geometric}. \cite{wang2022graphsurvey} focuses on the the molecular applications of GNNs; \cite{liu2023symmetry} proposes a benchmarking platform for various GNNs and molecular datasets; \cite{han2022geometrically, atz2021geometric} offer a short summary of the field. Overall, they all provide a descriptive and relatively general overview of the field, not always specific to Geometric GNNs. Our distinguishing contributions can be listed as follows: (1) a thorough description of all steps and variations in the Geometric GNNs modeling pipeline; (2) a novel taxonomy of methods containing a clear mathematical relation between them; (3) a concise description of how Geometric GNNs power different applications, with associated datasets; (4) a detailed list of promising future research directions, plus opinions on many ongoing discussions that divide the field; (5) the inclusion of almost all contextual material needed to understand the field; (6) a new notation scheme; (7) an exhaustive list of approaches and datasets that need to be updated by the community. We hope that our work will be useful to the whole community of researchers, helping experienced ones to efficiently navigate the field and guiding newcomers to integrate it.

\goingfurther{
    \begin{itemize}[leftmargin=*,label=\ding{213}]
        \setlength\itemsep{1em}
        \item \href{https://thegradient.pub/towards-geometric-deep-learning/}{
Towards Geometric Deep Learning} (Bronstein, 2023) \resourcetag{blog}
    \end{itemize}
}

\section{Data}
\label{app:sec:datasets}

\subsection{Data splits}
\label{app:subsec:data-split}

\begin{enumerate}
    \item \textbf{Random split}: ensures training, validation, and test data are sampled from the same underlying probability distribution.
    \begin{enumerate}
        \item [--] \textit{Pros}: Simple to implement. 
        \item [--] \textit{Cons}: May not preserve the underlying distribution of data. May result in variability in performance due to random variations in the data split. May not capture specific challenges or biases in the data.
        \item [--] \textit{When to use?} They can provide a good baseline evaluation, especially when the dataset is well-balanced and representative.
    \end{enumerate}
        \item \textbf{Stratified split}: The dataset is divided while maintaining a similar class distribution across sets (e.g., balanced representation of active and inactive compounds).It  ensures each of the training, validation, and test sets to cover the full range of provided labels. 
        \begin{enumerate}
            \item [--] \textit{Pros}: Helps ensure that each split contains a representative distribution of different classes or properties, reducing the risk of biased evaluations. 
            \item [--] \textit{Cons}: Requires class or property information for stratification, which may not always be available or applicable. Small or imbalanced datasets may still pose challenges. Not optimal to measure generalisation ability.
            \item [--] \textit{When to use ?} When there is a class or property imbalance in the dataset. 
        \end{enumerate}
        %
        %
        \item \textbf{Extrapolation split}: The dataset is divided such that some targeted molecules are placed in the test set without being included in the training set. This is often referred to as out-of-distribution (OOD). 
        \begin{enumerate}
            \item [--] \textit{Pros}: Aims to assess a model's ability to generalize to unseen chemical environments. It challenges the model to predict properties based on general chemical principles rather than memorizing specific training instances, yielding more robust models. 
            \item [--] \textit{Cons}: Such split requires careful curation of OOD examples that maintain chemical relevance while representing unseen combinations. The design of OOD examples may inadvertently introduce biases or artifacts (e.g. low density region), impacting the fairness of model evaluations.
            \item [--] \textit{When to use ?} When we care about ood generalisation of the model. For e.g., in materials or drug discovery applications. 
        \end{enumerate}
        \item \textbf{Time Split}: The dataset is divided based on a chronological order, such as using earlier time points for training and later time points for validation or testing.
        \begin{enumerate}
            \item [--] \textit{Pros}: Reflects real-world scenarios where models are trained on historical data and tested on future data. Allows evaluation of model performance under temporal variations.
            \item [--] \textit{Cons}: Assumes that data collected at different time points are representative of the same underlying distribution. May not be suitable for all datasets or tasks.
            \item [--] \textit{When to use?} In presence of meaningful temporal data. 
        \end{enumerate}
        \item \textbf{Group split}: The dataset is divided based on specific groups or categories present in the data (e.g., different targets, protein families, chemical series).
        \begin{enumerate}
            \item [--] \textit{Pros}: Enables evaluation of model performance on specific subsets of the data that may have distinct characteristics or challenges. Can provide insights into target or group-specific performance.
            \item [--] \textit{Cons}: Requires prior knowledge or information about the groups or categories. May introduce biases if the groups are not representative or if data in different groups have varying characteristics.
            \item [--] \textit{When to use?} When the data can be grouped based on specific characteristics or challenges.
        \end{enumerate}
\end{enumerate}

\subsection{Examples of predicted properties}
\label{app:subsec:prop-prediction-tasks-examples}

\begin{itemize}
    \item \textit{band gap}: determines a material's electronic behavior and is relevant in areas such as semiconductor design and solar cell applications.
    \item \textit{dielectric constant}: measures the ability of a substance or material to store electrical energy. It is an expression of the extent to which a material holds or concentrates electric flux. Dieletric constant has important applications in energy storage devices and electrical substation equipments. 
    \item \textit{Refractive index} is a measure of how light propagates through a material. It is an important property in optics and photonics applications, as it determines the material's ability to manipulate light.
    \item \textit{glass}: is a classification property indicating if a material is a glass former or not. Glass-forming materials are important in the field of materials science, as their amorphous structure offers unique properties and applications in fields like optics, electronics, and energy storage.
    \item \textit{jdft2d}: Exfoliation energy represents the energy required to separate or exfoliate a layered material into individual layers. This property is relevant in the field of 2D materials, where exfoliation plays a crucial role in obtaining thin layers with desired properties. 
    \item \textit{Formation energy} of a material provides insights into its stability and the energy involved in its formation. It is useful for understanding the feasibility of synthesizing new materials and their thermodynamic properties
    \item \textit{Phonon energy}: the frequency of the highest frequency optical phonon mode peak provides information about the lattice vibrations in a material. It is relevant for understanding thermal conductivity, phonon transport, and thermal properties of materials.
    \item \textit{Forces}: the forces acting on each atom, commonly used to perform molecular dynamics simulations. 
    \item \textit{Relaxed energy}: refers to the minimum energy state of a molecule or system, obtained through computational methods. It provides insights into the energetics of the molecule and aids in understanding its structural properties and interactions.
    \item \textit{Dipole Moment}: is a measure of the separation of positive and negative charges within a molecule. It quantifies the molecule's polarity and can be important for predicting its behavior in certain chemical reactions or interactions.
    \item \textit{HOMO} (Highest Occupied Molecular Orbital): refers to the highest energy level in a molecule that is occupied by electrons. It is an electronic property used to describe the reactivity and stability of molecules. 
    \item \textit{Polarizability}: is a property that describes the ability of a molecule to be deformed by an external electric field. It reflects the molecule's response to changes in the electric field and is important in studying intermolecular forces and interactions.
    \item \textit{Heat Capacity}: measures the amount of heat energy required to raise the temperature of a substance. It can be used to understand and predict how a molecule will respond to changes in temperature, providing insights into its thermodynamic properties.
    \item \textit{Toxicity}: refers to the degree to which a substance can cause harm or adverse effects to living organisms. Machine learning models trained to predict toxicity can help identify potentially toxic compounds and aid in drug development and safety assessment.
    \item \textit{Solubility}: is the property of a substance to dissolve in a solvent to form a homogeneous solution. Solubility prediction involves determining the likelihood of a compound to dissolve in a particular solvent or under specific conditions. Accurate solubility predictions are valuable in environmental studies, drug discovery and material science.
\end{itemize}


\subsection{Atom-type rescaling}
\label{app:subsec:data-preproc}

In this data pre-processing approach, the target values (e.g., molecule energy) are shifted or corrected based on a learned or calculated energy contribution from individual atom types. The idea is to factor out the contributions from different atom types and create a more interpretable target variable that represents the interaction energy between atoms. This can help the model focus on learning the residual energy variations after accounting for the atomic contributions.

To calculate the shift values (e.g. energy contributions) of each atom type based on the training dataset, one must first construct a matrix $\mA$ of shape $(n, z_{max}+1)$, where $n$ is the number of dataset samples and $z_{max}$ represents the largest atomic number. This matrix counts the occurrences of each atomic type for each structure.

Next, one must solve a linear equation $\mA \vx = \vq$ to calculate the per atom shifts, where $\vx$ is a vector of shape $(z_{max} + 1)$ representing the unknown shifts for each atom type, and $\vq$ is the target quantity of shape $(n)$, which can be the ground truth energy. The solution $\vx$ provides the shifts for each atom type.

Finally, to obtain the shifted energies $y'$, the sum of the shifts corresponding to every atom in the atomic system is subtracted from the ground truth energy $y$. This ensures that the energy values are adjusted according to the atom-type specific shifts, effectively balancing the contributions of different atom types to the overall potential energy.
\begin{equation*}
    y' = y - \sum_{i \in \graph{G}_j} x_i 
\end{equation*}



Target value shifting is particularly common in the field of computational chemistry. It's used to improve the prediction accuracy of ML models by factoring out known atomic contributions from the target values, allowing to focus on the finer details of atomic interactions. In different terms, the objective is to obtain rescaled target values exhibiting a desirable distribution and reduced variance, which enhances the quality of the training data and improves the stability/convergence of the GNN model during training. While it requires accurate calculation of energy contributions, this approach can help capture subtleties that might be challenging to learn directly from raw data. For instance, it is commonly applied on the QM7-X dataset \citep{hoja2021qm7}, as described \href{https://github.com/thorben-frank/mlff/blob/v0.1/mlff/examples/02_Multiple_Structure_Training.ipynb}{in this work}.

Finally, note that although slightly less common on materials and molecules, atom type shift can be applied to input features instead of target variables. This is the case when ML models use atomic properties having very different scales, ultimately making it easier for the model to learn patterns across different atom types. 

\begin{tcolorbox}[enhanced,attach boxed title to top left={yshift=-2mm,yshifttext=-1mm,xshift = 10mm},
colback=cyan!3!white,colframe=cyan!75!black,colbacktitle=white, coltitle=black, title=Opinion,fonttitle=\bfseries,
boxed title style={size=small,colframe=cyan!75!black} ]
If utilised, this pre-processing step must be reported clearly in the paper for reproducibility, with corresponding shift values. 
\end{tcolorbox}


\end{document}